\documentclass[twoside,11pt]{article}

\usepackage{jmlr2e}
\usepackage{hyperref}
\usepackage{amsmath}
\usepackage[export]{adjustbox}
\usepackage{subcaption}
\usepackage{caption}
\usepackage{wrapfig}
\usepackage{color}
\usepackage{multirow}
\usepackage{rotating}
\usepackage{graphicx}

\long\def\ignorethis#1{}

\newcommand{\pr}{^\prime}
\newcommand{\tr}{^\mathrm{T}}
\newcommand{\inv}{^{-1}}

\newcommand{\gauss}{\mathcal{N}}

\newcommand{\trace}{\text{tr}}

\newcommand{\vnorm}[1]{\|#1\|}
\newcommand{\lscnorm}[1]{\ell_{12}(#1)}

\newcommand{\badmmtraj}{\phi^{\trajdist}}
\newcommand{\badmmpol}{\phi^{\params}}
\newcommand{\lagtraj}{\mathcal{L}_{\trajdist}}
\newcommand{\lagpol}{\mathcal{L}_{\params}}

\newcommand{\admmrate}{\alpha}

\newcommand{\lgmult}{\lambda}

\newcommand{\lgmut}{\lambda_{\mu t}}
\newcommand{\admmrho}{\nu}

\newcommand{\trajdist}{p}
\newcommand{\policy}{\pi}

\newcommand{\params}{\theta}

\newcommand{\cost}{\ell}
\newcommand{\state}{\mathbf{x}}
\newcommand{\action}{\mathbf{u}}
\newcommand{\obs}{\mathbf{o}}

\newcommand{\traj}{\tau}

\newcommand{\trajmu}{\mu^\trajdist}

\newcommand{\polmu}{\mu^\policy}
\newcommand{\polsig}{\Sigma^\policy}

\newcommand{\ucovar}{\mathbf{C}}

\newcommand{\ucovart}{\mathbf{C}_t}

\newcommand{\detpolicy}{g}

\newcommand{\velx}{v_x}
\newcommand{\posy}{p_y}
\newcommand{\pos}{\mathbf{p}}
\newcommand{\torquepen}{w_\action}
\newcommand{\pospen}{w_\pos}
\newcommand{\velpen}{w_v}
\newcommand{\heightpen}{w_h}

\newcommand{\channel}{c}

\newcommand{\softmaxpix}{s_{cij}}

\newcommand{\responsepix}{a_{cij}}
\newcommand{\responsepixprime}{a_{ci'j'}}

\newcommand{\ms}{\text{m/s}}

\newcommand{\kl}{D_\text{KL}}
\newcommand{\ent}{\mathcal{H}}

\newcommand{\fc}{f_c}
\newcommand{\fx}{f_\state}
\newcommand{\fu}{f_\action}
\newcommand{\fct}{f_{c t}}
\newcommand{\fxt}{f_{\state t}}
\newcommand{\fut}{f_{\action t}}

\newcommand{\fyt}{f_{\state\action t}}

\newcommand{\Qxt}{Q_{\state t}}
\newcommand{\Qut}{Q_{\action t}}
\newcommand{\Qyt}{Q_{\state\action t}}
\newcommand{\Qxxt}{Q_{\state,\state t}}
\newcommand{\Quut}{Q_{\action,\action t}}
\newcommand{\Quxt}{Q_{\action,\state t}}

\newcommand{\Qyyt}{Q_{\state\action,\state\action t}}

\newcommand{\Vxt}{V_{\state t}}
\newcommand{\Vxxt}{V_{\state,\state t}}
\newcommand{\Vxtp}{V_{\state t+1}}
\newcommand{\Vxxtp}{V_{\state,\state t+1}}
\newcommand{\kpol}{\mathbf{k}}
\newcommand{\Kpol}{\mathbf{K}}

\newcommand{\noise}{\mathbf{F}}

\newcommand{\tcgradt}{\tilde{c}_{\state\action t}}
\newcommand{\tchesst}{\tilde{c}_{\state\action,\state\action t}}

\newcommand{\st}{\state_t}
\newcommand{\at}{\action_t}
\newcommand{\ot}{\obs_t}

\newcommand{\empsig}{\hat{\Sigma}}
\newcommand{\empmu}{\hat{\mu}}
\newcommand{\empn}{N}
\newcommand{\priorphi}{\mathbf{\Phi}}
\newcommand{\priormu}{\mu_0}

\newcommand{\priorm}{m}
\newcommand{\priorn}{n_0}

\jmlrheading{17}{2016}{1-40}{10/15}{4/16}{Sergey Levine, Chelsea Finn, Trevor Darrell, and Pieter Abbeel}

\ShortHeadings{End-to-End Training of Deep Visuomotor Policies}{Levine, Finn, Darrell, and Abbeel}
\firstpageno{1}

\begin{document}

\title{End-to-End Training of Deep Visuomotor Policies}

\author{\name Sergey Levine$^\dagger$ \email svlevine@eecs.berkeley.edu \\
\name Chelsea Finn$^\dagger$ \email cbfinn@eecs.berkeley.edu \\
\name Trevor Darrell \email trevor@eecs.berkeley.edu \\
\name Pieter Abbeel \email pabbeel@eecs.berkeley.edu \\
       {\addr Division of Computer Science \\
       University of California\\
       Berkeley, CA 94720-1776, USA\\
       }{\footnotesize\normalfont $^\dagger$These authors contributed equally.}}

\editor{Jan Peters}

\maketitle

\begin{abstract}

Policy search methods can allow robots to learn control policies for a wide range of tasks, but practical applications of policy search often require hand-engineered components for perception, state estimation, and low-level control. In this paper, we aim to answer the following question: does training the perception and control systems jointly end-to-end provide better performance than training each component separately? To this end, we develop a method that can be used to learn policies that map raw image observations directly to torques at the robot's motors. The policies are represented by deep convolutional neural networks (CNNs) with 92,000 parameters, and are trained using a guided policy search method, which transforms policy search into supervised learning, with supervision provided by a simple trajectory-centric reinforcement learning method. We evaluate our method on a range of real-world manipulation tasks that require close coordination between vision and control, such as screwing a cap onto a bottle, and present simulated comparisons to a range of prior policy search methods.

\end{abstract}


\section{Introduction}
\label{sec:intro}

Robots can perform impressive tasks under human control, including surgery~\citep{lcdm-rsacp-04} and household chores~\citep{wbvs-tprdp-08}. However, designing the perception and control software for autonomous operation remains a major challenge, even for basic tasks. Policy search methods hold the promise of allowing robots to automatically learn new behaviors through experience \citep{kop-rlarm-10,drf-lclcm-11,krps-lfcpc-11,dnp-spsr-13}. However, policies learned using such methods often rely on a number of hand-engineered components for perception and control, so as to present the policy with a more manageable and low-dimensional representation of observations and actions. The vision system in particular can be complex and prone to errors, and it is typically not improved during policy training, nor adapted to the goal of the task.

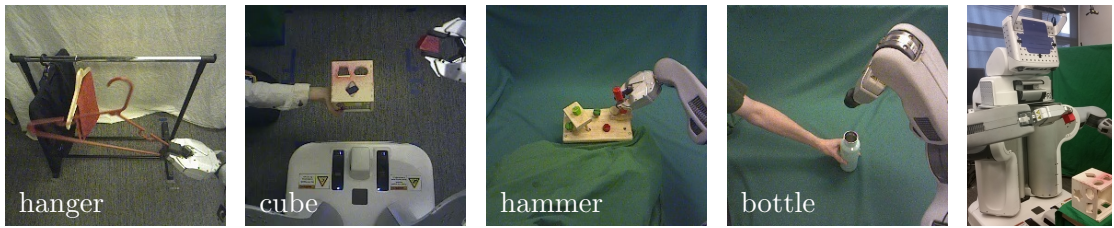
\begin{figure}
\setlength{\unitlength}{0.60\columnwidth}
\begin{picture}(1.9,0.3) \linethickness{0.5pt}
\put(0.025,-0.02){\includegraphics[height=0.192\columnwidth]{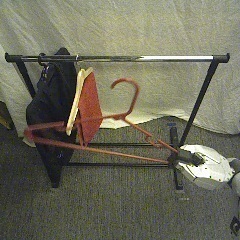}}
\put(0.375,-0.02){\includegraphics[height=0.192\columnwidth]{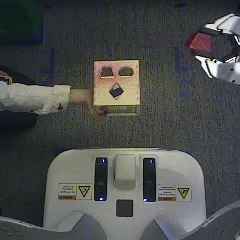}}
\put(0.725,-0.02){\includegraphics[height=0.192\columnwidth]{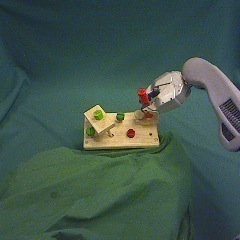}}
\put(1.075,-0.02){\includegraphics[height=0.192\columnwidth]{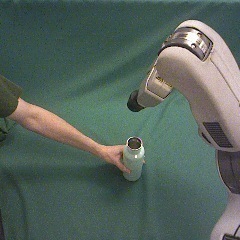}}

\put(1.425,-0.02){\includegraphics[height=0.192\columnwidth]{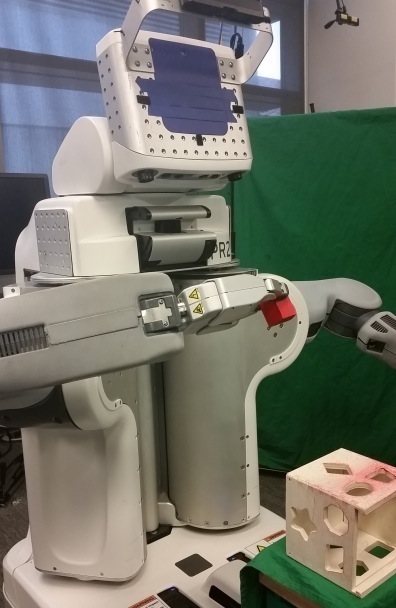}}

\put(0.045,0.0){\textcolor{white}{hanger}}
\put(0.395,0.0){\textcolor{white}{cube}}
\put(0.745,0.0){\textcolor{white}{hammer}}
\put(1.095,0.0){\textcolor{white}{bottle}}

\end{picture}
\caption{Our method learns visuomotor policies that directly use camera image observations (left) to set motor torques on a PR2 robot (right).
\label{fig:teaser}
}
\end{figure}

In this article, we aim to answer the following question: can we acquire more effective policies for sensorimotor control if the perception system is trained jointly with the control policy, rather than separately? In order to represent a policy that performs both perception and control, we use deep neural networks. Deep neural network representations have recently seen widespread success in a variety of domains, such as computer vision and speech recognition, and even playing video games.
However, using deep neural networks for real-world sensorimotor policies, such as robotic controllers that map image pixels and joint angles to motor torques, presents a number of unique challenges.
Successful applications of deep neural networks typically rely on large amounts of data and direct supervision of the output, neither of which is available in robotic control.
Real-world robot interaction data is scarce, and task completion is defined at a high level by means of a cost function, which means that the learning algorithm must determine on its own which action to take at each point. From the control perspective, a further complication is that observations from the robot's sensors do not provide us with the full state of the system. Instead, important state information, such as the positions of task-relevant objects, must be inferred from inputs such as camera images.

We address these challenges by developing a guided policy search algorithm for sensorimotor deep learning, as well as a novel CNN architecture designed for robotic control. Guided policy search converts policy search into supervised learning, by iteratively constructing the training data using an efficient model-free trajectory optimization procedure. We show that this can be formalized as an instance of Bregman ADMM (BADMM) \citep{wb-badmm-14}, which can be used to show that the algorithm converges to a locally optimal solution. In our method, the full state of the system is observable at training time, but not at test time. For most tasks, providing the full state simply requires positioning objects in one of several known positions for each trial during training. At test time, the learned CNN policy can handle novel, unknown configurations, and no longer requires full state information. Since the policy is optimized with supervised learning, we can use standard methods like stochastic gradient descent for training. Our CNNs have 92,000 parameters and 7 layers, including a novel spatial feature point transformation that provides accurate spatial reasoning and reduces overfitting. This allows us to train our policies with relatively modest amounts of data and only tens of minutes of real-world interaction time.

We evaluate our method by learning policies for inserting a block into a shape sorting cube, screwing a cap onto a bottle, fitting the claw of a toy hammer under a nail with various grasps, and placing a coat hanger on a rack with a PR2 robot (see Figure~\ref{fig:teaser}). These tasks require localization, visual tracking, and handling complex contact dynamics. Our results demonstrate improvements in consistency and generalization from training visuomotor policies end-to-end, when compared to training the vision and control components separately. We also present simulated comparisons that show that guided policy search outperforms a number of prior methods when training high-dimensional neural network policies. Some of the material in this article has previously appeared in two conference papers \citep{la-lnnpg-14,lwa-lnnpg-15}, which we extend to introduce visual input into the policy.

\section{Related Work}
\label{sec:related}

Reinforcement learning and policy search methods \citep{g-srlal-90,w-ssgfa-92} have been applied in robotics for playing games such as table tennis \citep{kop-rlarm-10}, object manipulation \citep{g-scuud-95,ps-rlmsp-08,kmkls-mtlh-10,drf-lclcm-11,krps-lfcpc-11}, locomotion \citep{bf-bdwrl-97,kp-pgrlf-04,tzs-spgrl-04,gpw-fbwrc-06,emmnc-lcbbl-08}, and flight \citep{nkjs-iahfr-04}. Several recent papers provide surveys of policy search in robotics \citep{dnp-spsr-13,kbp-rlrs-13}. Such methods are typically applied to one component of the robot control pipeline, which often sits on top of a hand-designed controller, such as a PD controller, and accepts processed input, for example from an existing vision pipeline \citep{krps-lfcpc-11}. Our method learns policies that map visual input and joint encoder readings directly to the torques at the robot's joints. By learning the entire mapping from perception to control, the perception layers can be adapted to optimize task performance, and the motor control layers can be adapted to imperfect perception.

We represent our policies with convolutional neural networks (CNNs). CNNs have a long history in computer vision and deep learning \citep{f-nson-80,lbdhh-hdrbp-89,s-dlnn-15}, and have recently gained prominence due to excellent results on a number of vision benchmarks \citep{cmmgs-fhpcn-11,ksh-incdc-12,cms-mcdnn-12,gddm-fhaod-13,tjlb-jcngm-14,lbh-dl-15,hzrs-drlir-15}.
Most applications of CNNs focus on classification, where locational information is discarded by means of successive pooling layers to provide for invariance \citep{lgrn-cdbn-09}. Applications to localization typically either use a sliding window \citep{gddm-fhaod-13} or object proposals \citep{dh-ciop-10,uvgs-ss-13,gddm-rcnn-14} to localize the object, reducing the task to classification, perform regression to a heatmap of manually labeled keypoints \citep{tjlb-jcngm-14}, requiring precise knowledge of the object position in the image and camera calibration, or use 3D models to localize previously scanned objects \citep{psgs-3ddpm-12,sl-3dgoc-07}. Many prior robotic applications of CNNs do not directly consider control, but employ CNNs for the perception component of a larger robotic system~\citep{hses-llrva-09,sjs-ropt-15,lls-dldrg-15,pg-ssltg-15}. We use a novel CNN architecture for our policies that automatically learn feature points that capture spatial information about the scene, without any supervision beyond the information from the robot's encoders and camera.

Applications of deep learning in robotic control have been less prevalent in recent years than in visual recognition. Backpropagation through the dynamics and the image formation process is typically impractical, since they are often non-differentiable, and such long-range backpropagation can lead to extreme numerical instability, since the linearization of a suboptimal policy is likely to be unstable. This issue has also been observed in the related context of recurrent neural networks \citep{hbfs-gfrnd-01,pb-odtrn-12}. The high dimensionality of the network also makes reinforcement learning difficult \citep{dnp-spsr-13}. Pioneering early work on neural network control used small, simple networks \citep{p-alvin-89,hszg-nncss-92,bg-nnr-92,lyj-nnc-98,bzgs-rrlim-03,mgwnk-srhsl-06}, and has largely been supplanted by methods that use carefully designed policies that can be learned efficiently with reinforcement learning \citep{kbp-rlrs-13}. More recent work on sensorimotor deep learning has tackled simple task-space motions \citep{lks-dmpc-15,lr-avsrg-13} and used unsupervised learning to obtain low-dimensional state spaces from images \citep{rlv-arlrv-12}. Such methods have been demonstrated on tasks with a low-dimensional underlying structure: \cite{lks-dmpc-15} controls the end-effector in 2D space, while \cite{rlv-arlrv-12} controls a 2-dimensional slot car with 1-dimensional actions. Our experiments include full torque control of 7-DoF robotic arms interacting with objects, with 30-40 state dimensions. In simple synthetic environments, control from images has been addressed with image features \citep{jp-cllvc-07}, nonparametric methods \citep{vpn-lnpcp-15}, and unsupervised state-space learning \citep{bgsmo-casrl-13,jb-srlru-14}. CNNs have also been trained to play video games with Q-learning, Monte Carlo tree search, and stochastic search \citep{mksga-padrl-13,kcsg-elsnnv-13,gsllw-amcts-14}, and have been applied to simple simulated control tasks \citep{wsbr-e2c-15,lhphe-ccdrl-15}. However, such methods have only been demonstrated on synthetic domains that lack the visual complexity of the real world, and require an impractical number of samples for real-world robotic learning. Our method is sample efficient, requiring only minutes of interaction time. To the best of our knowledge, this is the first method that can train deep visuomotor policies for complex, high-dimensional manipulation skills with direct torque control.

Learning visuomotor policies on a real robot requires handling complex observations and high dimensional policy representations. We tackle these challenges using guided policy search. In guided policy search, the policy is optimized using supervised learning, which scales gracefully with the dimensionality of the policy. The training set for supervised learning can be constructed using trajectory optimization under known dynamics \citep{lk-gps-13,lk-vpsto-13,lk-lcnnp-14,mt-cbfat-14} and trajectory-centric reinforcement learning methods that operate under unknown dynamics \citep{la-lnnpg-14,lwa-lnnpg-15}, which is the approach taken in this work. In both cases, the supervision is adapted to the policy, to ensure that the final policy can reproduce the training data. The use of supervised learning in the inner loop of iterative policy search has also been proposed in the context of imitation learning \citep{rgb-rilsp-11,rmswd-lmruc-13}. However, such methods typically do not address the question of how the supervision should be adapted to the policy.

The goal of our approach is also similar to visual servoing, which performs feedback control on feature points in a camera image \citep{ecr-navsr-92,mkd-vbcqp-14,whb-reecu-96}. However, our visuomotor policies are entirely learned from real-world data, and do not require feature points or feedback controllers to be specified by hand. This allows our method much more flexibility in choosing how to use the visual signal. Our approach also does not require any sort of camera calibration, in contrast to many visual servoing methods (though not all -- see e.g. \citet{jfn-eeuvs-97,ya-auvs-94}).

\section{Background and Overview}

In this section, we define the visuomotor policy learning problem and present an overview of our approach. The core component of our approach is a guided policy search algorithm that separates the problem of learning visuomotor policies into separate supervised learning and trajectory learning phases, each of which is easier than optimizing the policy directly. We also discuss a policy architecture suitable for end-to-end learning of vision and control, and a training setup that allows our method to be applied to real robotic platforms.

\subsection{Definitions and Problem Formulation}
\label{sec:defs}

In policy search, the goal is to learn a policy $\policy_\params(\at|\ot)$ that allows an agent to choose actions $\at$ in response to observations $\ot$ to control a dynamical system, such as a robot. The policy comes from some parametric class parameterized by $\params$, which could be, for example, the weights of a neural network. The system is defined by states $\st$, actions $\at$, and observations $\ot$. For example, $\st$ might include the joint angles of the robot, the positions of objects in the world, and their time derivatives, $\at$ might consist of motor torque commands, and $\ot$ might include an image from the robot's onboard camera. In this paper, we address finite horizon episodic tasks with $t\in[1,\dots,T]$. The states evolve in time according to the system dynamics $p(\state_{t+1}|\st,\at)$, and the observations are, in general, a stochastic consequence of the states, according to $p(\ot|\st)$. Neither the dynamics nor the observation distribution are assumed to be known in general. For notational convenience, we will use $\policy_\params(\at|\st)$ to denote the distribution over actions under the policy conditioned on the state. However, since the policy is conditioned on the observation $\ot$, this distribution is in fact given by $\policy_\params(\at|\st) = \int \policy_\params(\at|\ot)p(\ot|\st) d\ot$. The dynamics and $\policy_\params(\at|\st)$ together induce a distribution over trajectories $\traj = \{\state_1,\action_1,\state_2,\action_2,\dots,\state_T,\action_T\}$:
\[
\policy_\params(\traj) = p(\state_1)\prod_{t=1}^T \policy_\params(\at|\st) p(\state_{t+1}|\st,\at).
\]
The goal of a task is given by a cost function $\cost(\st,\at)$, and the objective in policy search is to minimize the expectation $E_{\policy_\params(\traj)}[\sum_{t=1}^T \cost(\st,\at)]$, which we will abbreviate as $E_{\policy_\params(\traj)}[\cost(\traj)]$. A summary of the notation used in the paper is provided in Table~\ref{tbl:notation}.

\begin{table}[ht]
\begin{center}
\footnotesize{
\begin{tabular}{| l | l | l |}
\hline
\!\!{\bf symbol} & \!\!{\bf definition} & \!\!{\bf example/details} \\
\hline
\!\!$\state_t$ & \!\!\parbox[c][0.4in][c]{0.41\columnwidth}{Markovian system state at time step $t \in [1, T]$}\!\! & \!\!\parbox[c][0.55in][c]{0.41\columnwidth}{joint angles, end-effector pose, object positions, and their velocities; dimensionality: 14 to 32}\!\! \\
\hline
\!\!$\action_t$ & \!\!\parbox[c][0.4in][c]{0.41\columnwidth}{control or action at time step \mbox{$t \in [1, T]$}}\!\! & \!\!\parbox[c][0.4in][c]{0.41\columnwidth}{joint motor torque commands; dimensionality: 7 (for the PR2 robot)}\!\! \\
\hline
\!\!$\obs_t$ & \!\!\parbox[c][0.4in][c]{0.41\columnwidth}{observation at time step \mbox{$t \in [1, T]$}}\!\! & \!\!\parbox[c][0.55in][c]{0.41\columnwidth}{RGB camera image, joint encoder readings \& velocities, end-effector pose; dimensionality: around 200,000}\!\! \\
\hline
\!\!$\traj$\!\! & \!\!\parbox[c][0.4in][c]{0.41\columnwidth}{trajectory:\\$\traj = \{\state_1,\action_1,\state_2,\action_2,\dots,\state_T,\action_T\}$}\!\! & \!\!\parbox[c][0.4in][c]{0.41\columnwidth}{notational shorthand for a sequence of states and actions}\!\! \\
\hline
\!\!$\cost(\st,\st)$\!\! & \!\!\parbox[c][0.4in][c]{0.41\columnwidth}{cost function that defines the goal of the task}\!\! & \!\!\parbox[c][0.4in][c]{0.41\columnwidth}{distance between an object in the gripper and the target}\!\! \\
\hline
\!\!$p(\state_{t+1}|\st,\at)$\!\! & \!\!\parbox[c][0.4in][c]{0.41\columnwidth}{unknown system dynamics}\!\! & \!\!\parbox[c][0.4in][c]{0.41\columnwidth}{physics that govern the robot and any objects it interacts with}\!\! \\
\hline
\!\!$p(\ot|\st)$\!\! & \!\!\parbox[c][0.4in][c]{0.41\columnwidth}{unknown observation distribution}\!\! & \!\!\parbox[c][0.4in][c]{0.41\columnwidth}{stochastic process that produces camera images from system state}\!\! \\
\hline
\!\!$\policy_\params(\at|\ot)$ & \!\!\parbox[c][0.4in][c]{0.41\columnwidth}{learned nonlinear global policy parameterized by weights $\params$}\!\! & \!\!\parbox[c][0.4in][c]{0.41\columnwidth}{convolutional neural network, such as the one in Figure~\ref{fig:nn}}\!\! \\
\hline
\!\!$\policy_\params(\at|\st)$ & \!\!\parbox[c][0.4in][c]{0.41\columnwidth}{$\int \policy_\params(\at|\ot)p(\ot|\st) d\ot$}\!\! & \!\!\parbox[c][0.4in][c]{0.41\columnwidth}{notational shorthand for observation-based policy conditioned on state}\!\! \\
\hline
\!\!$\trajdist_i(\at|\st)$ & \!\!\parbox[c][0.4in][c]{0.41\columnwidth}{learned local time-varying linear-Gaussian controller for initial state $\state_1^i$}\!\! & \!\!\parbox[c][0.4in][c]{0.41\columnwidth}{time-varying linear-Gaussian controller has form $\gauss(\Kpol_{ti}\st + \kpol_{ti}, \ucovar_{ti})$}\!\! \\
\hline
\!\!$\policy_\params(\traj)$ & \!\!\parbox[c][0.4in][c]{0.41\columnwidth}{trajectory distribution for $\policy_\params(\at|\st)$: $p(\state_1)\prod_{t=1}^T \policy_\params(\at|\st)p(\state_{t+1}|\st,\at)$}\!\! & \!\!\parbox[c][0.4in][c]{0.41\columnwidth}{notational shorthand for trajectory distribution induced by a policy}\!\! \\
\hline
\end{tabular}
}
\end{center}
\vspace{-0.2in}
\caption{Summary of the notation frequently used in this article.
\label{tbl:notation}
}
\vspace{-0.2in}
\end{table}

\subsection{Approach Summary}
\label{sec:overview}

\begin{wrapfigure}{r}{.37\columnwidth}
\setlength{\unitlength}{0.5\columnwidth}
\begin{picture}(1.5,0.94) \linethickness{0.5pt}

\put(-0.0,0.0){\includegraphics[width=0.37\columnwidth]{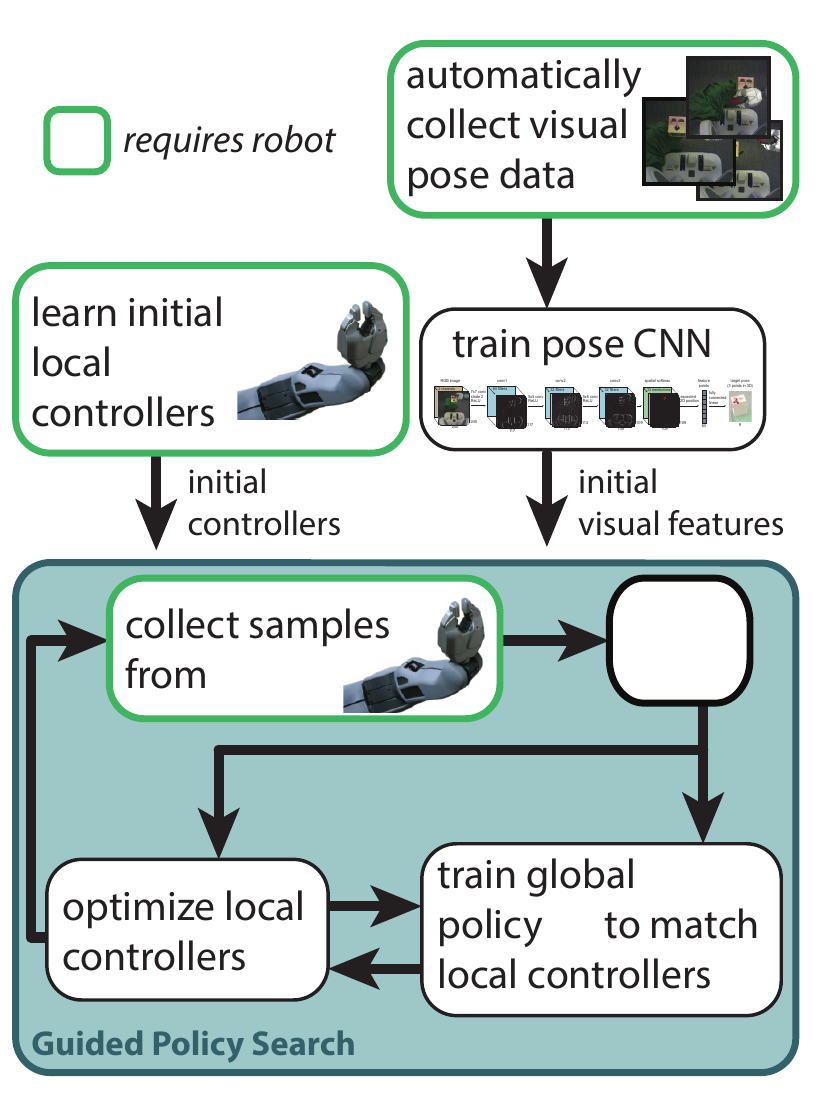}}

\put(0.20,0.368){\footnotesize{$\trajdist_i$}}
\put(0.238,0.115){\footnotesize{$\trajdist_i$}}
\put(0.658,0.095){\footnotesize{$\trajdist_i$}}

\put(0.5,0.14){\footnotesize{$\policy_\params$}}

\put(0.565,0.4){$\{\traj_i^j\}$}

\end{picture}
\caption{
Diagram of our approach, including the main guided policy search phase and initialization phases.
\label{fig:alg_diag}
\vspace{-0.15in}
}
\end{wrapfigure}
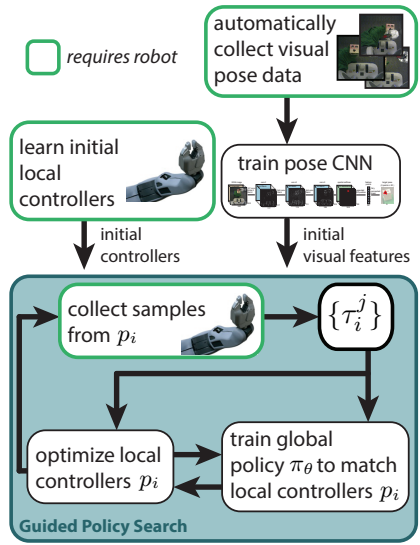

Our methods consists of two main components, which are illustrated in Figure~\ref{fig:alg_diag}. The first is a supervised learning algorithm that trains policies of the form $\policy_\params(\at|\ot) = \gauss(\polmu(\ot), \polsig(\ot))$, where both $\polmu(\ot)$ and $\polsig(\ot)$ are general nonlinear functions. In our implementation, $\polmu(\ot)$ is a deep convolutional neural network, while $\polsig(\ot)$ is an observation-independent learned covariance, though other representations are possible. The second component is a trajectory-centric reinforcement learning (RL) algorithm that generates guiding distributions $\trajdist_i(\at|\st)$ that provide the supervision used to train the policy. These two components form a policy search algorithm that can be used to learn complex robotic tasks using only a high-level cost function $\cost(\st,\at)$. During training, only samples from the guiding distributions $\trajdist_i(\at|\st)$ are generated by running rollouts on the physical system, which avoids the need to execute partially trained neural network policies on physical hardware.

Supervised learning will not, in general, produce a policy with good long-horizon performance, since a small mistake on the part of the policy will place the system into states that are outside the distribution in the training data, causing compounding errors. To avoid this issue, the training data must come from the policy's own state distribution \citep{rgb-rilsp-11}. We achieve this by alternating between trajectory-centric RL and supervised learning. The RL stage adapts to the current policy $\policy_\params(\at|\ot)$, providing supervision at states that are iteratively brought closer to the states visited by the policy. This is formalized as a variant of the BADMM algorithm \citep{wb-badmm-14} for constrained optimization, which can be used to show that, at convergence, the policy $\policy_\params(\at|\ot)$ and the guiding distributions $\trajdist_i(\at|\st)$ will exhibit the same behavior. This algorithm is derived in Section~\ref{sec:gps}. The guiding distributions are substantially easier to optimize than learning the policy parameters directly (e.g., using model-free reinforcement learning), because they use the full state of the system $\st$, while the policy $\policy_\params(\at|\ot)$ only uses the observations. This means that the method requires the full state to be known during training, but not at test time. This makes it possible to efficiently learn complex visuomotor policies, but imposes additional assumptions on the observability of $\st$ during training that we discuss in Section~\ref{sec:gps}.

\begin{figure*}[t]
\centering
\includegraphics[width=\textwidth]{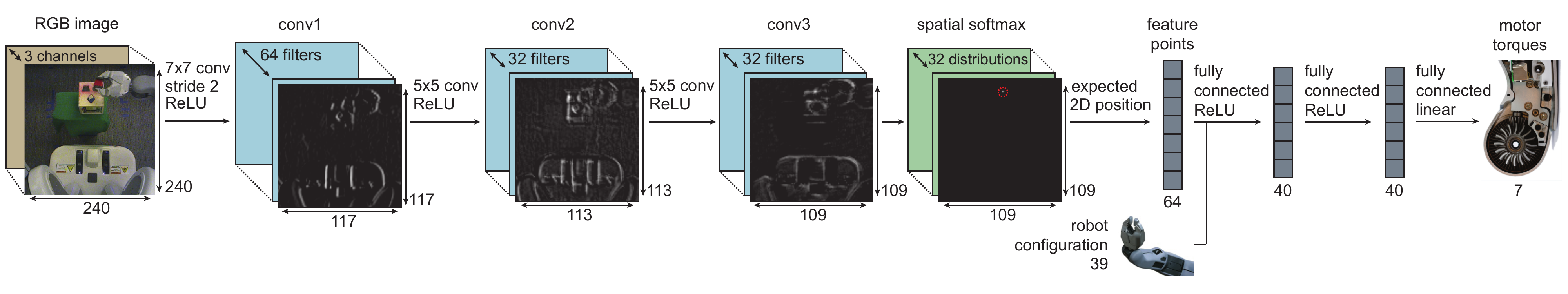}
\vspace{-0.3in}
\caption{Visuomotor policy architecture. The network contains three convolutional layers, followed by a spatial softmax and an expected position layer that converts pixel-wise features to feature points, which are better suited for spatial computations. The points are concatenated with the robot configuration, then passed through three fully connected layers to produce the torques.
}
\label{fig:nn}
\end{figure*}

When learning visuomotor tasks, the policy $\policy_\params(\at|\ot)$ is represented by a novel convolutional neural network (CNN) architecture, which we describe in Section~\ref{sec:training}. CNNs have enjoyed considerable success in computer vision \citep{lbh-dl-15}, but the most popular architectures rely on large datasets and focus on semantic tasks such as classification, often intentionally discarding spatial information. Our architecture, illustrated in Figure~\ref{fig:nn}, uses a fixed transformation from the last convolutional layer to a set of spatial feature points, which form a concise representation of the visual scene suitable for feedback control. Our network has 7 layers and around 92,000 parameters, which presents a major challenge for standard policy search methods \citep{dnp-spsr-13}.

To reduce the amount of experience needed to train visuomotor policies, we also introduce a pretraining scheme that allows us to train effective policies with a relatively small number of iterations. The pretraining steps are illustrated in Figure~\ref{fig:alg_diag}. The intuition behind our pretraining is that, although we ultimately seek to obtain sensorimotor policies that combine both vision and control, low-level aspects of vision can be initialized independently. To that end, we pretrain the convolutional layers of our network by predicting elements of $\st$ that are not provided in the observation $\ot$, such as the positions of objects in the scene. We also initially train the guiding trajectory distributions $\trajdist_i(\at|\st)$ independently of the convolutional network until the trajectories achieve a basic level of competence at the task, and then switch to full guided policy search with end-to-end training of $\policy_\params(\at|\ot)$. In our implementation, we also initialize the first layer filters from the model of \citet{sljsr-gdwc-14}, which is trained on ImageNet~\citep{ddsll-lshid-09} classification. The initialization and pretraining scheme is described in Section~\ref{sec:training}.

\section{Guided Policy Search with BADMM}
\label{sec:gps}

Guided policy search transforms policy search into a supervised learning problem, where the training set is generated by a simple trajectory-centric RL algorithm. This algorithm optimizes linear-Gaussian controllers $\trajdist_i(\at|\st)$, and is described in Section~\ref{sec:gpstraj}. We refer to the trajectory distribution induced by $\trajdist_i(\at|\st)$ as $\trajdist_i(\traj)$. Each $\trajdist_i(\at|\st)$ succeeds from different initial states.
For example, in the task of placing a cap on a bottle, these initial states correspond to different positions of the bottle. By training on trajectories for multiple bottle positions, the final CNN policy can succeed from all initial states, and can generalize to other states from the same distribution.

\begin{wrapfigure}{r}{.29\columnwidth}
\setlength{\unitlength}{0.29\columnwidth}
\begin{picture}(0.99,0.68) \linethickness{0.5pt}

\put(-0.03,-0.22){\includegraphics[width=0.29\columnwidth]{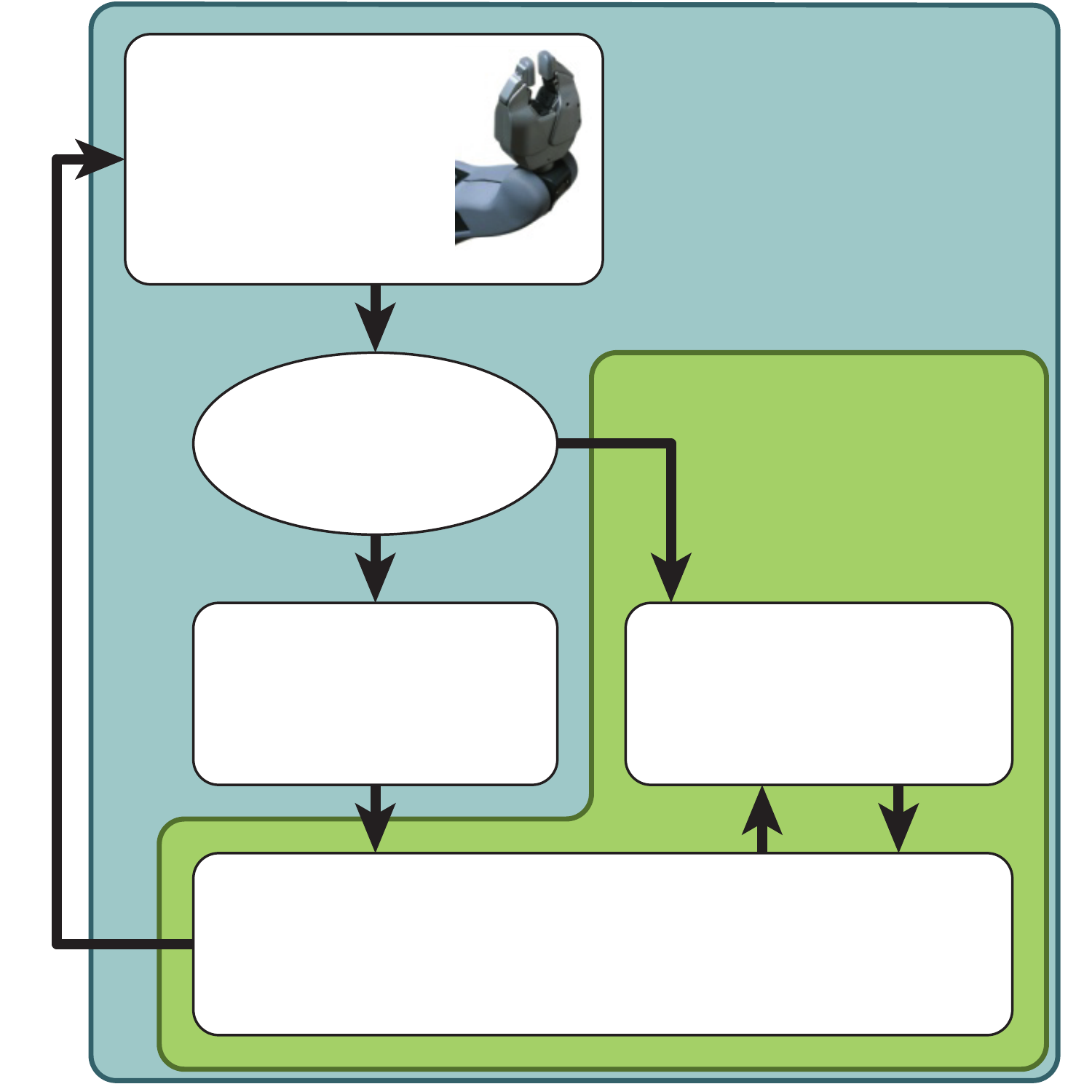}}

\scriptsize{
\put(0.105,0.69){run each}
\put(0.105,0.62){$\trajdist_i(\at|\st)$}
\put(0.105,0.55){on robot}

\put(0.20,0.385){samples}
\put(0.25,0.325){$\{\traj_i^j\}$}

\put(0.55,0.16){optimize $\policy_\params$}
\put(0.59,0.09){w.r.t. $\lagpol$}

\put(0.28,0.16){fit}
\put(0.175,0.09){dynamics}

\put(0.24,-0.07){optimize each $\trajdist_i(\traj)$}
\put(0.415,-0.14){w.r.t. $\lagtraj$}

\put(0.74,0.39){inner}
\put(0.76,0.32){loop}

\put(0.745,0.72){outer}
\put(0.76,0.65){loop}
}

\end{picture}
\label{fig:gps}
\end{wrapfigure}
The final policy $\policy_\params(\at|\ot)$ learned with guided policy search is only provided with observations $\ot$ of the full state $\st$, and the dynamics are assumed to be unknown. A diagram of this method, which corresponds to an expanded version of the guided policy search box in Figure~\ref{fig:alg_diag}, is shown on the right. In the outer loop, we draw sample trajectories $\{\traj_i^j\}$ for each initial state on the physical system by running the corresponding controller $\trajdist_i(\at|\st)$. The samples are used to fit the dynamics $\trajdist_i(\state_{t+1}|\st,\at)$ that are used to improve $\trajdist_i(\at|\st)$, and serve as training data for the policy. The inner loop alternates between optimizing each $\trajdist_i(\traj)$ and optimizing the policy to match these trajectory distributions. The policy is trained to predict the actions along each trajectory from the observations $\ot$, rather than the full state $\st$. This allows the policy to directly use raw observations at test time. This alternating optimization can be framed as an instance of the BADMM algorithm \citep{wb-badmm-14}, which converges to a solution where the trajectory distributions and the policy have the same state distribution. This allows greedy supervised training of the policy to produce a policy with good long-horizon performance.

\subsection{Algorithm Derivation}
\label{sec:gpsalg}

Policy search methods minimize the expected cost $E_{\policy_\params}[\cost(\traj)]$, where \mbox{$\traj = \{\state_1,\action_1,\dots,\state_T,\action_T\}$} is a trajectory, and \mbox{$\cost(\traj)=\sum_{t=1}^T \cost(\st,\at)$} is the cost of an episode. In the fully observed case, the expectation is taken under $\policy_\params(\traj) = p(\state_1)\prod_{t=1}^T\policy_\params(\at|\st)p(\state_{t+1}|\st,\at)$. The final policy $\policy_\params(\at|\ot)$ is conditioned on the observations $\ot$, but $\policy_\params(\at|\st)$ can be recovered as $\policy_\params(\at|\st) = \int \policy_\params(\at|\ot)p(\ot|\st) d\ot$. We will present the derivation in this section for $\policy_\params(\at|\st)$, but we do not require knowledge of $p(\ot|\st)$ in the final algorithm. As discussed in Section~\ref{sec:gpspolicy}, the integral will be evaluated with samples from the real system, which include both $\st$ and $\ot$. We begin by rewriting the expected cost minimization as a constrained problem:
\begin{equation}
\min_{\trajdist,\policy_\params} E_{\trajdist}[\cost(\traj)] \text{ s.t. } \trajdist(\at|\st) = \policy_\params(\at|\st) \,\, \forall \, \st,\at,t,\label{eqn:cgps}
\end{equation}
\noindent where we will refer to $\trajdist(\traj)$ as a guiding distribution. This formulation is equivalent to the original problem, since the constraint forces the two distributions to be identical. However, if we approximate the initial state distribution $p(\state_1)$ with samples $\state_1^i$, we can choose $\trajdist(\traj)$ to be a class of distributions that is much easier to optimize than $\policy_\params$, as we will show later. This will allow us to use simple local learning methods for $\trajdist(\traj)$, without needing to train the complex neural network policy $\policy_\params(\at|\ot)$ directly with reinforcement learning, which would require a prohibitive amount of experience on real physical systems.

The constrained problem can be solved by a dual descent method, which alternates between minimizing the Lagrangian with respect to the primal variables, and incrementing the Lagrange multipliers by their subgradient. Minimization of the Lagrangian with respect to $\trajdist(\traj)$ and $\params$ is done in alternating fashion: minimizing with respect to $\params$ corresponds to supervised learning (making $\policy_\params$ match $\trajdist(\traj)$), and minimizing with respect to $\trajdist(\traj)$ consists of one or more trajectory optimization problems. The dual descent method we use is based on BADMM \citep{wb-badmm-14}, a variant of ADMM \citep{bpcpe-dosla-11} that augments the Lagrangian with a Bregman divergence between the constrained variables. We use the KL-divergence as the Bregman constraint, which is particularly convenient for working with probability distributions. We will also modify the constraint $\trajdist(\at|\st) = \policy_\params(\at|\st)$ by multiplying both sides by $\trajdist(\st)$, to get $\trajdist(\at|\st)\trajdist(\st) = \policy_\params(\at|\st)\trajdist(\st)$. This constraint is equivalent, but has the convenient property that we can express the Lagrangian in terms of expectations. The BADMM augmented Lagrangians for $\params$ and $\trajdist$ are therefore given by
\begin{align*}
\lagpol(\params,\trajdist) &= \sum_{t=1}^T E_{\trajdist(\st,\at)}[\cost(\st,\at)] + E_{\trajdist(\st)\policy_\params(\at|\st)}[\lgmult_{\st,\at}] - E_{\trajdist(\st,\at)}[\lgmult_{\st,\at}] + \admmrho_t \badmmpol_t(\params,\trajdist) \\
\lagtraj(\trajdist,\params) &= \sum_{t=1}^T E_{\trajdist(\st,\at)}[\cost(\st,\at)] + E_{\trajdist(\st)\policy_\params(\at|\st)}[\lgmult_{\st,\at}] - E_{\trajdist(\st,\at)}[\lgmult_{\st,\at}] + \admmrho_t \badmmtraj_t(\params,\trajdist),
\end{align*}
where $\lgmult_{\st,\at}$ is the Lagrange multiplier for state $\st$ and action $\at$ at time $t$, and $\badmmpol_t(\params,\trajdist)$ are $\badmmtraj_t(\params,\trajdist)$ are expectations of the KL-divergences:
\begin{align*}
\badmmtraj_t(\trajdist,\params) &= E_{\trajdist(\st)}[\kl(\trajdist(\at|\st)\|\policy_\params(\at|\st))] \\
\badmmpol_t(\params,\trajdist) &= E_{\trajdist(\st)}[\kl(\policy_\params(\at|\st))\|\trajdist(\at|\st)].
\end{align*}
Dual descent with alternating primal minimization is then described by the following steps:
\begin{align*}
\params &\leftarrow \arg\min_{\params} \sum_{t=1}^T E_{\trajdist(\st)\policy_\params(\at|\st)}[\lgmult_{\st,\at}] + \admmrho_t \badmmpol_t(\params,\trajdist) \\
\trajdist &\leftarrow \arg\min_{\trajdist} \sum_{t=1}^T E_{\trajdist(\st,\at)}[\cost(\st,\at) - \lgmult_{\st,\at}] +
\admmrho_t \badmmtraj_t(\trajdist,\params) \\
\lgmult_{\st,\at} &\leftarrow \lgmult_{\st,\at} +  \admmrate \admmrho_t(\policy_\params(\at|\st)\trajdist(\st) - \trajdist(\at|\st)\trajdist(\st)).
\end{align*}
This procedure is an instance of BADMM, and therefore inherits its convergence guarantees.
Note that we drop terms that are independent of the optimization variables on each line. The parameter $\admmrate$ is a step size. As with most augmented Lagrangian methods, the weight $\admmrho_t$ is set heuristically, as described in Appendix~\ref{app:badmm}.

The dynamics only affect the optimization with respect to $\trajdist(\traj)$. In order to make this optimization efficient, we choose $\trajdist(\traj)$ to be a mixture of $N$ Gaussians $\trajdist_i(\traj)$, one for each initial state sample $\state_1^i$. This makes the action conditionals $\trajdist_i(\at|\st)$ and the dynamics $\trajdist_i(\state_{t+1}|\st,\at)$ linear-Gaussian, as discussed in Section~\ref{sec:gpstraj}. This is a reasonable choice when the system is deterministic, or the noise is Gaussian or small,
and we found that this approach is sufficiently tolerant to noise for use on real physical systems. Our choice of $\trajdist$ also assumes that the policy $\policy_\params(\at|\ot)$ is conditionally Gaussian. This is also reasonable, since the mean and covariance of $\policy_\params(\at|\ot)$ can be any nonlinear function of the observations $\ot$, which themselves are a function of the unobserved state $\st$. In Section~\ref{sec:gpstraj}, we show how these assumptions enable each $\trajdist_i(\traj)$ to be optimized very efficiently. We will refer to $\trajdist_i(\traj)$ as guiding distributions, since they serve to focus the policy on good, low-cost behaviors.

Aside from learning $\trajdist_i(\traj)$, we must choose a tractable way to represent the infinite set of constraints $\trajdist(\at|\st)\trajdist(\st) = \policy_\params(\at|\st)\trajdist(\st)$. One approximate approach proposed in prior work is to replace the exact constraints with expectations of features \citep{pma-reps-10}. When the features consist of linear, quadratic, or higher order monomial functions of the random variable, this can be viewed as a constraint on the moments of the distributions. If we only use the first moment, we get a constraint on the expected action: $E_{\trajdist(\at|\st)\trajdist(\st)}[\at] = E_{\policy_\params(\at|\st)\trajdist(\st)}[\at]$. If the stochasticity in the dynamics is low, as we assumed previously, the optimal solution for each $\trajdist_i(\tau)$ will have low entropy, making this first moment constraint a reasonable approximation. The KL-divergence terms in the augmented Lagrangians will still serve to softly enforce agreement between the higher moments. While this simplification is quite drastic, we found that it was more stable in practice than including higher moments, likely because these higher moments are harder to estimate accurately with a limited number of samples. The alternating optimization is now given by
\begin{align}
\params &\leftarrow \arg\min_{\params} \sum_{t=1}^T E_{\trajdist(\st)\policy_\params(\at|\st)}[\at\tr\lgmut] + \admmrho_t \badmmpol_t(\params,\trajdist) \label{eqn:lg1} \\
\trajdist &\leftarrow \arg\min_{\trajdist} \sum_{t=1}^T E_{\trajdist(\st,\at)}[\cost(\st,\at) - \at\tr\lgmut] +
\admmrho_t \badmmtraj_t(\trajdist,\params) \label{eqn:lg2} \\
\lgmut &\leftarrow \lgmut +  \admmrate \admmrho_t (E_{\policy_\params(\at|\st)\trajdist(\st)}[\at] - E_{\trajdist(\at|\st)\trajdist(\st)}[\at]),\nonumber
\end{align}
\noindent where $\lgmut$ is the Lagrange multiplier on the expected action at time $t$. In the rest of the paper, we will use $\lagpol(\params,\trajdist)$ and $\lagtraj(\trajdist,\params)$ to denote the two augmented Lagrangians in Equations~(\ref{eqn:lg1}) and (\ref{eqn:lg2}), respectively. In the next two sections, we will describe how $\lagtraj(\trajdist,\params)$ can be optimized with respect to $\trajdist$ under unknown dynamics, and how $\lagpol(\params,\trajdist)$ can be optimized for complex, high-dimensional policies. Implementation details of the BADMM optimization are presented in Appendix~\ref{app:badmm}.

\subsection{Trajectory Optimization under Unknown Dynamics}
\label{sec:gpstraj}

Since the Lagrangian $\lagtraj(\trajdist,\params)$ in the previous section factorizes over the mixture elements in $\trajdist(\traj) = \sum_i \trajdist_i(\traj)$, we describe the trajectory optimization method for a single Gaussian $\trajdist(\traj)$. When there are multiple mixture elements, this procedure is applied in parallel to each $\trajdist_i(\traj)$. Since $\trajdist(\traj)$ is Gaussian, the conditionals $\trajdist(\state_{t+1}|\st,\at)$ and $\trajdist(\at|\st)$, which correspond to the dynamics and the controller, are time-varying linear-Gaussian, and given by
\[
\trajdist(\at|\st) = \gauss(\Kpol_t\st + \kpol_t, \ucovart) \hspace{0.45in} \trajdist(\state_{t+1}|\st,\at) = \gauss(\fxt\st + \fut\at + \fct, \noise_t).
\]
This type of controller can be learned efficiently with a small number of real-world samples, making it a good choice for optimizing the guiding distributions. Since a different set of time-varying linear-Gaussian dynamics is fitted for each initial state, this dynamics representation can model any continuous deterministic system that can be locally linearized. Stochastic dynamics can violate the local linearity assumption in principle, but we found that in practice this representation was well suited for a wide variety of noisy real-world tasks.

The dynamics are determined by the environment. If they are known, $\trajdist(\at|\st)$ can be optimized with a variant of the iterative linear-quadratic-Gaussian regulator (iLQG) \citep{lt-ilqr-04,lk-gps-13}, which is a variant of DDP \citep{jm-ddp-70}. In the case of unknown dynamics, we can fit $\trajdist(\state_{t+1}|\st,\at)$ to sample trajectories sampled from the trajectory distribution at the previous iteration, denoted $\hat{\trajdist}(\traj)$. If $\hat{\trajdist}(\traj)$ is too different from $\trajdist(\traj)$, these samples will not give a good estimate of $\trajdist(\state_{t+1}|\st,\at)$, and the optimization will diverge. To avoid this, we can bound the change from $\hat{\trajdist}(\traj)$ to $\trajdist(\traj)$ in terms of their KL-divergence by a step size $\epsilon$, producing the following constrained problem:
\[
\min_{\trajdist(\traj)\in\gauss(\traj)} \lagtraj(\trajdist,\params) \text{ s.t. } \kl(\trajdist(\traj)\|\hat{\trajdist}(\traj)) \leq \epsilon.
\]
This type of policy update has previously been proposed by several authors in the context of policy search \citep{bs-cps-03,ps-rlmsp-08,pma-reps-10,la-lnnpg-14}. In the case when $\trajdist(\traj)$ is Gaussian, this problem can be solved efficiently using dual gradient descent, while the dynamics $\trajdist(\state_{t+1}|\st,\at)$ are fitted to samples gathered by running the previous controller $\hat{\trajdist}(\at|\st)$ on the robot. Fitting a global Gaussian mixture model to tuples $(\st,\at,\state_{t+1})$ and using it as a prior for fitting the dynamics $\trajdist(\state_{t+1}|\st,\at)$ serves to greatly reduce the sample complexity. We describe the dynamics fitting procedure in detail in Appendix~\ref{app:dynamics}.

Note that the trajectory optimization cost function $\lagtraj(\trajdist,\params)$ also depends on the policy $\policy_\params(\at|\st)$, while we only have access to $\policy_\params(\at|\ot)$. In order to compute a local quadratic expansion of the KL-divergence term $\kl(\trajdist(\at|\st)\|\policy_\params(\at|\st))$ inside $\lagtraj(\trajdist,\params)$ for iLQG, we also estimate a linearization of the mean of the conditionally Gaussian policy $\policy_\params(\at|\ot)$ with respect to the state $\st$, using the same procedure that we use to linearize the dynamics. The data for this estimation consists of tuples $\{\st^i, E_{\policy_\params(\at|\ot^i)}[\at]\}$, which we can obtain because both the states $\st^i$ and the observations $\ot^i$ are available for all of the samples evaluated on the real physical system.

This constrained optimization is performed in the ``inner loop'' of the optimization described in the previous section, and the KL-divergence constraint $\kl(\trajdist(\traj)\|\hat{\trajdist}(\traj)) \leq \epsilon$ imposes a step size on the trajectory update. The overall algorithm then becomes an instance of generalized BADMM \citep{wb-badmm-14}. Note that the augmented Lagrangian $\lagtraj(\trajdist,\params)$ consists of an expectation under $\trajdist(\traj)$ of a quantity that is independent of $\trajdist$. We can locally approximate this quantity with a quadratic by using a quadratic expansion of $\cost(\st,\at)$, and fitting a linear-Gaussian to $\policy_\params(\at|\st)$ with the same method we used for the dynamics. We can then solve the primal optimization in the dual gradient descent procedure with a standard LQR backward pass. This is significantly simpler and much faster than the forward-backward dynamic programming procedure employed in previous work \citep{la-lnnpg-14,lk-lcnnp-14}. This improvement is enabled by the use of BADMM, which allows us to always formulate the KL-divergence term in the Lagrangian with the distribution being optimized as the first argument. Since the KL-divergence is convex in its first argument, this makes the corresponding optimization significantly easier. The details of this LQR-based dual gradient descent algorithm are derived in Appendix~\ref{app:trajopt}.

We can further improve the efficiency of the method by allowing samples from multiple trajectories $\trajdist_i(\traj)$ to be used to fit a shared dynamics $\trajdist(\state_{t+1}|\st,\at)$, while the controllers $\trajdist_i(\at|\st)$ are allowed to vary. This makes sense when the initial states of these trajectories are similar, and they therefore visit similar regions. This allows us to draw just a single sample from each $\trajdist_i(\traj)$ at each iteration, allowing us to handle many more initial states.

\subsection{Supervised Policy Optimization}
\label{sec:gpspolicy}

Since the policy parameters $\params$ participate only in the constraints of the optimization problem in Equation~(\ref{eqn:cgps}), optimizing the policy corresponds to minimizing the KL-divergence between the policy and trajectory distribution, as well as the expectation of $\lgmut\tr\at$. For a conditional Gaussian policy of the form $\policy_\params(\at|\ot) = \gauss(\polmu(\ot),\polsig(\ot))$, the objective is
\begin{align*}
&\lagpol(\params,\trajdist) \!=\!\! \frac{1}{2N} \!\sum_{i=1}^N \!\sum_{t=1}^T \! E_{\trajdist_i(\st,\ot)} \! \left[ \trace[\ucovar_{ti}\inv\polsig(\ot)] \!-\! \log|\polsig(\ot)| \right. \nonumber \\
& \left. + (\polmu(\ot) \!-\! \trajmu_{ti}(\st))\ucovar_{ti}\inv(\polmu(\ot) \!-\! \trajmu_{ti}(\st)) + 2\lgmut\tr\polmu(\ot) \right], 
\end{align*}
\noindent where $\trajmu_{ti}(\st)$ is the mean of $\trajdist_i(\at|\st)$ and $\ucovar_{ti}$ is the covariance, and the expectation is evaluated using samples from each $\trajdist_i(\traj)$ with corresponding observations $\ot$. The observations are sampled from $p(\ot|\st)$ by recording camera images on the real system. Since the input to $\polmu(\ot)$ and $\polsig(\ot)$ is not the state $\st$, but only an observation $\ot$, we can train the policy to directly use raw observations. Note that $\lagpol(\params,\trajdist)$ is simply a weighted quadratic loss on the difference between the policy mean and the mean action of the trajectory distribution, offset by the Lagrange multiplier. The weighting is the precision matrix of the conditional in the trajectory distribution, which is equal to the curvature of its cost-to-go function \citep{lk-gps-13}. This has an intuitive interpretation: $\lagpol(\params,\trajdist)$ penalizes deviation from the trajectory distribution, with a penalty that is locally proportional to its cost-to-go. At convergence, when the policy $\policy_\params(\at|\ot)$ takes the same actions as $\trajdist_i(\at|\st)$, their Q-functions are equal, and the supervised policy objective becomes equivalent to the policy iteration objective \citep{lk-lcnnp-14}

In this work, we optimize $\lagpol(\params,\trajdist)$ with respect to $\params$ using stochastic gradient descent (SGD), a standard method for neural network training. The covariance of the Gaussian policy does not depend on the observation in our implementation, though adding this dependence would be straightforward. Since training complex neural networks requires a substantial number of samples, we found it beneficial to include sampled observations from previous iterations into the policy optimization, evaluating the action $\trajmu_{ti}(\st)$ at their corresponding states using the current trajectory distributions. Since these samples come from the wrong state distribution, we use importance sampling and weight them according to the ratio of their probability under the current distribution $\trajdist(\st)$ and the one they were sampled from, which is straightforward to evaluate under the estimated linear-Gaussian dynamics \citep{lk-vpsto-13}.

\subsection{Comparison with Prior Guided Policy Search Methods}
\label{sec:gpsprior}

We presented a guided policy search method where the policy is trained on observations, while the trajectories are trained on the full state. The BADMM formulation of guided policy search is new to this work, though several prior guided policy search methods based on constrained optimization have been proposed. \citet{lk-lcnnp-14} proposed a formulation similar to Equation~(\ref{eqn:cgps}), but with a constraint on the KL-divergence between $\trajdist(\traj)$ and $\policy_\params$. This results in a more complex, non-convex forward-backward trajectory optimization phase. Since the BADMM formulation solves a convex problem during the trajectory optimization phase, it is substantially faster and easier to implement and use, especially when the number of trajectories $\trajdist_i(\traj)$ is large.

The use of ADMM for guided policy search was also proposed by \citet{mt-cbfat-14} for deterministic policies under known dynamics. This approach requires known, deterministic dynamics and trains deterministic policies. Furthermore, because this approach uses a simple quadratic augmented Lagrangian term, it further requires penalty terms on the gradient of the policy to account for local feedback. Our approach enforces this feedback behavior due to the higher moments included in the KL-divergence term, but does not require computing the second derivative of the policy.

\section{End-to-End Visuomotor Policies}
\label{sec:policy}

Guided policy search allows us to optimize complex, high-dimensional policies with raw observations, such as when the input to the policy consists of images from a robot's onboard camera. However, leveraging this capability to directly learn policies for visuomotor control requires designing a policy representation that is both data-efficient and capable of learning complex control strategies directly from raw visual inputs. In this section, we describe a deep convolutional neural network (CNN) model that is uniquely suited to this task. Our approach combines a novel spatial soft-argmax layer with a pretraining procedure that provides for flexibility and data-efficiency.

\subsection{Visuomotor Policy Architecture}
\label{sec:policyarch}

Our visuomotor policy runs at 20 Hz on the robot, mapping monocular RGB images and the robot configurations to joint torques on a 7 DoF arm. The configuration includes the angles of the joints and the pose of the end-effector (defined by 3 points in the space of the end-effector), as well as their velocities, but does not include the position of the target object or goal, which must be determined from the image. CNNs often use pooling to discard the locational information that is necessary to determine positions, since it is an irrelevant distractor for tasks such as object classification \citep{lgrn-cdbn-09}. Because locational information is important for control, our policy does not use pooling. Additionally, CNNs built for spatial tasks such as human pose estimation often also rely on the availability of location labels in image-space, such as hand-labeled keypoints~\citep{tjlb-jcngm-14}. We propose a novel CNN architecture capable of estimating spatial information from an image without direct supervision in image space. Our pose estimation experiments, discussed in Section~\ref{sec:training}, show that this network can learn useful visual features using only 3D position information provided by the robot, and no camera calibration. Further training the network with guided policy search to directly output motor torques causes it to acquire \emph{task-specific} visual features. Our experiments in Section~\ref{sec:generalization} show that this improves performance beyond the level achieved with features trained only for pose estimation.

Our network architecture is shown in Figure~\ref{fig:nn}. The visual processing layers of the network consist of three convolutional layers, each of which learns a bank of filters that are applied to patches centered on every pixel of its input. These filters form a hierarchy of local image features. Each convolutional layer is followed by a rectifying nonlinearity of the form $\responsepix=\max(0,z_{cij})$ for each channel $\channel$ and each pixel coordinate $(i,j)$. The third convolutional layer contains $32$ response maps with resolution $109 \times 109$. These response maps are passed through a spatial softmax function of the form $\softmaxpix = {e^{\responsepix}}/{\sum_{i' j'}{e^{\responsepixprime}}}$.
Each output channel of the softmax is a probability distribution over the location of a feature in the image. To convert from this distribution to a coordinate representation $(f_{cx},f_{cy})$, the network calculates the expected image position of each feature, yielding a 2D coordinate for each channel: $f_{cx} = \sum_{ij} \softmaxpix x_{ij}$ and $f_{cy} = \sum_{ij} \softmaxpix y_{ij}$, where $(x_{ij},y_{ij})$ is the image-space position of the point $(i,j)$ in the response map. Since this is a linear operation, it corresponds to a fixed, sparse fully connected layer with weights $W_{cix} = x_{ij}$ and $W_{cjy} = y_{ij}$. The combination of the spatial softmax and expectation operator implement a kind of soft-argmax. The spatial feature points $(f_{cx},f_{cy})$ are concatenated with the robot's configuration and fed into two fully connected layers, each with 40 rectified units, followed by linear connections to the torques. The full network contains about 92,000 parameters, of which 86,000 are in the convolutional layers.

The spatial softmax and the expected position computation serve to convert pixel-wise representations in the convolutional layers to spatial coordinate representations, which can be manipulated by the fully connected layers into 3D positions or motor torques. The softmax also provides lateral inhibition, which suppresses low, erroneous activations, only keeping strong activations that are more likely to be accurate. This makes our policy more robust to distractors, providing generalization to novel visual variation. We compare our architecture with more standard alternatives in Section~\ref{sec:poseeval} and evaluate robustness to visual distractors in Section~\ref{sec:generalization}. However, the proposed architecture is also in some sense more specialized for visuomotor control, in contrast to more general standard convolutional networks. For example, not all perception tasks require information that can be coherently summarized by a set of spatial locations.

\subsection{Visuomotor Policy Training}
\label{sec:training}

\begin{wrapfigure}{r}{.28\columnwidth}
\vspace{-0.12in}
\hspace{-0.000\columnwidth}\includegraphics[width=0.28\columnwidth]{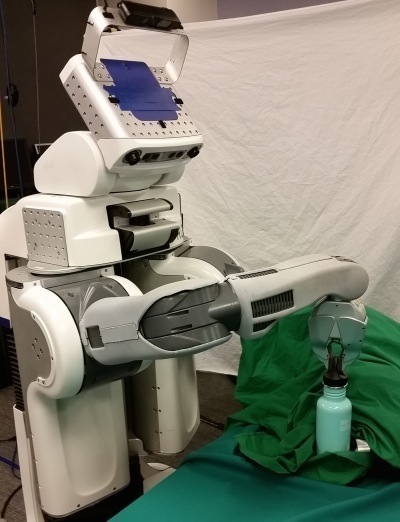}
\put(-45.00,110.00){\includegraphics[width=0.09\columnwidth,frame]{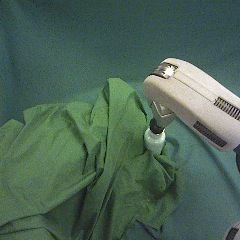}}
\vspace{-0.12in}
\end{wrapfigure}
The guided policy search trajectory optimization phase uses the full state of the system, though the final policy only uses the observations. This type of instrumented training is a natural choice for many robotics tasks, where the robot is trained under controlled conditions, but must then act intelligently in uncontrolled, real-world situations. In our tasks, the unobserved variables are the pose of a target object (e.g. the bottle on which a cap must be placed). During training, this target object is typically held in the robot's left gripper, while the robot's right arm performs the task, as shown to the right. This allows the robot to move the target through a range of known positions. The final visuomotor policy does not receive this position as input, but must instead use the camera images. Due to the modest amount of training data, distractors that are correlated with task-relevant variables can hamper generalization. For this reason, the left arm is covered with cloth to prevent the policy from associating its appearance with the object's position.

While we can train the visuomotor policy entirely from scratch, the algorithm would spend a large number of iterations learning basic visual features and arm motions that can more efficiently be learned by themselves, before being incorporated into the policy. To speed up learning, we initialize both the vision layers in the policy and the trajectory distributions for guided policy search by leveraging the fully observed training setup. To initialize the vision layers, the robot moves the target object through a range of random positions, recording camera images and the object's pose, which is computed automatically from the pose of the gripper. This dataset is used to train a pose regression CNN, which consists of the same vision layers as the policy, followed by a fully connected layer that outputs the 3D points that define the target. Since the training set is still small (we use 1000 images collected from random arm motions), we initialize the filters in the first layer with weights from the model of \citet{sljsr-gdwc-14}, which is trained on ImageNet~\citep{ddsll-lshid-09} classification. After training on pose regression, the weights in the convolutional layers are transferred to the policy CNN. This enables the robot to learn the appearance of the objects prior to learning the behavior.

To initialize the linear-Gaussian controllers for each of the initial states, we take 15 iterations of guided policy search without optimizing the visuomotor policy. This allows for much faster training in the early iterations, when the trajectories are not yet successful, and optimizing the full visuomotor policy is unnecessarily time consuming. Since we still want the trajectories to arrive at compatible strategies for each target position, we replace the visuomotor policy during these iterations with a small network that receives the full state, which consisted of two layers with 40 rectified linear hidden units in our experiments. This network serves only to constrain the trajectories and avoid divergent behaviors from emerging for similar initial states, which would make subsequent policy learning difficult.

\begin{wrapfigure}{r}{.28\columnwidth}
\setlength{\unitlength}{0.28\columnwidth}
\begin{picture}(1.05,1.07) \linethickness{0.5pt}

\put(-0.0,-0.27){\includegraphics[width=0.28\columnwidth]{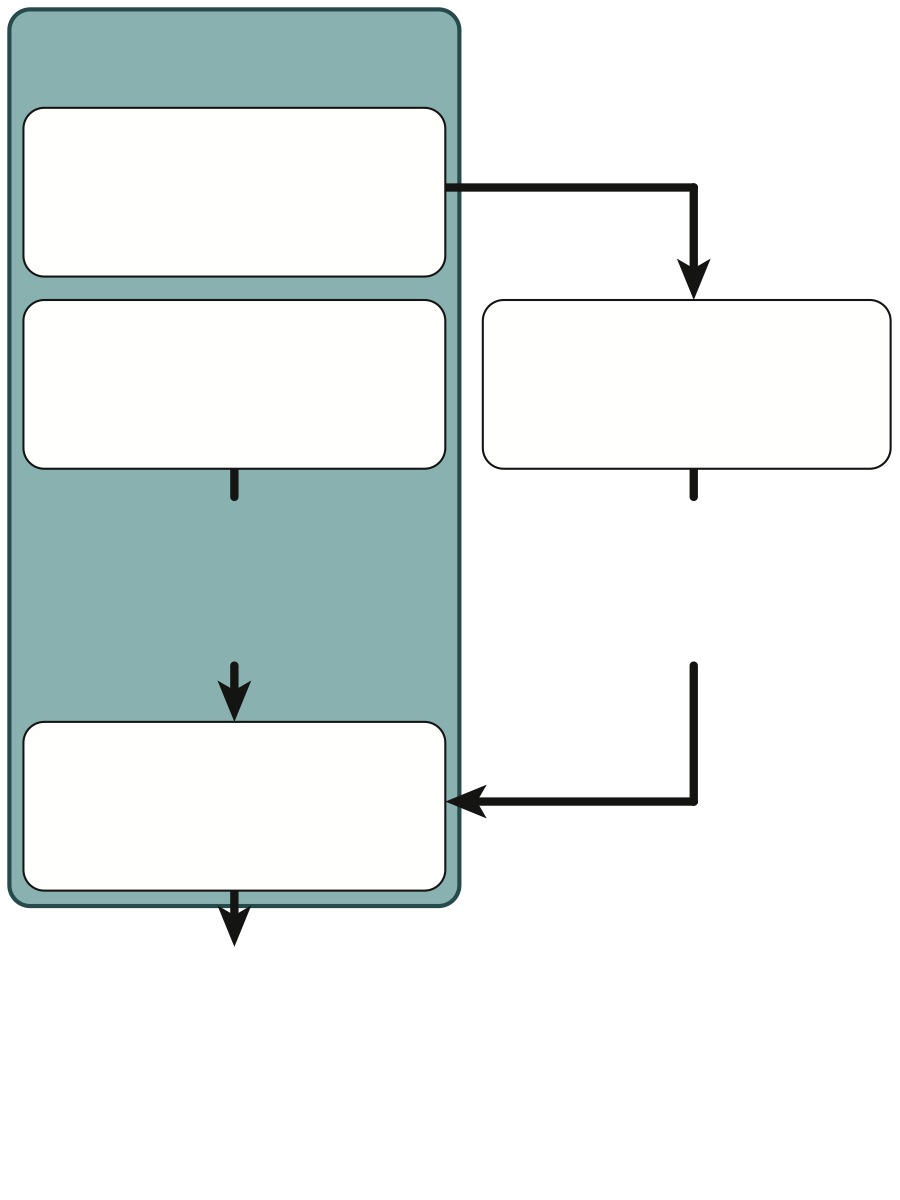}}

\scriptsize{
\put(0.03,0.98){ requires robot}

\put(0.06,0.86){collect visual}
\put(0.1,0.795){pose data}

\put(0.13,0.65){pretrain}
\put(0.085,0.58){trajectories}

\put(0.62,0.65){train pose}
\put(0.68,0.58){CNN}

\put(0.58,0.44){initial visual}
\put(0.65,0.37){features}

\put(0.165,0.44){initial}
\put(0.09,0.37){trajectories}

\put(0.10,0.19){end-to-end}
\put(0.13,0.12){training}

\put(0.175,-0.04){policy}
}

\end{picture}
\vspace{-0.25in}

\end{wrapfigure}
After initialization, we train the full visuomotor policy with guided policy search. During the supervised policy optimization phase, the fully connected motor control layers are first optimized by themselves, since they are not initialized with pretraining. This can be done very quickly because these layers are small. Then, the entire network is further optimized end-to-end. We found that first training the upper layers before end-to-end optimization prevented the convolutional layers from forgetting useful features learning during pretraining, when the error signal due to the untrained upper layers is very large.
The entire pretraining scheme is summarized in the diagram on the right. Note that the trajectories can be pretrained in parallel with the vision layer pretraining, which does not require access to the physical system. Furthermore, the entire initialization procedure does not use any additional information that is not already available from the robot.

\section{Experimental Evaluation}
\label{sec:results}

In this section, we present a series of experiments aimed at evaluating our approach and answering the following questions:
\begin{enumerate}
\vspace{-0.03in}
\item How does the guided policy search algorithm compare to other policy search methods for training complex, high-dimensional policies, such as neural networks?
\vspace{-0.05in}
\item Does our trajectory optimization algorithm work on a real robotic platform with unknown dynamics, for a range of different tasks?
\vspace{-0.05in}
\item How does our spatial softmax architecture compare to other, more standard convolutional neural network architectures?
\vspace{-0.05in}
\item Does training the perception and control systems in a visuomotor policy jointly end-to-end provide better performance than training each component separately?
\vspace{-0.03in}
\end{enumerate}
Evaluating a wide range of policy search algorithms on a real robot would be extremely time consuming, particularly for methods that require a large number of samples. We therefore answer question (1) by using a physical simulator and simpler policies that do not use vision. This also allows us to test the generality of guided policy search on tasks that include manipulation, walking, and swimming. To answer question (2), we present a wide range of experiments on a PR2 robot. These experiments allow us to evaluate the sample efficiency of our trajectory optimization algorithm. To address question (3), we compare a range of different policy architectures on the task of localizing a target object (the cube in the shape sorting cube task). Since localizing the target object is a prerequisite for completing the shape sorting cube task, this serves as a good proxy for evaluating different architectures. Finally, we answer the last and most important question (4) by training visuomotor policies for hanging a coat hanger on a clothes rack, inserting a block into a shape sorting cube, fitting the claw of a toy hammer under a nail with various grasps, and screwing on a bottle cap. These tasks are illustrated in Figure~\ref{fig:tasks}.

\subsection{Simulated Comparisons to Prior Policy Search Methods}
\label{sec:simulated}

In this section, we compare our method against prior policy search techniques on a range of simulated robotic control tasks. These results previously appeared in our conference paper that introduced the trajectory optimization procedure with local linear models \citep{la-lnnpg-14}. In these tasks, the state $\st$ consists of the joint angles and velocities of each robot, and the actions $\at$ consist of the torques at each joint. The neural network policies used one hidden layer and soft rectifier nonlinearities of the form $a = \log(1+\exp(z))$. Since these policies use the state as input, they only have a few hundred parameters, far fewer than our visuomotor policies. However, even this number of parameters can pose a major challenge for prior policy search methods \citep{dnp-spsr-13}.

\paragraph{Experimental tasks.} We simulated 2D and 3D peg insertion, octopus arm control, and planar swimming and walking. The difficulty in the peg insertion tasks stems from the need to align the peg with the slot and the complex contacts between the peg and the walls, which result in discontinuous dynamics. Octopus arm control involves moving the tip of a flexible arm to a goal position \citep{esv-lcoag-05}. The challenge in this task stems from its high dimensionality: the arm has $25$ degrees of freedom, corresponding to $50$ state dimensions. The swimming task requires controlling a three-link snake, and the walking task requires a seven-link biped to maintain a target velocity. The challenge in these tasks comes from underactuation. Details of the simulation and cost for each task are in Appendix~\ref{app:taskssim}.

\paragraph{Prior methods.} We compare to REPS \citep{pma-reps-10}, reward-weighted regression (RWR) \citep{ps-aenac-07,kb-lmpr-09}, the cross-entropy method (CEM) \citep{rk-cem-04}, and PILCO \citep{dr-pmbde-11}. We also use iLQG~\citep{lt-ilqr-04} with a known model as a baseline, shown as a black horizontal line in all plots. REPS is a model-free method that, like our approach, enforces a KL-divergence constraint between the new and old policy. We compare to a variant of REPS that also fits linear dynamics to generate 500 pseudo-samples \citep{lpnp-sbits-14}, which we label ``REPS (20 + 500).'' RWR is an EM algorithm that fits the policy to previous samples weighted by the exponential of their reward, and CEM fits the policy to the best samples in each batch. With Gaussian trajectories, CEM and RWR only differ in the weights. These methods represent a class of RL algorithms that fit the policy to weighted samples, including PoWER and PI2 \citep{kb-lmpr-09,tbs-rlmsh-10,ss-pipic-12}. PILCO is a model-based method that uses a Gaussian process to learn a global dynamics model that is used to optimize the policy. We used the open-source implementation of PILCO provided by the authors. Both REPS and PILCO require solving large nonlinear optimizations at each iteration, while our method does not. Our method used $5$ rollouts with the Gaussian mixture model prior, and $20$ without. Due to its computational cost, PILCO was provided with $5$ rollouts per iteration, while other prior methods used $20$ and $100$. For all prior methods with free hyperparameters (such as the fraction of elites for CEM), we performed hyperparameter sweeps and chose the most successful settings for the comparison.

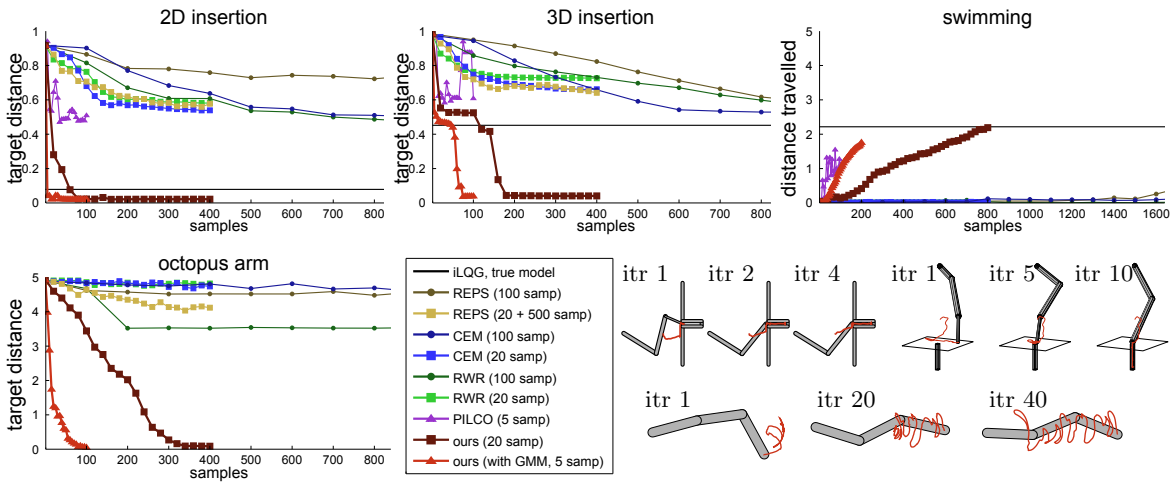
\begin{figure}
\setlength{\unitlength}{0.5\columnwidth}
\begin{picture}(1.99,0.8) \linethickness{0.5pt}

\put(-0.03,-0.03){\includegraphics[height=0.2\columnwidth]{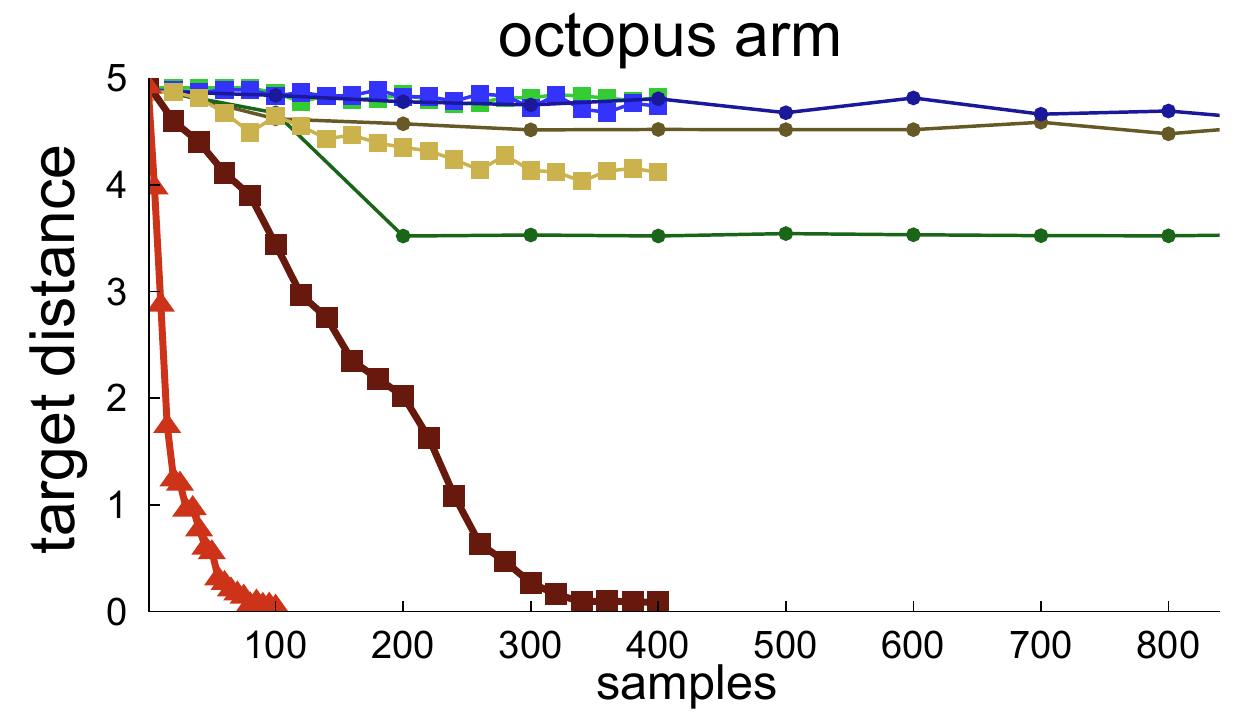}}

\put(-0.03, 0.4){\includegraphics[height=0.2\columnwidth]{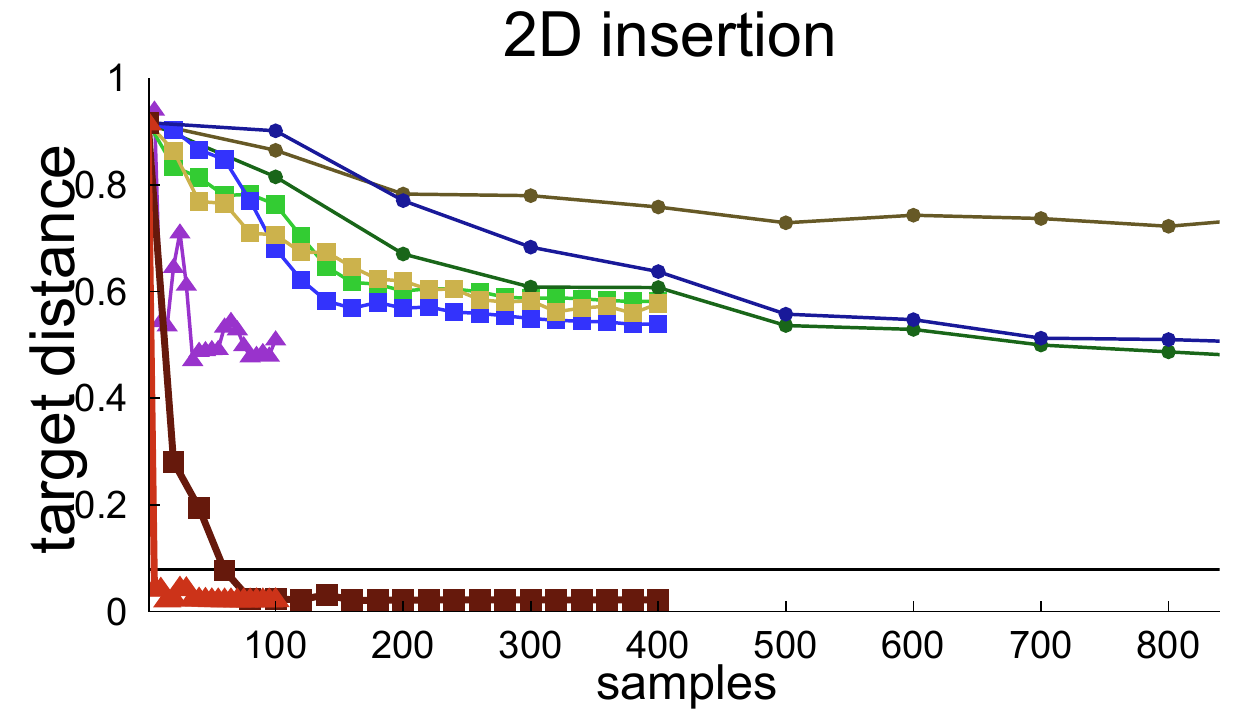}}
\put(0.645, 0.4){\includegraphics[height=0.2\columnwidth]{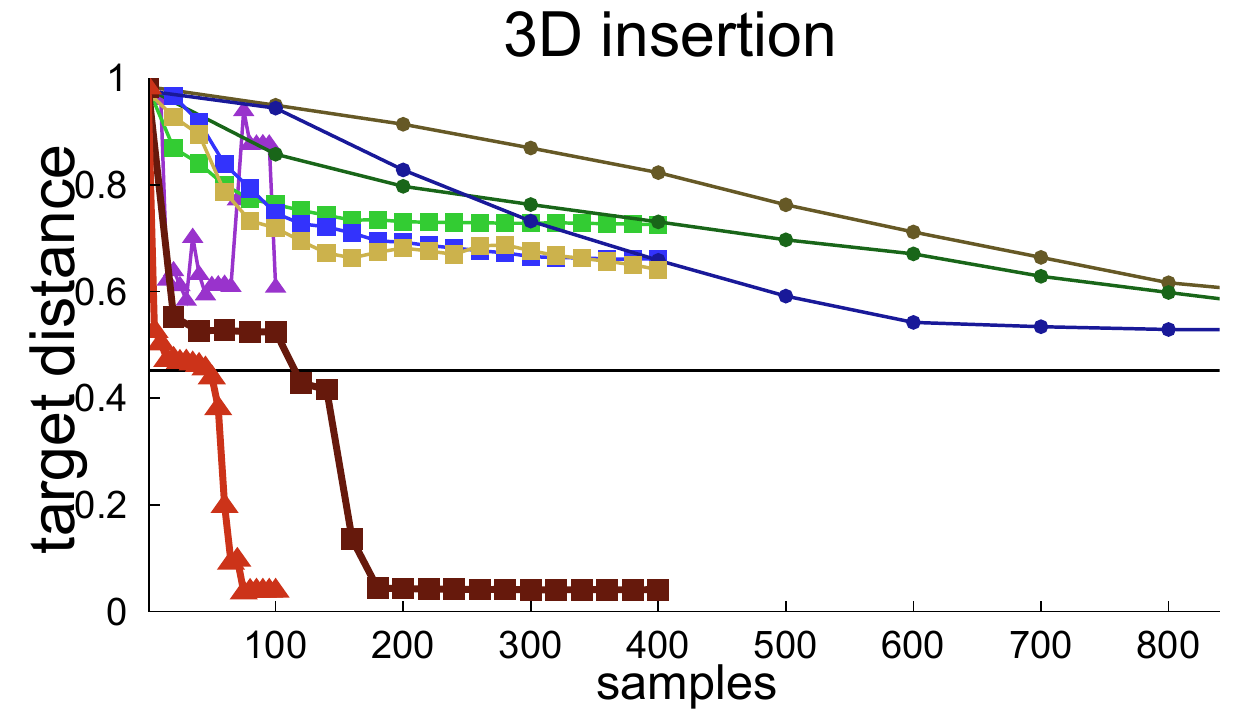}}
\put(1.32, 0.4){\includegraphics[height=0.2\columnwidth]{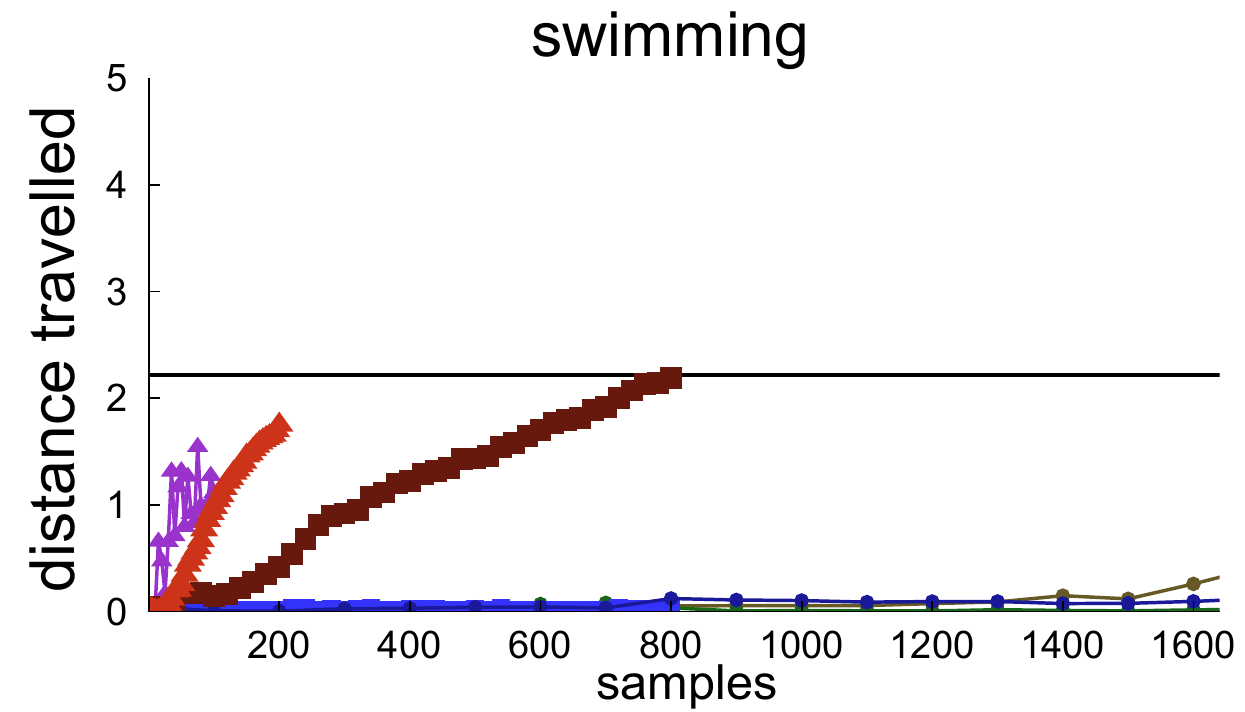}}

\put(0.665,-0.03){\includegraphics[height=0.2\columnwidth]{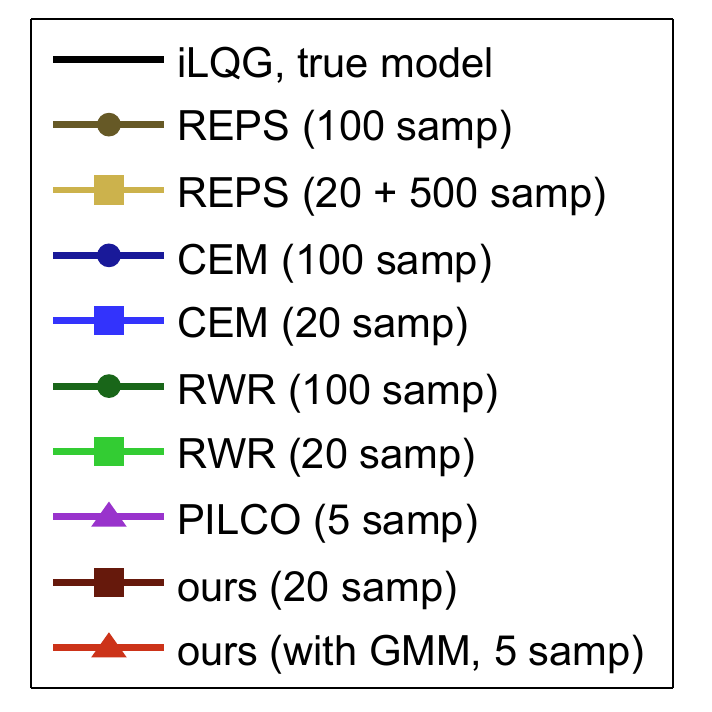}}

\put(1.03, 0.145){\includegraphics[height=0.1\columnwidth]{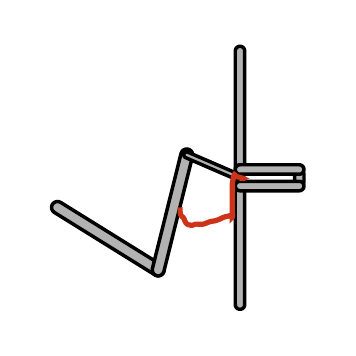}}
\put(1.18, 0.145){\includegraphics[height=0.1\columnwidth]{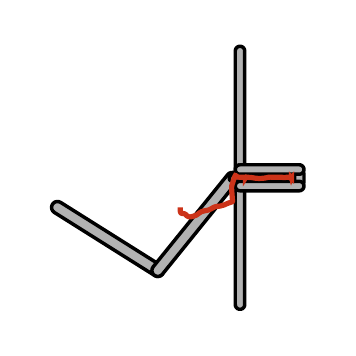}}
\put(1.33, 0.145){\includegraphics[height=0.1\columnwidth]{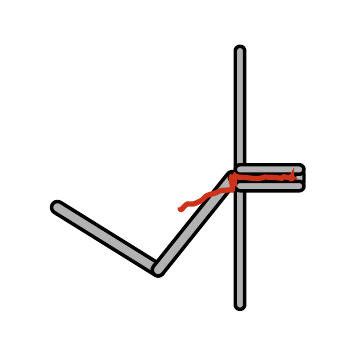}}

\put(1.50, 0.145){\includegraphics[height=0.11\columnwidth]{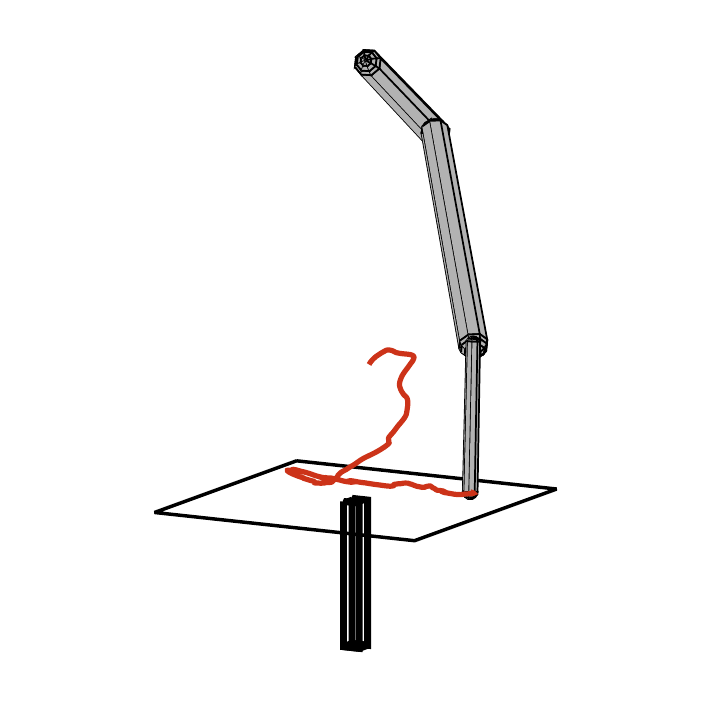}}
\put(1.67, 0.145){\includegraphics[height=0.11\columnwidth]{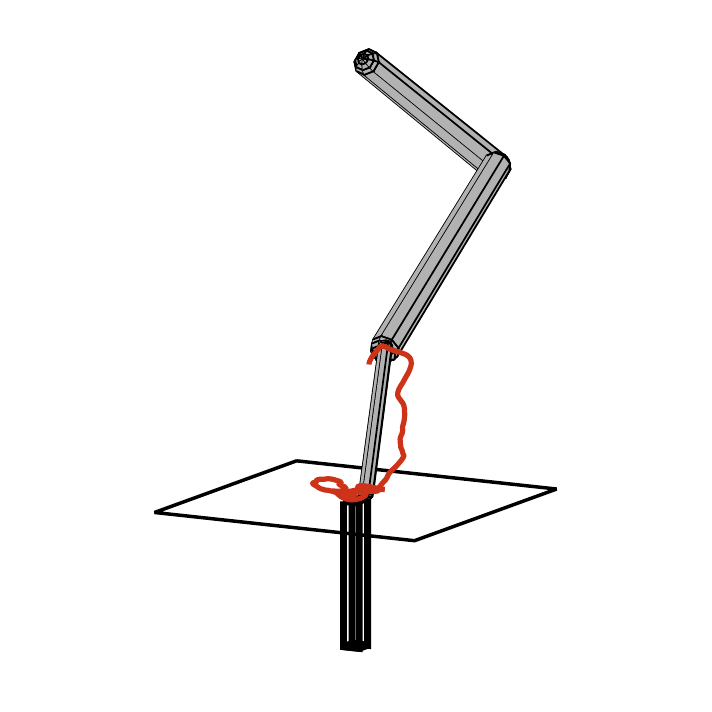}}
\put(1.84, 0.145){\includegraphics[height=0.11\columnwidth]{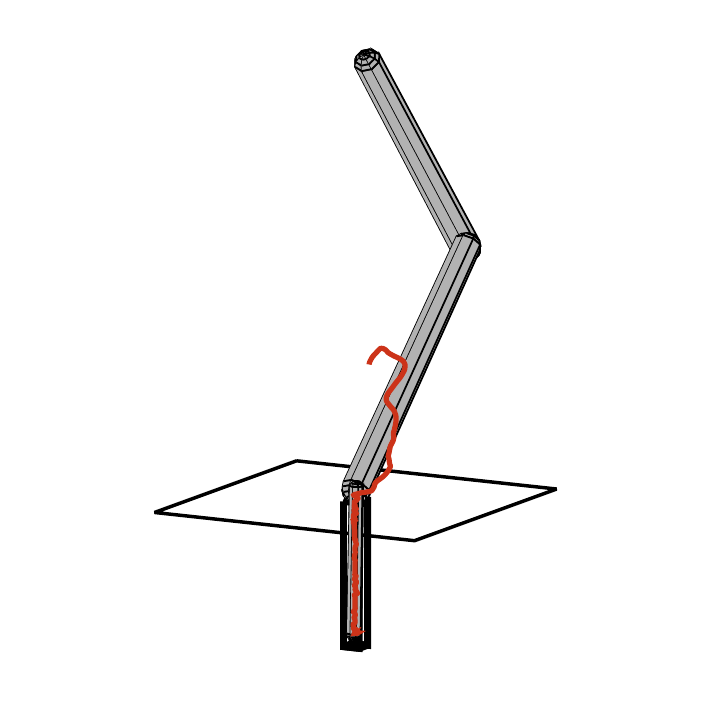}}

\put(1.05,-0.02){\includegraphics[height=0.08\columnwidth]{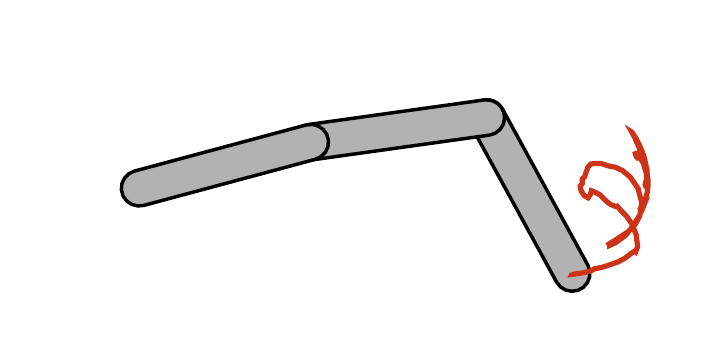}}
\put(1.35,-0.02){\includegraphics[height=0.08\columnwidth]{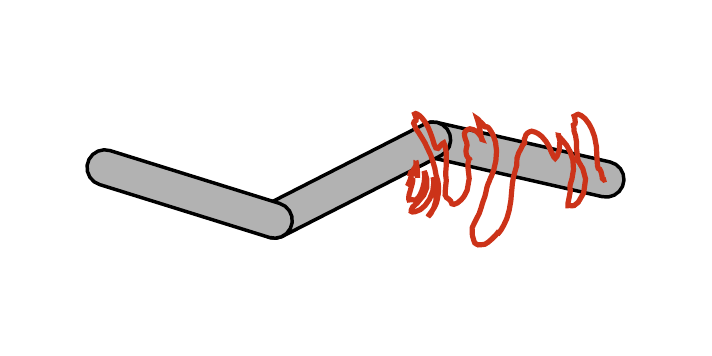}}
\put(1.65,-0.02){\includegraphics[height=0.08\columnwidth]{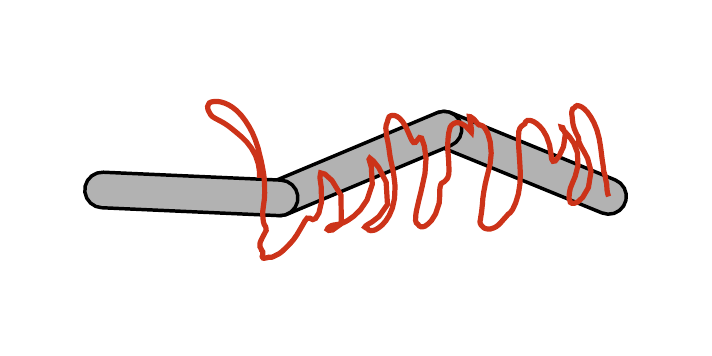}}

{\footnotesize
\put(1.06, 0.32){itr 1}
\put(1.21, 0.32){itr 2}
\put(1.36, 0.32){itr 4}

\put(1.53, 0.32){itr 1}
\put(1.70, 0.32){itr 5}
\put(1.85, 0.32){itr 10}

\put(1.1, 0.1){itr 1}
\put(1.4, 0.1){itr 20}
\put(1.7, 0.1){itr 40}
}

\end{picture}
\vspace{-0.05in}
\caption{Results for learning linear-Gaussian controllers for 2D and 3D insertion, octopus arm, and swimming. Our approach uses fewer samples and finds better solutions than prior methods, and the GMM further reduces the required sample count. Images in the lower-right show the last time step for each system at several iterations of our method, with red lines indicating end effector trajectories.
\label{fig:trajopt}
\vspace{-0.15in}
}
\end{figure}

\paragraph{Gaussian trajectory distributions.} In the first set of comparisons, we evaluate only the trajectory optimization procedure for training linear-Gaussian controllers under unknown dynamics to determine its sample-efficiency and applicability to complex, high-dimensional problems. The results of this comparison for the peg insertion, octopus arm, and swimming tasks appears in Figure~\ref{fig:trajopt}. The horizontal axis shows the total number of samples, and the vertical axis shows the minimum distance between the end of the peg and the bottom of the slot, the distance to the target for the octopus arm, or the total distance travelled by the swimmer. Since the peg is $0.5$ units long, distances above this amount correspond to controllers that cannot perform an insertion. Our method learned much more effective controllers with fewer samples, especially when using the Gaussian mixture model prior. On 3D insertion, it outperformed the iLQG baseline, which used a known model. Contact discontinuities cause problems for derivative-based methods like iLQG, as well as methods like PILCO that learn a smooth global dynamics model. We use a time-varying local model, which preserves more detail, and fitting the model to samples has a smoothing effect that mitigates discontinuity issues. Prior policy search methods could servo to the hole, but were unable to insert the peg. On the octopus arm, our method succeeded despite the high dimensionality of the state and action spaces.\footnote{The high dimensionality of the octopus arm made it difficult to run PILCO, though in principle, such methods should perform well on this task given the arm's smooth dynamics.} Our method also successfully learned a swimming gait, while prior model-free methods could not initiate forward motion. PILCO also learned an effective gait due to the smooth dynamics of this task, but its GP-based optimization required orders of magnitude more computation time than our method, taking about 50 minutes per iteration. In the case of prior model-free methods, the high dimensionality of the time-varying linear-Gaussian controllers likely caused considerable difficulty \citep{dnp-spsr-13}, while our approach exploits the structure of linear-Gaussian controllers for efficient learning.

\begin{figure}
\setlength{\unitlength}{0.5\columnwidth}
\begin{picture}(1.99,0.8) \linethickness{0.5pt}

\put(-0.03,-0.03){\includegraphics[height=0.2\columnwidth]{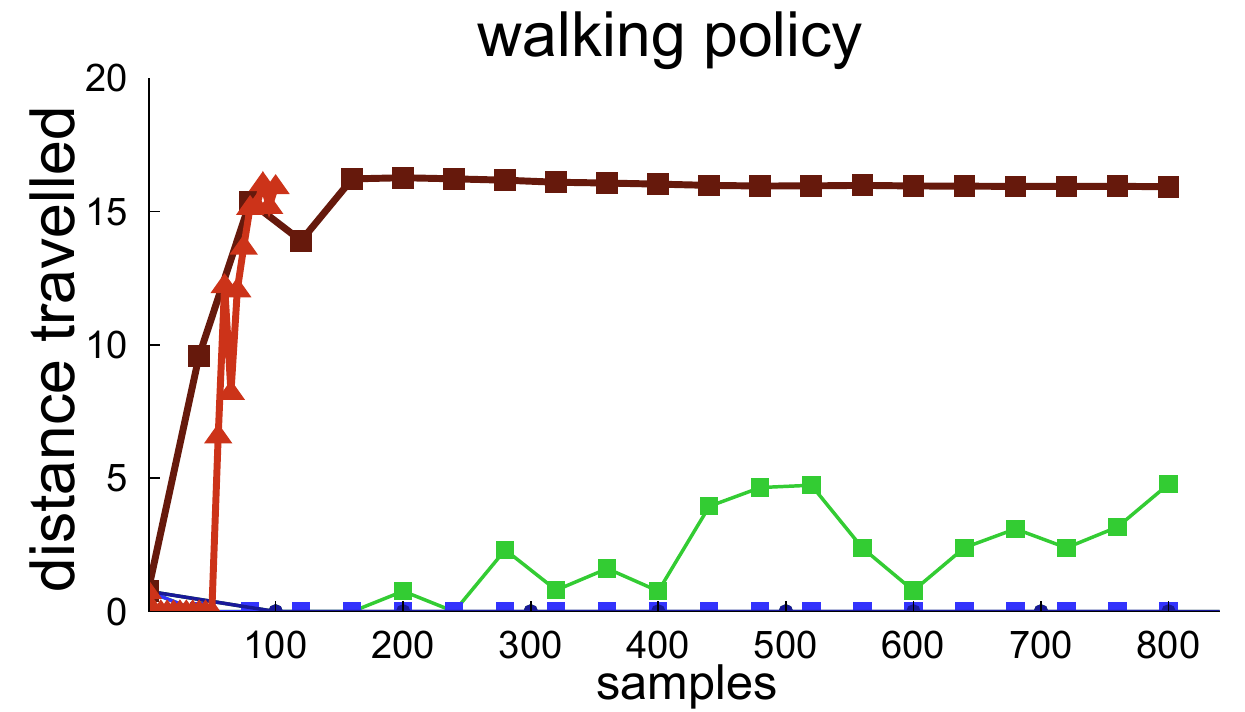}}
\put(-0.03, 0.4){\includegraphics[height=0.2\columnwidth]{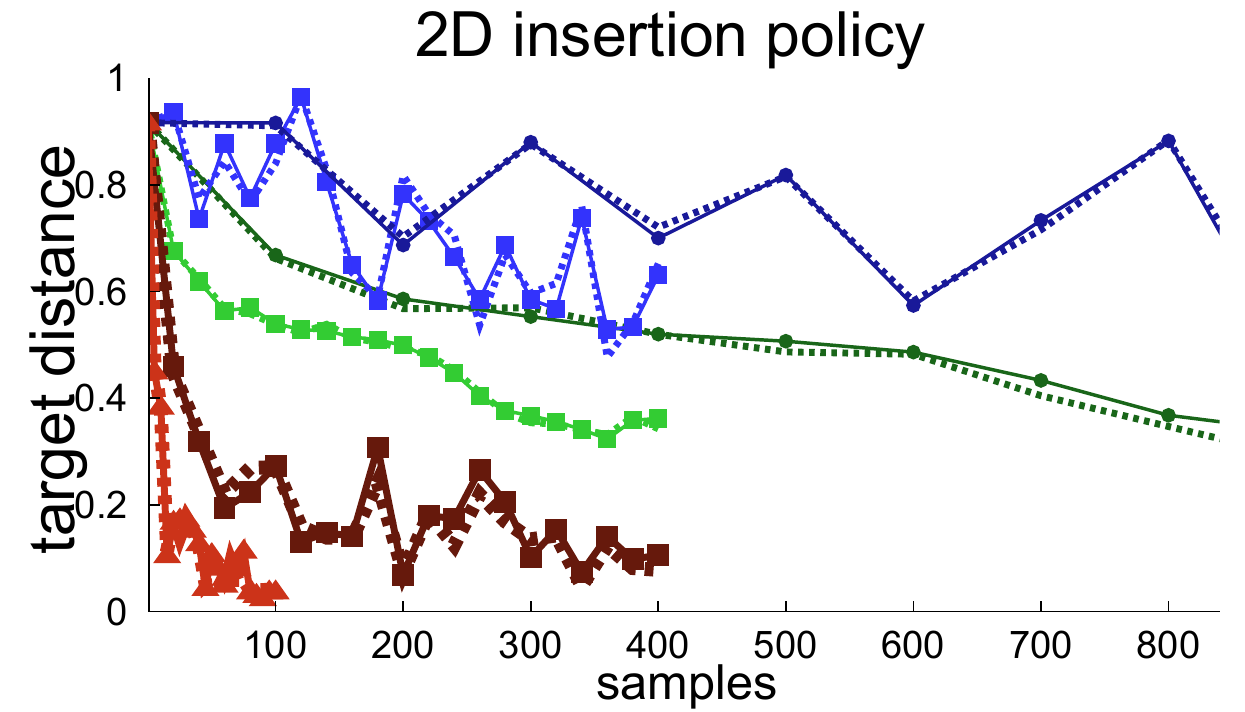}}
\put(0.645, 0.4){\includegraphics[height=0.2\columnwidth]{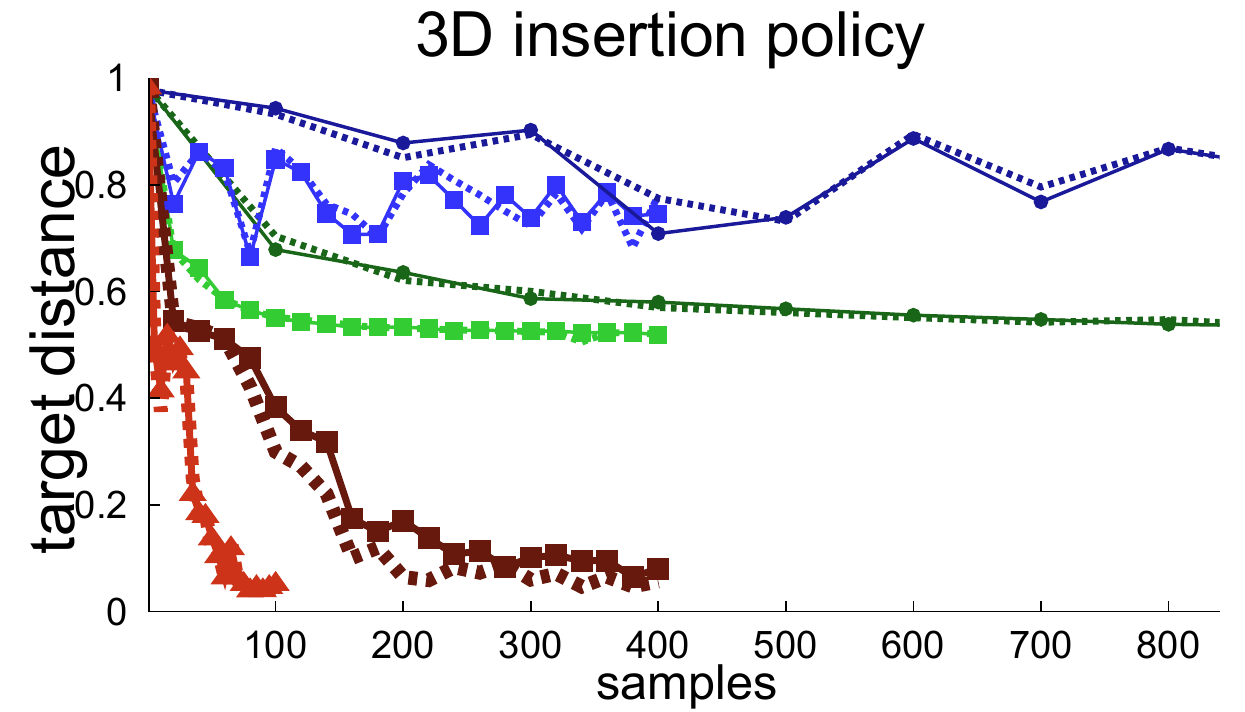}}
\put(1.32, 0.4){\includegraphics[height=0.2\columnwidth]{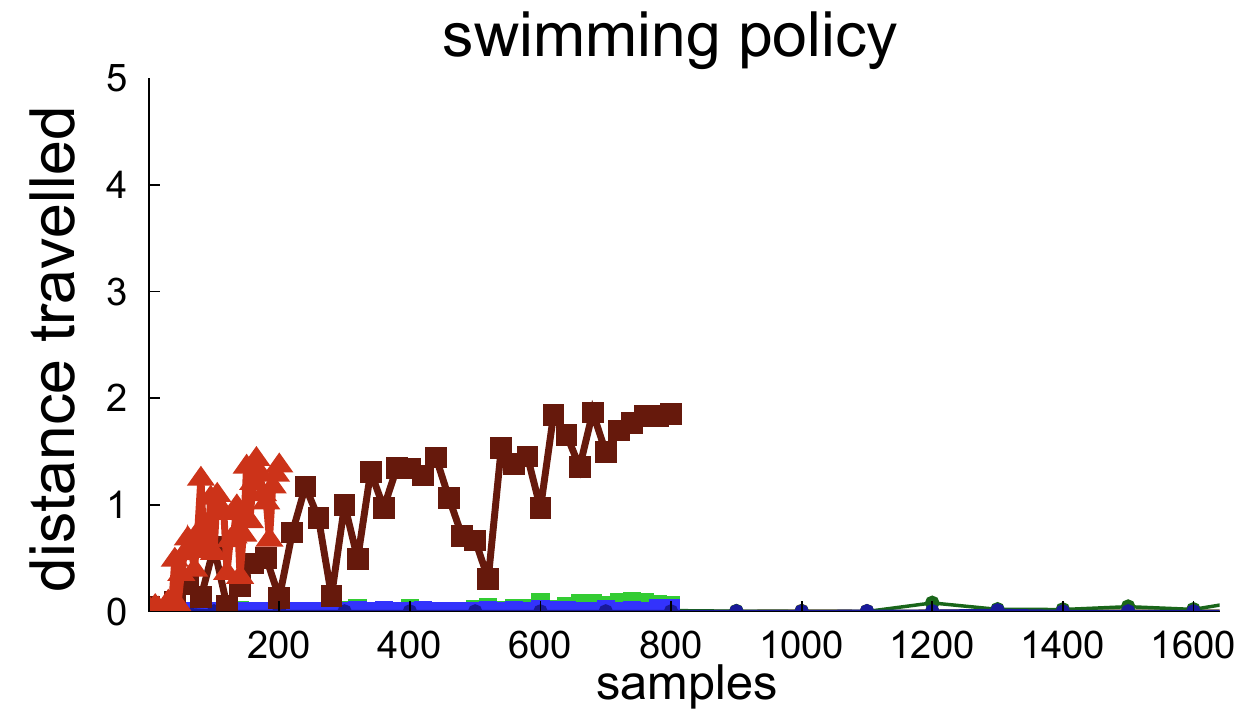}}
\put(0.665,-0.03){\includegraphics[height=0.2\columnwidth]{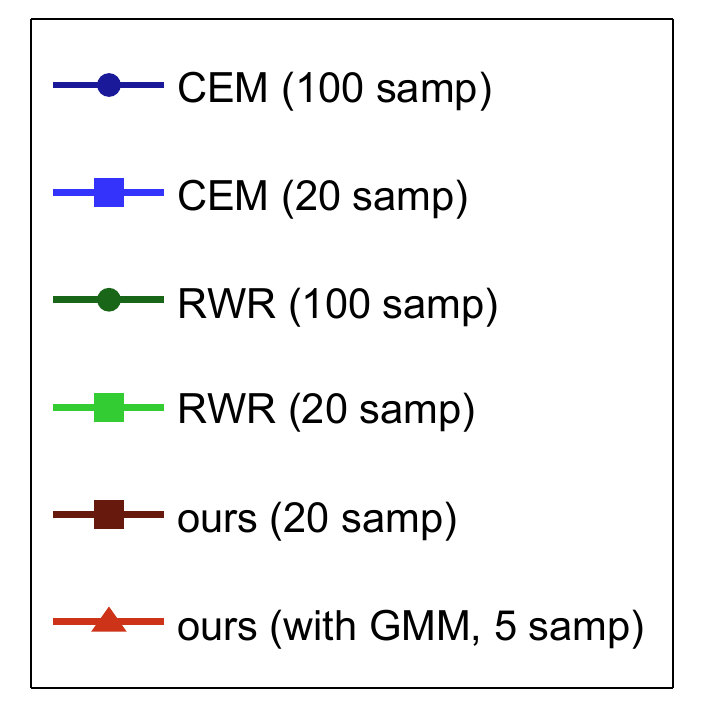}}

\put(1.36, 0.165){\includegraphics[height=0.1\columnwidth]{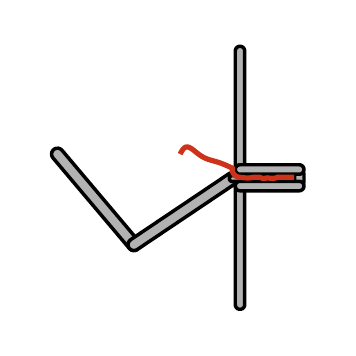}}
\put(1.51, 0.165){\includegraphics[height=0.1\columnwidth]{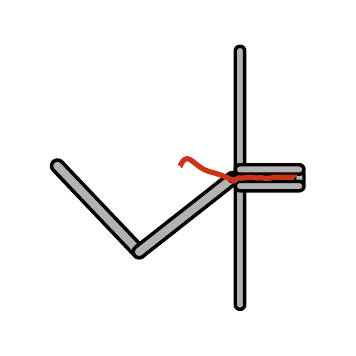}}
\put(1.66, 0.165){\includegraphics[height=0.1\columnwidth]{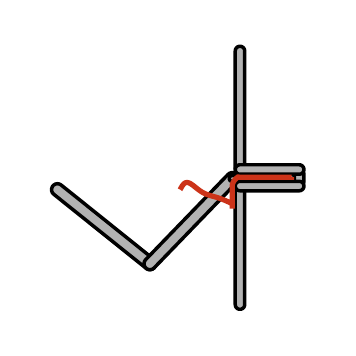}}
\put(1.81, 0.165){\includegraphics[height=0.1\columnwidth]{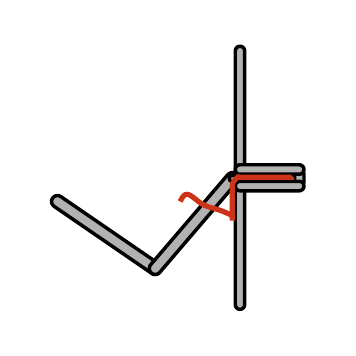}}

\put(1.35,-0.035){\includegraphics[height=0.11\columnwidth]{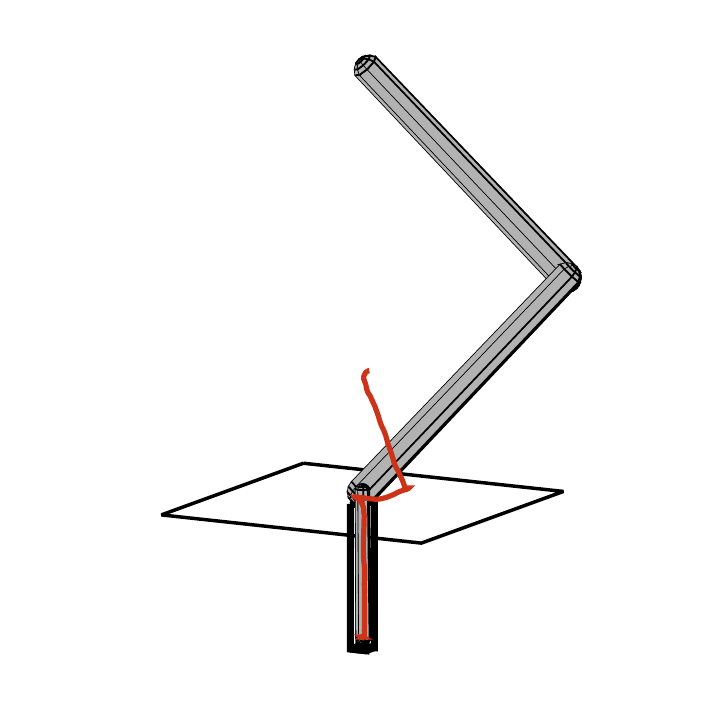}}
\put(1.50,-0.035){\includegraphics[height=0.11\columnwidth]{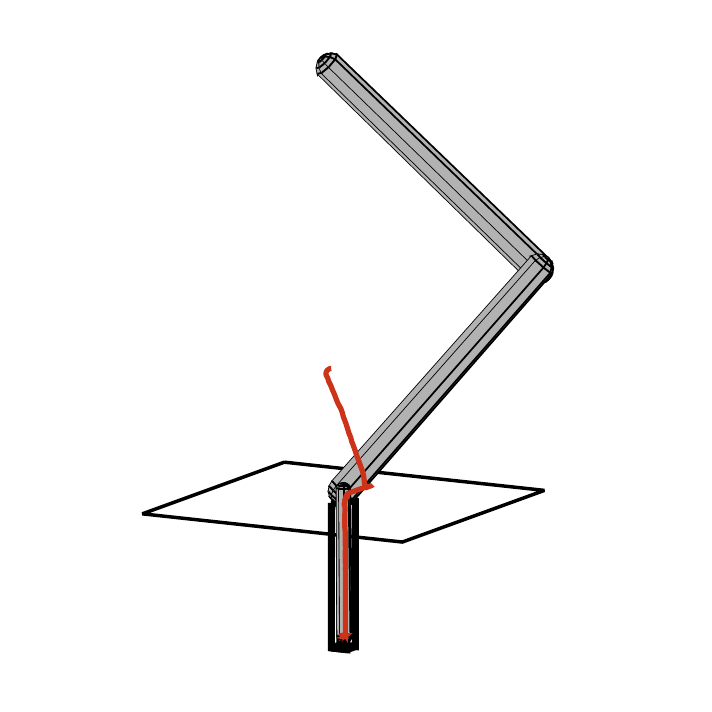}}
\put(1.65,-0.035){\includegraphics[height=0.11\columnwidth]{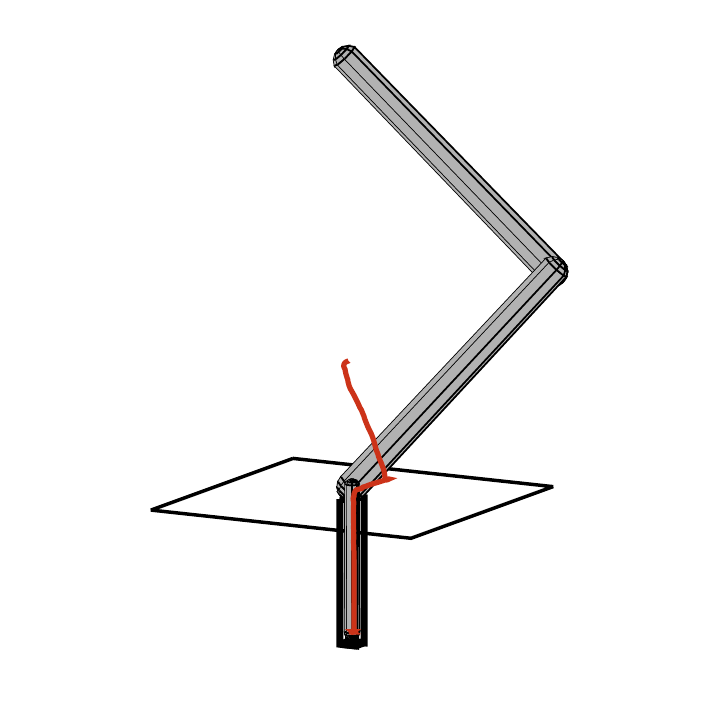}}
\put(1.80,-0.035){\includegraphics[height=0.11\columnwidth]{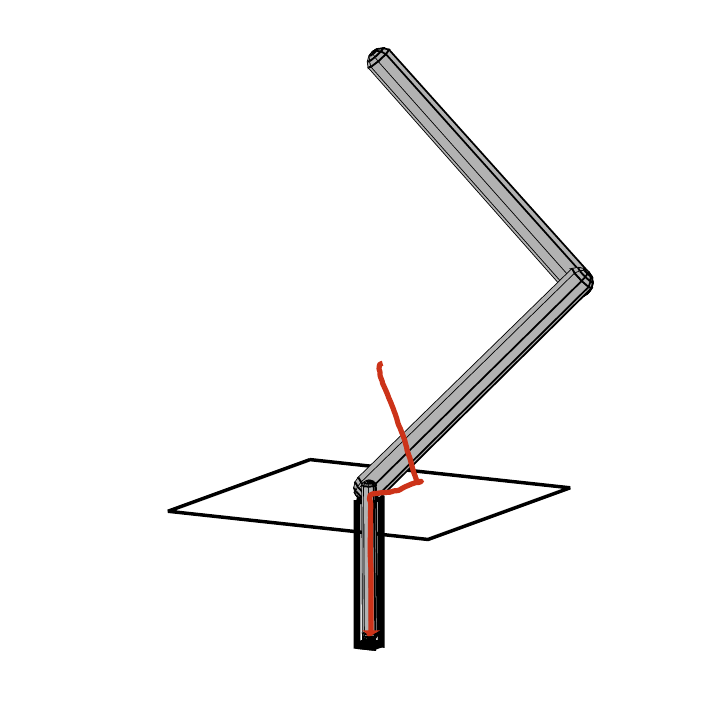}}

\put(1.03, 0.18){\includegraphics[height=0.08\columnwidth]{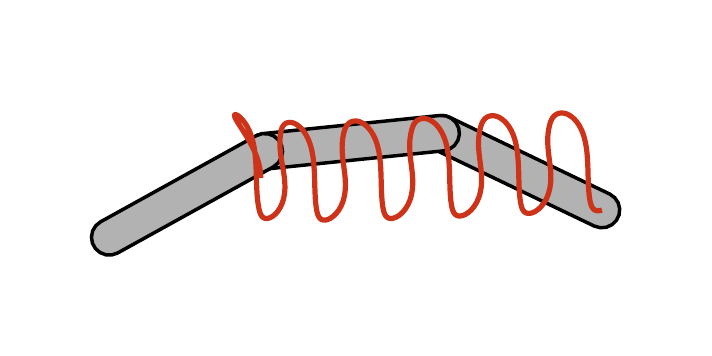}}

\put(1.05,-0.02){\includegraphics[height=0.08\columnwidth]{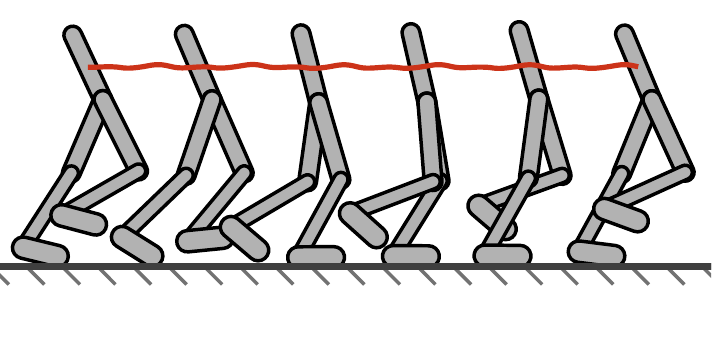}}

{\footnotesize
\put(1.38, 0.32){\#1}
\put(1.53, 0.32){\#2}
\put(1.68, 0.32){\#3}
\put(1.83, 0.32){\#4}

\put(1.38, 0.16){\#1}
\put(1.53, 0.16){\#2}
\put(1.68, 0.16){\#3}
\put(1.83, 0.16){\#4}
}

\end{picture}
\vspace{-0.05in}
\caption{Comparison on neural network policies. For insertion, the policy was trained to search for an unknown slot position on four slot positions (shown above). Generalization to new positions is graphed with dashed lines. Note how the end effector (red) follows the surface to find the slot, and how the swimming gait is smoother due to the stationary policy.
\label{fig:policy}
\vspace{-0.28in}
}
\end{figure}
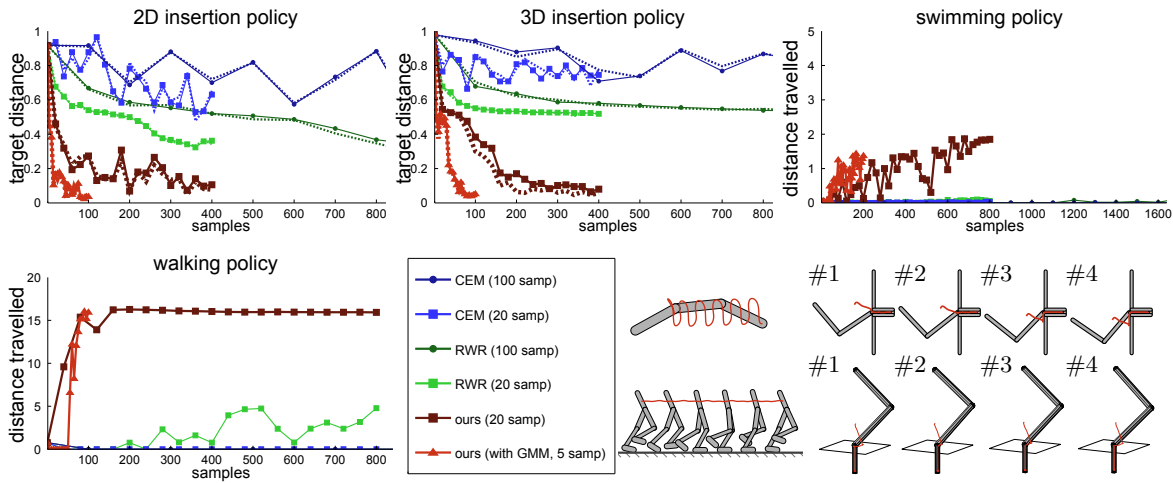

\paragraph{Neural network policies.} In the second set of comparisons, shown in Figure~\ref{fig:policy}, we compare guided policy search to RWR and CEM\footnote{PILCO cannot optimize neural network policies, and we could not obtain reasonable results with REPS. Prior applications of REPS generally focus on simpler, lower-dimensional policy classes \citep{pma-reps-10,lpnp-sbits-14}.} on the challenging task of training high-dimensional neural network policies for the peg insertion and locomotion tasks. The variant of guided policy search used in this comparison differs somewhat from the version described in Section~\ref{sec:gps}, in that it used a simpler dual gradient descent formulation, rather than BADMM. In practice, we found the performance of these methods to be very similar, though the BADMM variant was substantially faster and easier to implement.

On swimming, our method achieved similar performance to the linear-Gaussian case, but since the neural network policy was stationary, the resulting gait was much smoother. Previous methods could only solve this task with $100$ samples per iteration, with RWR eventually obtaining a distance of 0.5m after 4000 samples, and CEM reaching 2.1m after 3000. Our method was able to reach such distances with many fewer samples. Following prior work \citep{lk-gps-13}, the walker trajectory was initialized from a demonstration, which was stabilized with simple linear feedback. The RWR and CEM policies were initialized with samples from this controller to provide a fair comparison. The graph shows the average distance travelled on rollouts that did not fall, and shows that only our method was able to learn walking policies that succeeded consistently.

On peg insertion, the neural network was trained to insert the peg without precise knowledge of the position of the hole, resulting in a partially observed problem. The holes were placed in a region of radius 0.2 units in 2D and 0.1 units in 3D. The policies were trained on four different hole positions, and then tested on four new hole positions to evaluate generalization. The hole position was not provided to the neural network, and the policies therefore had to search for the hole, with only joint angles and velocities as input. Only our method could acquire a successful strategy to locate both the training and test holes, although RWR was eventually able to insert the peg into one of the four holes in 2D.

These comparisons show that training even medium-sized neural network policies for continuous control tasks with a limited number of samples is very difficult for many prior policy search algorithms. Indeed, it is generally known that model-free policy search methods struggle with policies that have over 100 parameters \citep{dnp-spsr-13}. In subsequent sections, we will evaluate our method on real robotic tasks, showing that it can scale from these simulated tasks all the way up to end-to-end learning of visuomotor control.

\subsection{Learning Linear-Gaussian Controllers on a PR2 Robot}
\label{sec:lg}

\begin{figure*}
\setlength{\unitlength}{0.25\columnwidth}
\begin{picture}(1.99,0.75) \linethickness{0.5pt}

\put(0.0,0.37){\includegraphics[width=0.12\columnwidth]{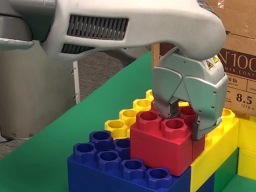}}
\put(0.52,0.37){\includegraphics[width=0.12\columnwidth]{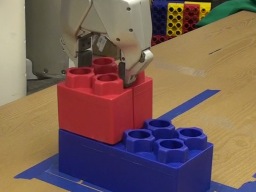}}
\put(0.0,0.0){\includegraphics[width=0.25\columnwidth]{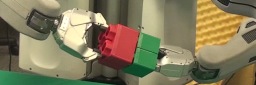}}

\put(1.05,0.0){\includegraphics[height=0.1825\columnwidth]{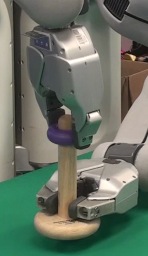}}
\put(1.53,0.0){\includegraphics[height=0.1825\columnwidth]{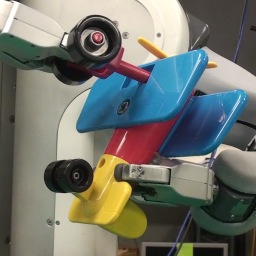}}
\put(2.305,0.0){\includegraphics[height=0.1825\columnwidth]{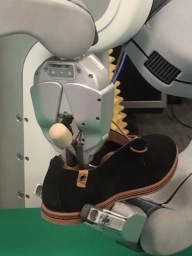}}

\put(2.90,0.0){\includegraphics[height=0.1825\columnwidth]{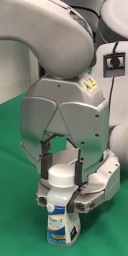}}
\put(3.31,0.0){\includegraphics[height=0.1825\columnwidth]{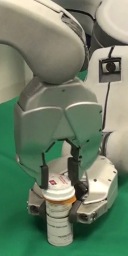}}
\put(3.72,0.0){\includegraphics[height=0.1825\columnwidth]{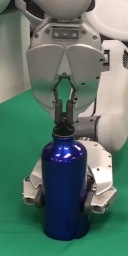}}

\put(0.005,0.39){\textcolor{white}{(a)}}
\put(0.525,0.39){\textcolor{white}{(b)}}
\put(0.005,0.02){\textcolor{white}{(c)}}
\put(1.055,0.02){\textcolor{white}{(d)}}
\put(1.535,0.02){\textcolor{white}{(e)}}
\put(2.31,0.02){\textcolor{white}{(f)}}
\put(2.905,0.02){\textcolor{white}{(g)}}
\put(3.315,0.02){\textcolor{white}{(h)}}
\put(3.725,0.02){\textcolor{white}{(i)}}

\end{picture}
\caption{Tasks for linear-Gaussian controller evaluation: (a) stacking lego blocks on a fixed base, (b) onto a free-standing block, (c) held in both gripper; (d) threading wooden rings onto a peg; (e) attaching the wheels to a toy airplane; (f) inserting a shoe tree into a shoe; (g,h) screwing caps onto pill bottles and (i) onto a water bottle.
\label{fig:objects}
\vspace{-0.2in}
}
\end{figure*}
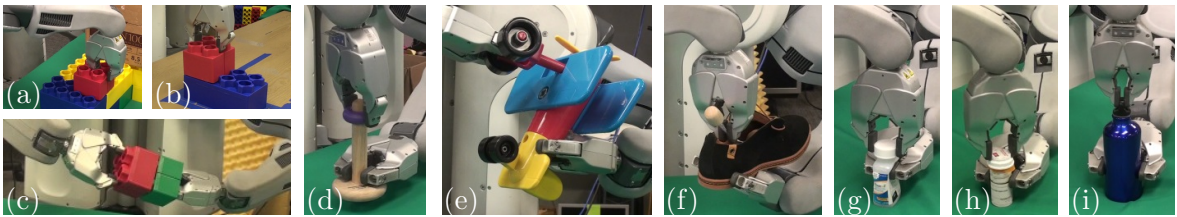
\label{sec:trajresults}

\begin{wrapfigure}{r}{.5\columnwidth}
\setlength{\unitlength}{0.5\columnwidth}
\begin{picture}(1.99,0.46) \linethickness{0.5pt}

\put(0.0,0.0){\includegraphics[width=0.5\columnwidth]{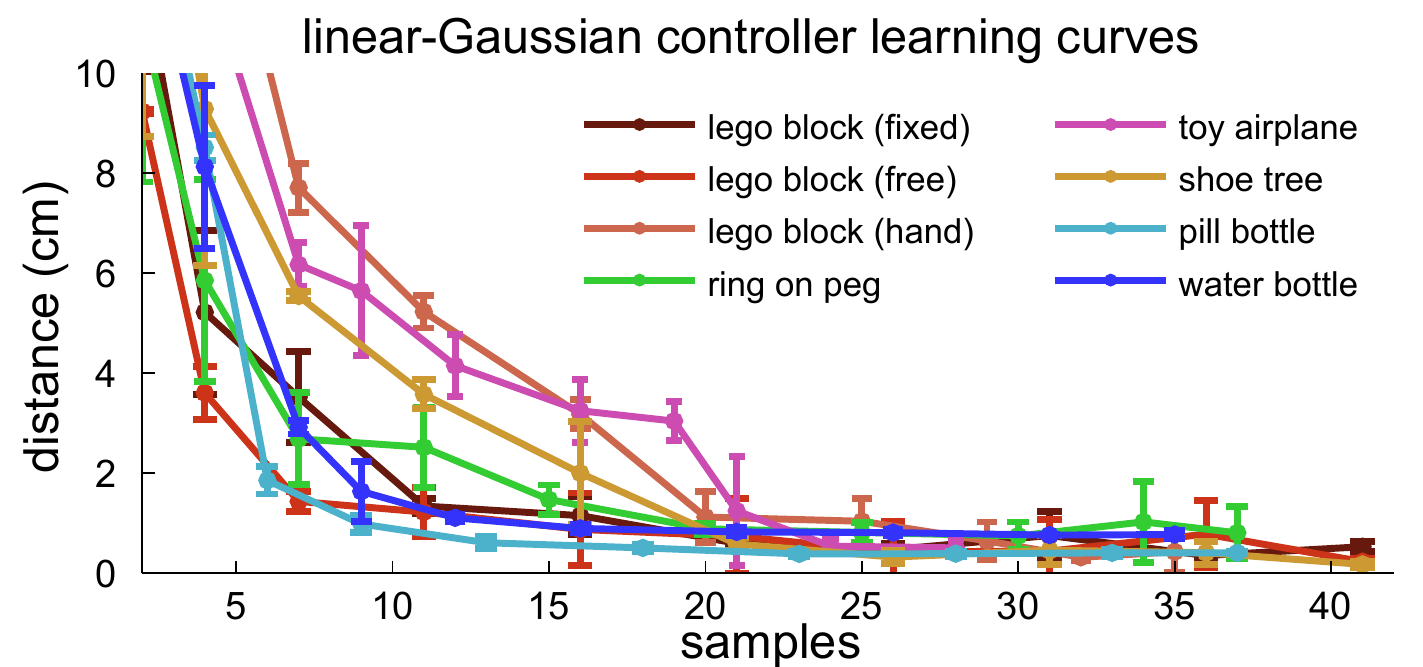}}

\end{picture}
\caption{Distance to target point during training of linear-Gaussian controllers. The actual target may differ due to perturbations. Error bars indicate one standard deviation.
\label{fig:lgresults}
\vspace{-0.15in}
}
\end{wrapfigure}

In this section, we demonstrate the range of manipulation tasks that can be learned using our trajectory optimization algorithm on a real PR2 robot. These experiments previously appeared in our conference paper on guided policy search \citep{lwa-lnnpg-15}. Since performing trajectory optimization is a prerequisite for guided policy search to learn effective visuomotor policies, it is important to evaluate that our trajectory optimization can learn a wide variety of robotic manipulation tasks under unknown dynamics. The tasks in these experiments are shown in Figure~\ref{fig:objects}, while Figure~\ref{fig:lgresults} shows the learning curves for each task. For all robotic experiments in this article, the tasks were learned entirely from scratch,
with the initialization of the controllers $\trajdist(\at|\st)$ described in Appendix~\ref{app:tasksviz}. The number of samples required to learn each controller is around 20-25, substantially lower than many prior policy search methods in the literature \citep{ps-rlmsp-08,kop-rlarm-10,tbs-rlmsh-10,dnp-spsr-13}. Total learning time was about ten minutes for each task, of which only 3-4 minutes involved system interaction. The rest of the time was spent resetting the robot to the initial state and on computation.

The linear-Gaussian controllers are optimized for a specific condition -- e.g., a specific position of the target lego block. To evaluate their robustness to errors in the specified target position, we conducted experiments on the lego block and ring tasks where the target object (the lower block and the peg) was perturbed at each trial during training, and then tested with various perturbations. For each task, controllers were trained with Gaussian perturbations with standard deviations of $0$, $1$, and $2$ cm in the position of the target object, and each controller was tested with perturbations of radius $0$, $1$, $2$, and $3$ cm. Note that with a radius of $2$ cm, the peg would be placed about one ring-width away from the expected position. The results are shown in Table~\ref{tbl:var}. All controllers were robust to perturbations of $1$ cm, and would often succeed at $2$ cm. Robustness increased slightly when more noise was injected during training, but even controllers trained without noise exhibited considerable robustness, since the linear-Gaussian controllers themselves add noise during sampling. We also evaluated a kinematic baseline for each perturbation level, which planned a straight path from a point 5 cm above the target to the expected (unperturbed) target location. This baseline was only able to place the lego block in the absence of perturbations. The rounded top of the peg provided an easier condition for the baseline, with occasional successes at higher perturbation levels. Our controllers outperformed the baseline by a wide margin.

All of the robotic experiments discussed in this section may be viewed in the corresponding supplementary video, available online: \url{http://rll.berkeley.edu/icra2015gps}. A video illustration of the visuomotor policies, discussed in the following sections, is also available: \url{http://sites.google.com/site/visuomotorpolicy}.

\begin{table}[t]
\begin{center}
\footnotesize{
\begin{tabular}{| l | l | l | l | l | l || l | l | l | l |}
\cline{3-10}
\multicolumn{2}{l|}{} & \multicolumn{8}{c|}{test perturbation} \\
\cline{3-10}
\multicolumn{2}{l|}{} & \multicolumn{4}{c||}{lego block} & \multicolumn{4}{c|}{ring on peg} \\
\cline{3-10}
\multicolumn{2}{l|}{} & 0 cm & 1 cm & 2 cm & 3 cm & 0 cm & 1 cm & 2 cm & 3 cm \\
\hline
\multirow{4}{*}{\rotatebox{90}{\parbox[b][0.6cm][t]{1.3cm}{training\\perturb.}}} & 0 cm & 5/5 & 5/5 & 3/5 & 2/5 & 5/5 & 5/5 & 0/5 & 0/5 \\
& 1 cm & 5/5 & 5/5 & 3/5 & 2/5 & 5/5 & 5/5 & 3/5 & 0/5 \\
& 2 cm & 5/5 & 5/5 & 5/5 & 3/5 & 5/5 & 5/5 & 3/5 & 0/5 \\
& kinematic baseline & 5/5 & 0/5 & 0/5 & 0/5 & 5/5 & 3/5 & 0/5 & 0/5 \\
\hline
\end{tabular}
}
\end{center}
\vspace{-0.1in}
\caption{Success rates of linear-Gaussian controllers under target object perturbation.
\label{tbl:var}
}
\vspace{-0.2in}
\end{table}

\subsection{Spatial Softmax CNN Architecture Evaluation}
\label{sec:poseeval}

In this section, we evaluate the neural network architecture that we propose in Section~\ref{sec:policyarch} in comparison to more standard convolutional networks. To isolate the architectures from other confounding factors, we measure their accuracy on the pose estimation pretraining task described in Section~\ref{sec:training}. This is a reasonable proxy for evaluating how well the network can overcome two major challenges in visuomotor learning: the ability to handle relatively small datasets without overfitting, and the capability to learn tasks that are inherently spatial. We compare to a network where the expectation operator after the softmax is replaced with a learned fully connected layer, as is standard in the literature, a network where both the softmax and the expectation operators are replaced with a fully connected layer, and a version of this network that also uses $3\times 3$ max pooling with stride $2$ at the first two layers. These alternative architectures have many more parameters, since the fully connected layer takes the entire bank of response maps from the third convolutional layer as input. Pooling helps to reduce the number of parameters, but not to the same degree as the spatial softmax and expectation operators in our architecture.

The results in Table~\ref{tbl:posebaseline} indicate that using the softmax and expectation operators improves pose estimation accuracy substantially. Our network is able to outperform the more standard architectures because it is forced by the softmax and expectation operators to learn feature points, which provide a concise representation suitable for spatial inference. Since most of the parameters in this architecture are in the convolutional layers, which benefit from extensive weight sharing, overfitting is also greatly reduced. By removing pooling, our network also maintains higher resolution in the convolutional layers, improving spatial accuracy. Although we did attempt to regularize the larger standard architectures with higher weight decay and dropout, we did not observe a significant improvement on this dataset. We also did not extensively optimize the parameters of this network, such as filter size and number of channels, and investigating these design decisions further would be valuable to investigate in future work.

\begin{table}[h!]
  \begin{center}
\footnotesize{
    \begin{tabular}{| l | l | l | }
    \hline
    network architecture & test error (cm)  \\
    \hline
    softmax + feature points ({\bf ours}) &  {\bf 1.30 $\pm$ 0.73} \\
    \hline
    softmax + fully connected layer &  2.59 $\pm$ 1.19 \\
    \hline
    fully connected layer &  4.75 $\pm$ 2.29  \\
    \hline
    max-pooling + fully connected &  3.71 $\pm$ 1.73 \\
    \hline
    \end{tabular}
}
  \end{center}
\vspace{-0.1in}
  \caption{Average pose estimation accuracy and standard deviation with various architectures, measured as average Euclidean error for the three target points in 3D, with ground truth determined by forward kinematics from the left arm.
}
  \label{tbl:posebaseline}
\vspace{-0.05in}
\end{table}

\subsection{Deep Visuomotor Policy Evaluation}
\label{sec:generalization}

In this section, we present an evaluation of our full visuomotor policy training algorithm on a PR2 robot. The aim of this evaluation is to answer the following question: does training the perception and control systems in a visuomotor policy jointly end-to-end provide better performance than training each component separately?

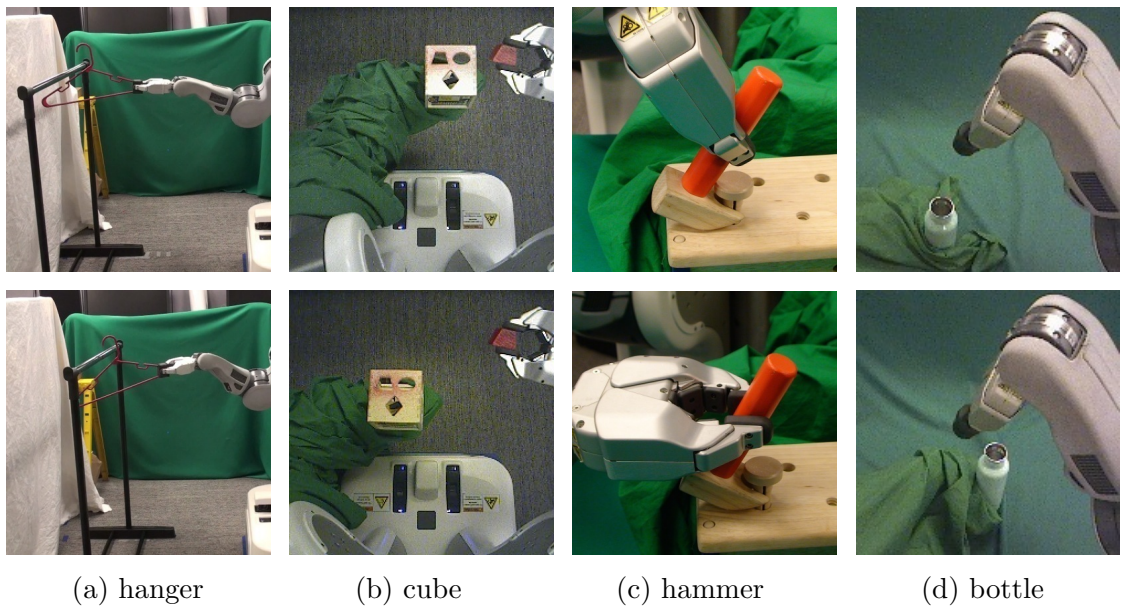
\begin{figure}
\centering
\setlength{\unitlength}{1.9\columnwidth}
\begin{picture}(0.5,0.29) \linethickness{0.5pt}

\put(0.0,0.04){\includegraphics[width=0.23\columnwidth]{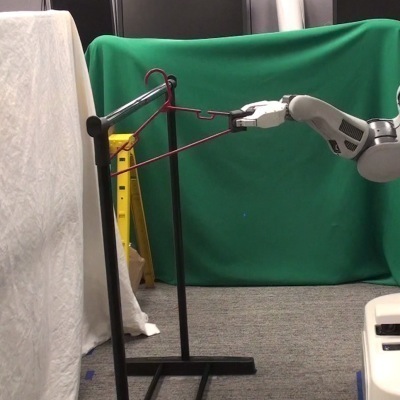}}
\put(0.0,0.17){\includegraphics[width=0.23\columnwidth]{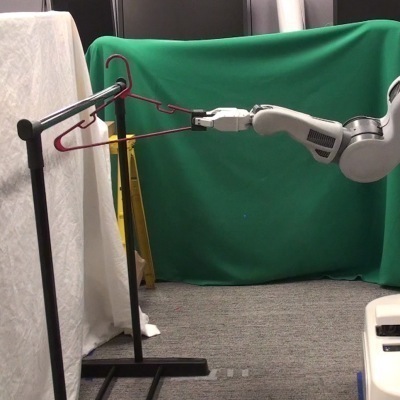}}

\put(0.13,0.04){\includegraphics[width=0.23\columnwidth]{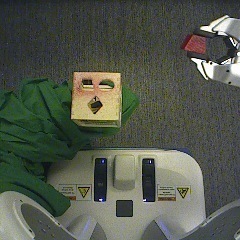}}
\put(0.13,0.17){\includegraphics[width=0.23\columnwidth]{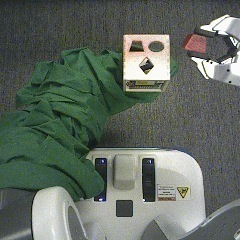}}

\put(0.26,0.04){\includegraphics[width=0.23\columnwidth]{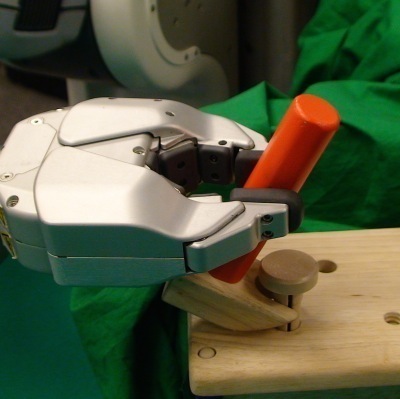}}
\put(0.26,0.17){\includegraphics[width=0.23\columnwidth]{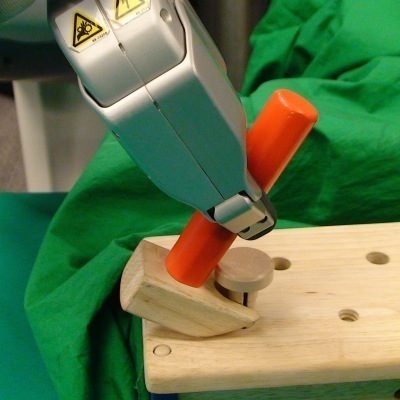}}

\put(0.39,0.04){\includegraphics[width=0.23\columnwidth]{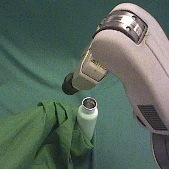}}
\put(0.39,0.17){\includegraphics[width=0.23\columnwidth]{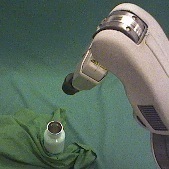}}

\put(0.03,0.02){(a) hanger}
\put(0.16,0.02){(b) cube}
\put(0.28,0.02){(c) hammer}
\put(0.42,0.02){(d) bottle}

\end{picture}
\vspace{-0.23in}
\caption{Illustration of the tasks in our visuomotor policy experiments, showing the variation in the position of the target for the hanger, cube, and bottle tasks, as well as two of the three grasps for the hammer, which also included variation in position (not shown).}
\label{fig:tasks}
\vspace{-0.2in}
\end{figure}

\paragraph{Experimental tasks.} We trained policies for hanging a coat hanger on a clothes rack, inserting a block into a shape sorting cube, fitting the claw of a toy hammer under a nail with various grasps, and screwing on a bottle cap. The cost function for these tasks encourages low distance between three points on the end-effector and corresponding target points, low torques, and, for the bottle task, spinning the wrist. The equations for these cost functions and the details of each task are presented in Appendix~\ref{app:tasksviz}. The tasks are illustrated in Figure~\ref{fig:tasks}. Each task involved variation of 10-20 cm in each direction in the position of the target object (the rack, shape sorting cube, nail, and bottle). In addition, the coat hanger and hammer tasks were trained with two and three grasps, respectively. The current angle of the grasp was not provided to the policy, but had to be inferred from observing the robot's gripper in the camera images. All tasks used the same policy architecture and model parameters.

\begin{figure}
\setlength{\unitlength}{0.5\columnwidth}
\begin{picture}(1.99,1.10) \linethickness{0.5pt}

\put(0.06,0.77){\includegraphics[width=0.14\columnwidth]{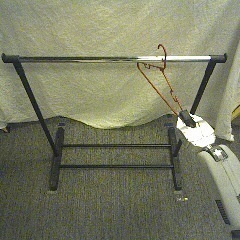}}
\put(0.353,0.77){\includegraphics[width=0.14\columnwidth]{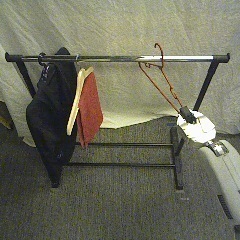}}

\put(0.06,0.48){\includegraphics[width=0.14\columnwidth]{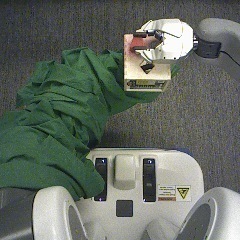}}
\put(0.353,0.48){\includegraphics[width=0.14\columnwidth]{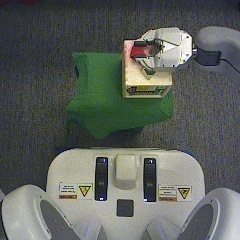}}

\put(0.06,0.19){\includegraphics[width=0.14\columnwidth]{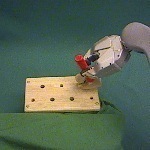}}
\put(0.353,0.19){\includegraphics[width=0.14\columnwidth]{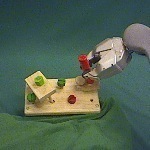}}

\put(0.06,-0.1){\includegraphics[width=0.14\columnwidth]{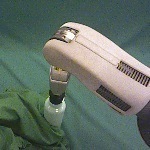}}
\put(0.353,-0.1){\includegraphics[width=0.14\columnwidth]{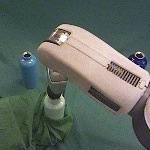}}

\put(0.0,0.82){\rotatebox{90}{hanger}}
\put(0.0,0.56){\rotatebox{90}{cube}}
\put(0.0,0.24){\rotatebox{90}{hammer}}
\put(0.0,-0.02){\rotatebox{90}{bottle}}
\put(0.11,1.07){training}
\put(0.38,1.07){visual test}

\small{
\put(0.67,0.55){
\begin{tabular}{| l | l | l | l |}
    \hline
    \!coat hanger &\!training (18)\!\!&\!\!spatial test (24)\!\!&\!\!visual test (18)\!\!\\
    \hline
    \!end-to-end & {\bf 100\%} & {\bf 100\%} & {\bf 100\%} \\
    \hline
    \!pose features & 88.9\% & 87.5\% & 83.3\% \\
    \hline
    \!pose prediction\!\!& 55.6\% & 58.3\% & 66.7\% \\
    \hline
    \hline
    \!shape cube\!&\!training (27)\!\!&\!\!spatial test (36)\!\!&\!\!visual test (40)\!\!\\
    \hline
    \!end-to-end & {\bf 96.3\%} & {\bf 91.7\%} & {\bf 87.5\%} \\
    \hline
    \!pose features & 70.4\% & 83.3\% & 40\%\\
    \hline
    \!pose prediction\!\!& 0\% & 0\% & n/a\\
    \hline
    \hline
    \!toy hammer\!&\!training (45)\!\!&\!\!spatial test (60)\!\!&\!\!visual test (60)\!\!\\
    \hline
    \!end-to-end & {\bf 91.1\%} & {\bf 86.7\%} & {\bf 78.3\%} \\
    \hline
    \!pose features & 62.2\% & 75.0\% & 53.3\%\\
    \hline
    \!pose prediction\!\!& 8.9\% & 18.3\% & n/a\\
    \hline
    \hline
    \!bottle cap &\!training (27)\!\!&\!\!spatial test (12)\!\!&\!\!visual test (40)\!\!\\
    \hline
    \!end-to-end & {\bf 88.9\%} & {\bf 83.3\%} & {\bf 62.5\%}\\
    \hline
    \!pose features\!\!& 55.6\% & 58.3\% & 27.5\%\\
    \hline
    \end{tabular}
}
}
\put(0.69,0.010){\parbox{3.9in}{
Success rates on training positions, on novel test positions, and in the presence of visual distractors. The number of trials per test is shown in parentheses.
}
}

\end{picture}
\vspace{-0cm}
\caption{Training and visual test scenes as seen by the policy (left), and experimental results (right). The hammer and bottle images were cropped for visualization only.
}
\label{fig:traintest}
\vspace{-0.4cm}
\end{figure}

\paragraph{Experimental conditions.} We evaluated the visuomotor policies in three conditions: (1) the training target positions and grasps, (2) new target positions not seen during training and, for the hammer, new grasps (spatial test), and (3) training positions with visual distractors (visual test). A selection of these experiments is shown in the supplementary video.\footnote{The video can be viewed at \url{http://sites.google.com/site/visuomotorpolicy}} For the visual test, the shape sorting cube was placed on a table rather than held in the gripper, the coat hanger was placed on a rack with clothes, and the bottle and hammer tasks were done in the presence of clutter. Illustrations of this test are shown in Figure~\ref{fig:traintest}.

\paragraph{Comparison.} The success rates for each test are shown in Figure~\ref{fig:traintest}. We compared to two baselines, both of which train the vision layers in advance for pose prediction, instead of training the entire policy end-to-end. The features baseline discards the last layer of the pose predictor and uses the feature points, resulting in the same architecture as our policy, while the prediction baseline feeds the predicted pose into the control layers. The pose prediction baseline is analogous to a standard modular approach to policy learning, where the vision system is first trained to localize the target, and the policy is trained on top of it. This variant achieves poor performance. As discussed in Section~\ref{sec:poseeval}, the pose estimate is accurate to about 1 cm. However, unlike the tasks in Section~\ref{sec:lg}, where robust controllers could succeed even with inaccurate perception, many of these tasks have tolerances of just a few millimeters. In fact, the pose prediction baseline is only successful on the coat hanger, which requires comparatively little accuracy. Millimeter accuracy is difficult to achieve even with calibrated cameras and checkerboards. Indeed, prior work has reported that the PR2 can maintain a camera to end effector accuracy of about 2 cm during open loop motion \citep{mwgcm-adopi-10}. This suggests that the failure of this baseline is not atypical, and that our visuomotor policies are learning visual features and control strategies that improve the robot's accuracy.
When provided with pose estimation features, the policy has more freedom in how it uses the visual information, and achieves somewhat higher success rates. However, full end-to-end training performs significantly better, achieving high accuracy even on the challenging bottle task, and successfully adapting to the variety of grasps on the hammer task. This suggests that, although the vision layer pretraining is clearly beneficial for reducing computation time, it is not sufficient by itself for discovering good features for visuomotor policies.

\paragraph{Visual distractors.} The policies exhibit moderate tolerance to distractors that are visually separated from the target object. This is enabled in part by the spatial softmax, which has a lateral inhibition effect that suppresses non-maximal activations. Since distractors are unlikely to activate each feature as much as the true object, their activations are therefore suppressed. However, as expected, the learned policies tend to perform poorly under drastic changes to the backdrop, or when the distractors are adjacent to or occluding the manipulated objects, as shown in the supplementary video. A standard solution to this issue to expose the policy to a greater variety of visual situations during training. This issue could also be mitigated by artificially augmenting the image samples with synthetic transformations, as discussed in prior work in computer vision \citep{ssp-bpcnn-03}, or even incorporating ideas from transfer and semi-supervised learning.

\subsection{Features Learned with End-to-End Training}

\begin{figure*}
\setlength{\unitlength}{0.5\columnwidth}
\begin{picture}(2.0,1.5) \linethickness{0.5pt}

\put(0.0,0.91){\includegraphics[width=0.245\columnwidth]{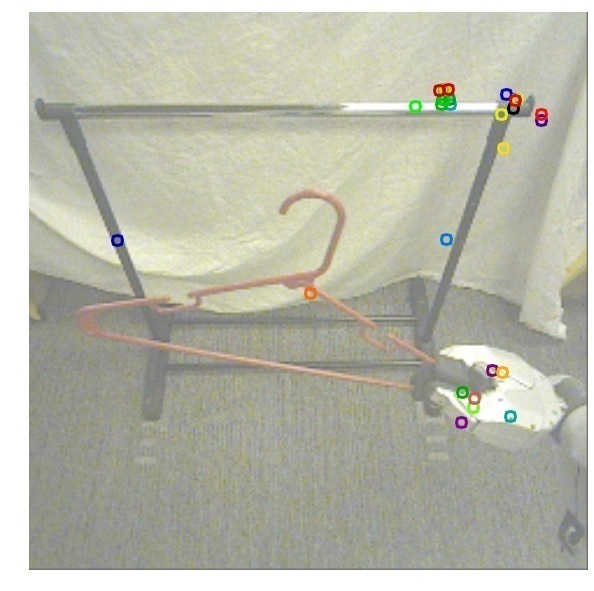}}
\put(0.0,0.46){\includegraphics[width=0.245\columnwidth]{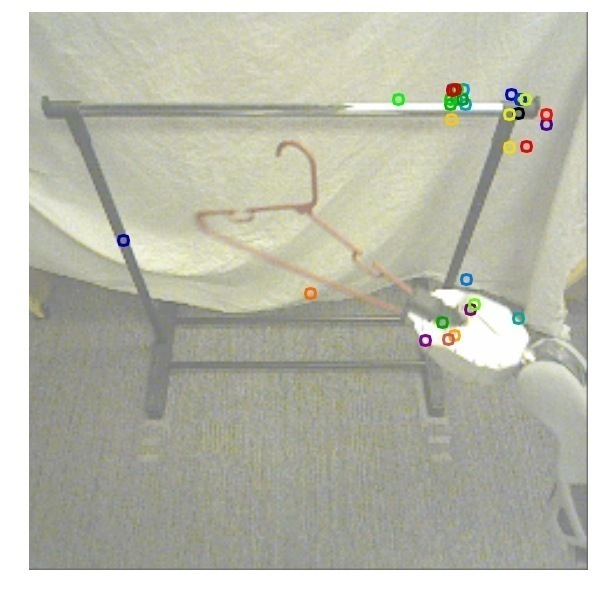}}
\put(0.0,0.01){\includegraphics[width=0.245\columnwidth]{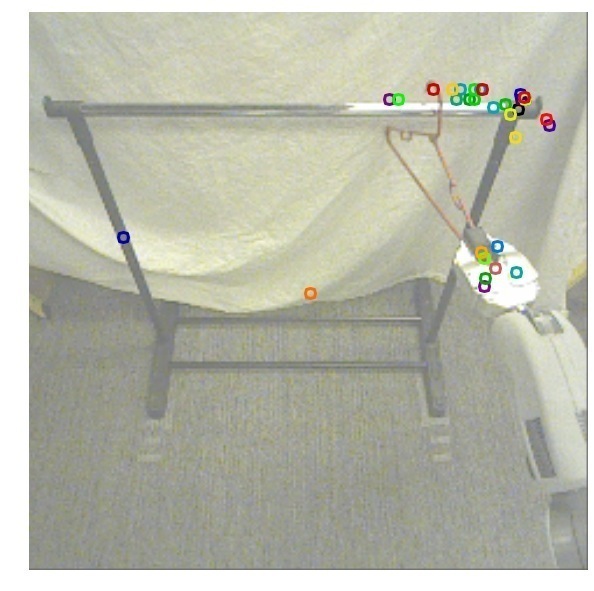}}

\put(0.5,0.91){\includegraphics[width=0.245\columnwidth]{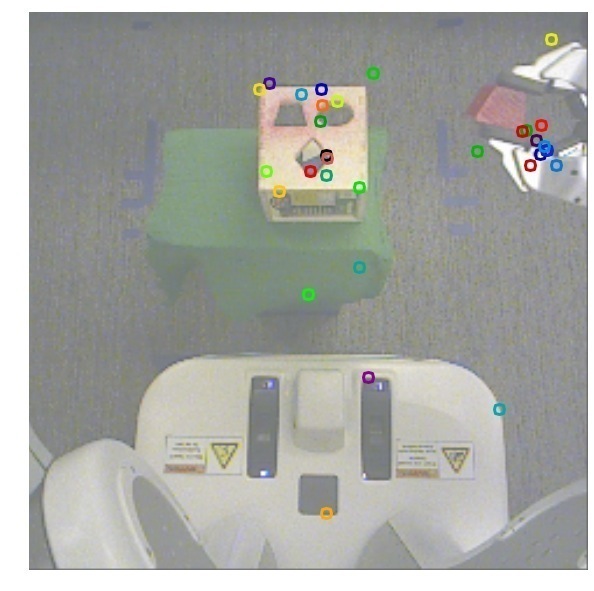}}
\put(0.5,0.46){\includegraphics[width=0.245\columnwidth]{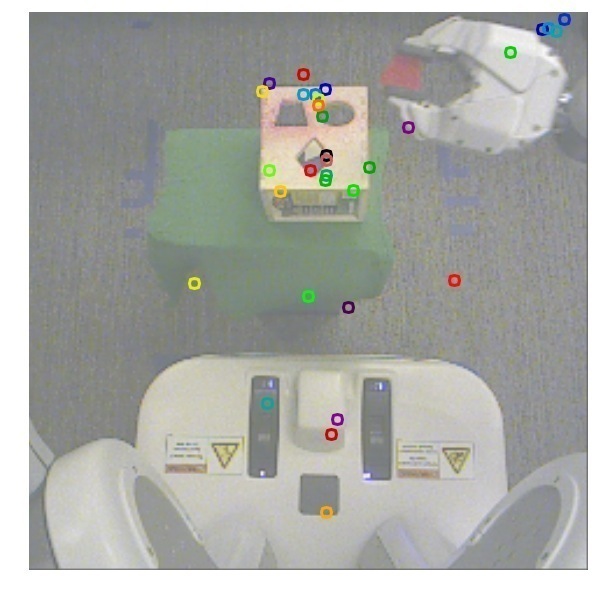}}
\put(0.5,0.01){\includegraphics[width=0.245\columnwidth]{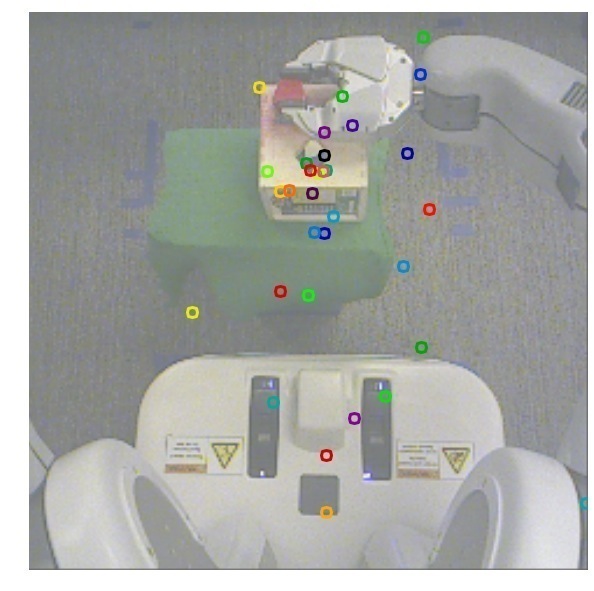}}

\put(1.0,0.91){\includegraphics[width=0.245\columnwidth]{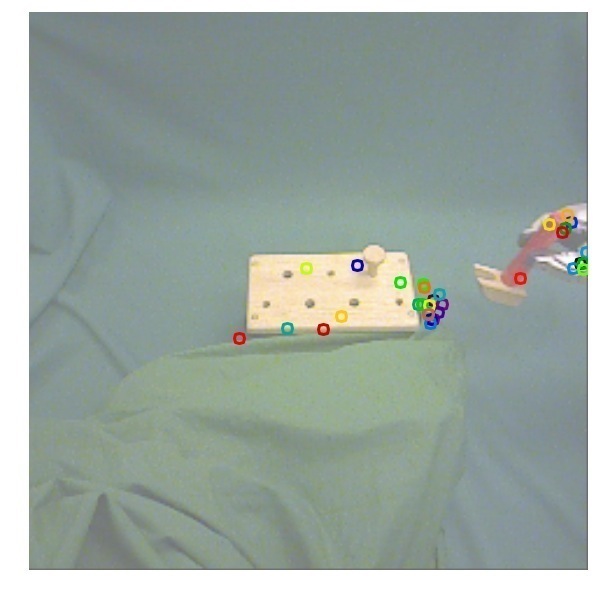}}
\put(1.0,0.46){\includegraphics[width=0.245\columnwidth]{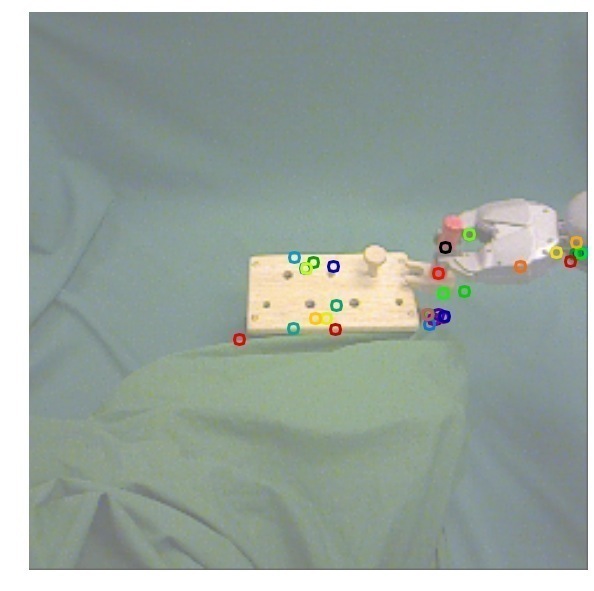}}
\put(1.0,0.01){\includegraphics[width=0.245\columnwidth]{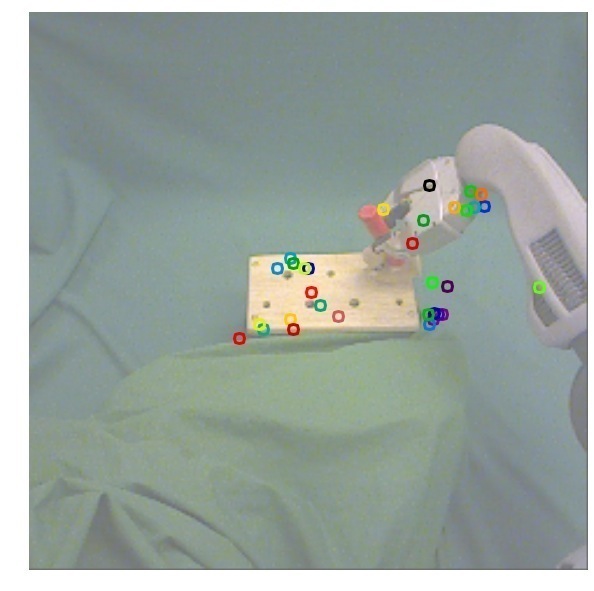}}

\put(1.5,0.91){\includegraphics[width=0.245\columnwidth]{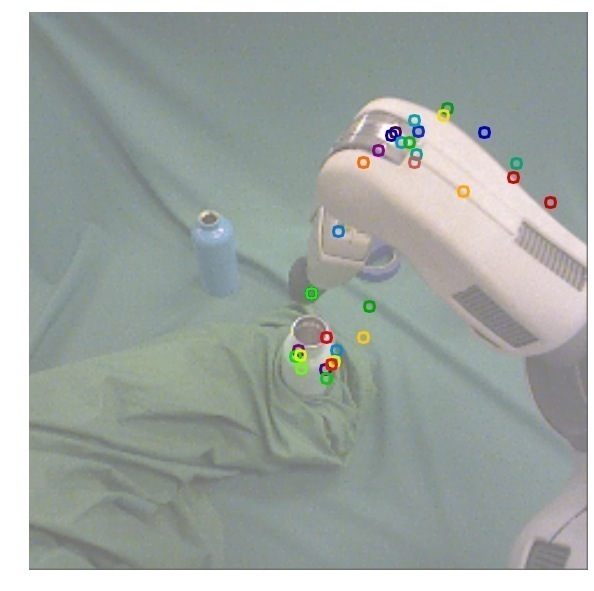}}
\put(1.5,0.46){\includegraphics[width=0.245\columnwidth]{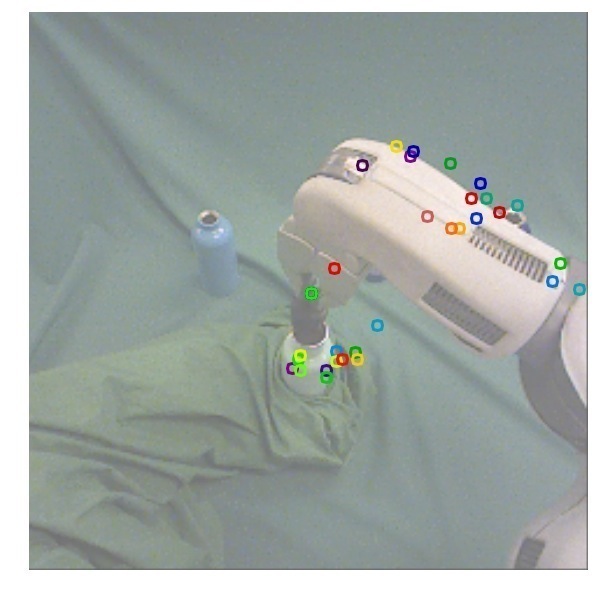}}
\put(1.5,0.01){\includegraphics[width=0.245\columnwidth]{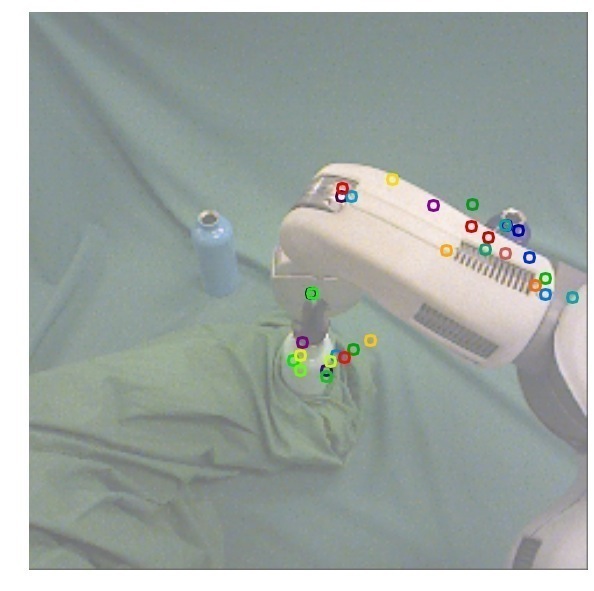}}

\put(0.15,-0.02){(a) hanger}
\put(0.66,-0.02){(b) cube}
\put(1.13,-0.02){(c) hammer}
\put(1.64,-0.02){(d) bottle}

\end{picture}
\caption{Feature points tracked by the policy during task execution for each of the four tasks. Each feature point is displayed in a different random color, with consistent coloring across images. The policy finds features on the target object and the robot gripper and arm. In the bottle cap task, note that the policy correctly ignores the distractor bottle in the background, even though it was not present during training.}
\label{fig:ptstasks}
\end{figure*}

The visual processing layers of our architecture automatically learn features points using the spatial softmax and expectation operators. These feature points encapsulate all of the visual information received by the motor layers of the policy. In Figure~\ref{fig:ptstasks}, we show the features points discovered by our visuomotor policy through guided policy search. Each policy learns features on the target object and the robot manipulator, both clearly relevant to task execution. The policy tends to pick out robust, distinctive features on the objects, such as the left pole of the clothes rack, the left corners of the shape-sorting cube and the bottom-left corner of the toy tool bench. In the bottle task, the end-to-end trained policy outputs points on both sides of the bottle, including one on the cap, while the pose prediction network only finds points on the right edge of the bottle.

\begin{figure*}
\setlength{\unitlength}{0.5\columnwidth}
\begin{picture}(2.0,0.5) \linethickness{0.5pt}

\put(0.0,0.01){\includegraphics[width=0.245\columnwidth]{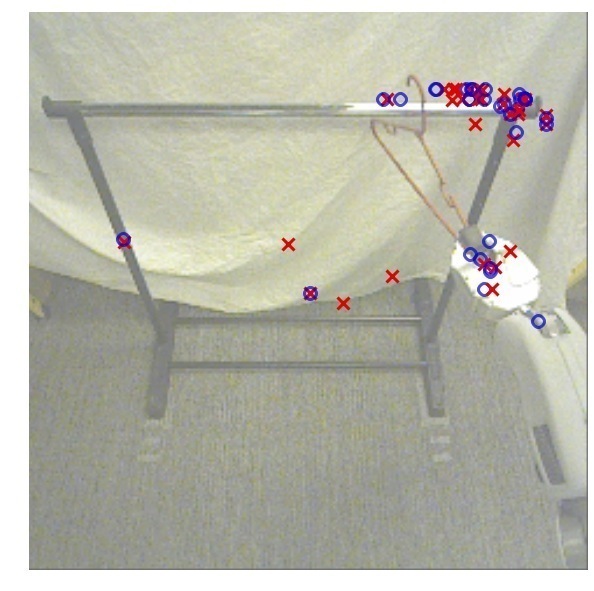}}

\put(0.5,0.01){\includegraphics[width=0.245\columnwidth]{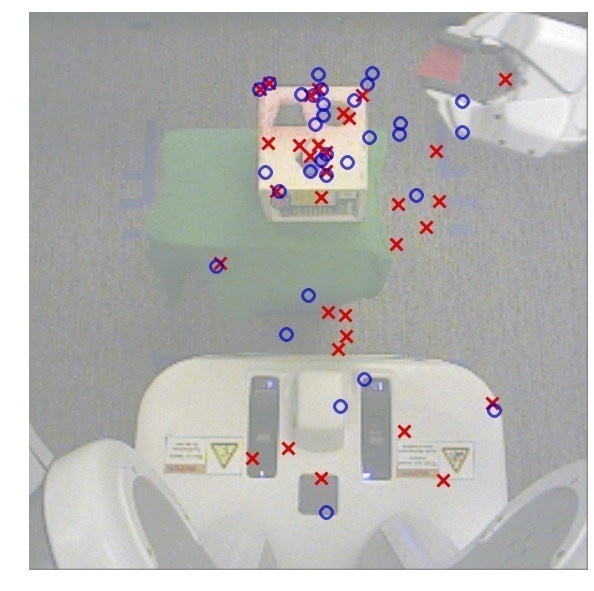}}

\put(1.0,0.01){\includegraphics[width=0.245\columnwidth]{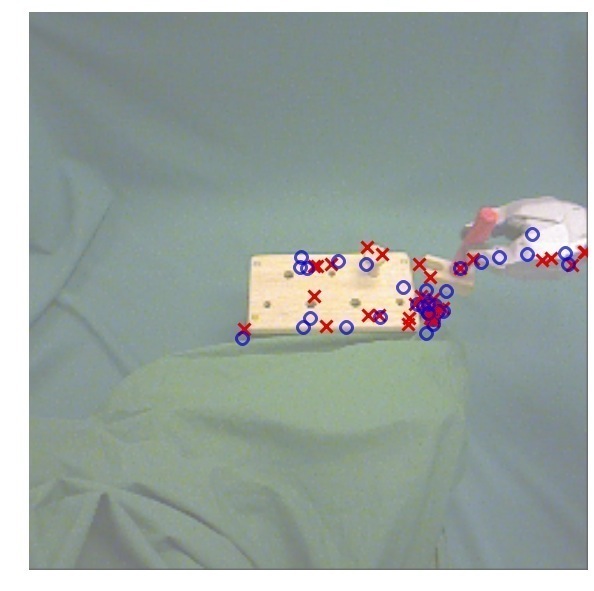}}

\put(1.5,0.01){\includegraphics[width=0.245\columnwidth]{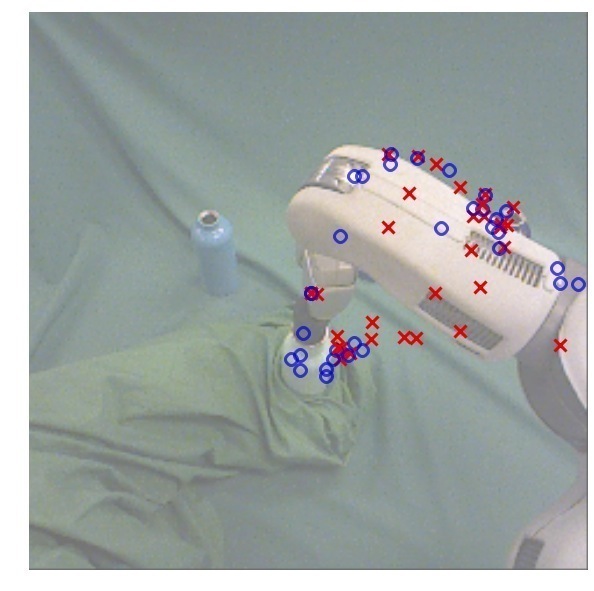}}

\put(0.15,-0.02){(a) hanger}
\put(0.66,-0.02){(b) cube}
\put(1.13,-0.02){(c) hammer}
\put(1.64,-0.02){(d) bottle}

\end{picture}
        \caption{Feature points learned for each task. For each input image, the feature points produced by the policy are shown in blue, while the feature points of the pose prediction network are shown in red. The end-to-end trained policy tends to discover more feature points on the target object and the robot arm than the pose prediction network.}
	\label{fig:comppoints}
\end{figure*}
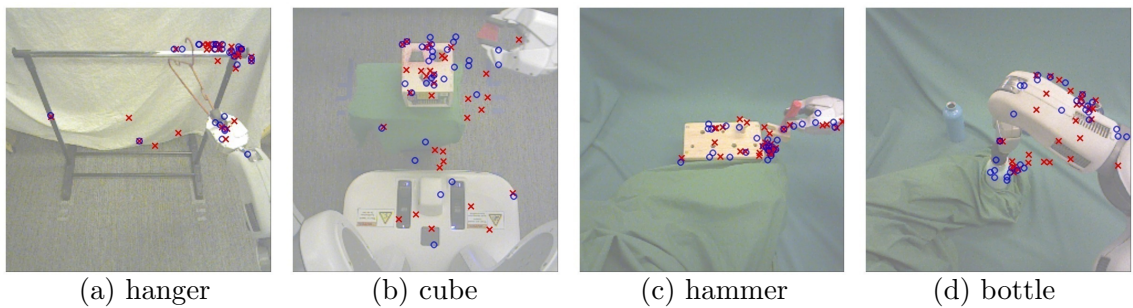

In Figure~\ref{fig:comppoints}, we compare the feature points learned through guided policy search to those learned by a CNN trained for pose prediction. After end-to-end training, the policy acquired a distinctly different set of feature points compared to the pose prediction CNN used for initialization. The end-to-end trained model finds more feature points on task-relevant objects and fewer points on background objects. This suggests that the policy improves its performance by acquiring \emph{goal-driven} visual features that differ from those learned for object localization.

The feature point representation is very simple, since it assumes that the learned features are present at all times, and only one instance of each feature is ever present in the image. While this is a drastic simplification, both the pose predictor and the policy still achieve good results. A more flexible architecture that still learns a concise feature point representation could further improve policy performance. We hope to explore this in future work.

\subsection{Computational Performance and Sample Efficiency}

We used the Caffe deep learning library~\citep{jsdkl-caffe-14} for CNN training. Each visuomotor policy required a total of 3-4 hours of training time: 20-30 minutes for the pose prediction data collection on the robot, 40-60 minutes for the fully observed trajectory pretraining on the robot and offline pose pretraining (which can be done in parallel), and between 1.5 and 2.5 hours for end-to-end training with guided policy search. The coat hanger task required two iterations of guided policy search, the shape sorting cube and the hammer required three, and the bottle task required four. Only about 15 minutes of the training time consisted of executing trials on the robot. Since training was dominated by computation, we expect significant speedup from a more efficient implementation. The number of samples for training each policy is shown in Table~\ref{tbl:samples}. Each trial was five seconds in length, and the numbers do not include the time needed to collect about 1000 images for pretraining the visual processing layers of the policy.

\begin{table}[h]
\begin{center}
\footnotesize{
\begin{tabular}{|c|c|c|c|}
\cline{2-4}
\multicolumn{1}{l|}{}& \multicolumn{3}{c|}{number of trials}\\
\hline
task & trajectory pretraining & end-to-end training & total \\
\hline
coat hanger & 120 & 36 & 156 \\
\hline
shape cube & 90 & 81 & 171 \\
\hline
toy hammer & 150 & 90 & 240 \\
\hline
bottle cap & 180 & 108 & 288 \\
\hline
\end{tabular}
}
\end{center}
\vspace{-0.2in}
\caption{Total number of trials used for learning each visuomotor policy.
\label{tbl:samples}
}
\vspace{-0.2in}
\end{table}

\section{Discussion and Future Work}
\label{sec:conclusion}

In this paper, we presented a method for learning robotic control policies that use raw input from a monocular camera. These policies are represented by a novel convolutional neural network architecture, and can be trained end-to-end using our guided policy search algorithm, which decomposes the policy search problem in a trajectory optimization phase that uses full state information and a supervised learning phase that only uses the observations. This decomposition allows us to leverage state-of-the-art tools from supervised learning, making it straightforward to optimize extremely high-dimensional policies. Our experimental results show that our method can execute complex manipulation skills, and that end-to-end training produces significant improvements in policy performance compared to using fixed vision layers trained for pose prediction.

Although we demonstrate moderate generalization over variations in the scene, our current method does not generalize to dramatically different settings, especially when visual distractors occlude the manipulated object or break up its silhouette in ways that differ from the training. The success of CNNs on exceedingly challenging vision tasks suggests that this class of models is capable of learning invariance to irrelevant distractor features \citep{lbh-dl-15}, and in principle this issue can be addressed by training the policy in a variety of environments, though this poses certain logistical challenges. More practical alternatives that could be explored in future work include simultaneously training the policy on multiple robots, each of which is located in a different environment, developing more sophisticated regularization and pretraining techniques to avoid overfitting, and introducing artificial data augmentation to encourage the policy to be invariant to irrelevant clutter. However, even without these improvements, our method has numerous applications in, for example, an industrial setting where the robot must repeatedly and efficiently perform a task that requires visual feedback under moderate variation in background and clutter conditions.

Our method takes advantage of a known, fully observed state space during training. This is both a weakness and a strength. It allows us to train linear-Gaussian controllers for guided policy search using a very small number of samples, far more efficiently than standard policy search methods. However, the requirement to observe the full state during training limits the tasks to which the method can be applied. In many cases, this limitation is minor, and the only ``instrumentation'' required at training is to position the objects in the scene at consistent positions. However, tasks that require, for example, manipulating freely moving objects require more extensive instrumentation, such as motion capture. A promising direction for addressing this limitation is to combine our method with unsupervised state-space learning, as proposed in several recent works, including our own \citep{rlv-arlrv-12,wsbr-e2c-15,ftddl-lvfsr-15}.

In future work, we hope to explore more complex policy architectures, such as recurrent policies that can deal with extensive occlusions by keeping a memory of past observations. We also hope to extend our method to a wider range of tasks that can benefit from visual input, as well as a variety of other rich sensory modalities, including haptic input from pressure sensors and auditory input. With a wider range of sensory modalities, end-to-end training of sensorimotor policies will become increasingly important: while it is often straightforward to imagine how vision might help to localize the position of an object in the scene, it is much less apparent how sound can be integrated into robotic control. A learned sensorimotor policy would be able to naturally integrate a wide range of modalities and utilize them to directly aid in control.

\section*{Acknowledgements}

This research was funded in part by DARPA through a Young Faculty Award, the Army Research Office through the MAST program, NSF awards IIS-1427425 and IIS-1212798, the Berkeley Vision and Learning Center, and a Berkeley EECS Department Fellowship.

\appendix

\section{Guided Policy Search Algorithm Details}

In this appendix, we describe a number of implementation details of our BADMM-based guided policy search algorithm and our linear-Gaussian controller optimization method.

\subsection{BADMM Dual Variables and Weight Adjustment}
\label{app:badmm}

Recall that the inner loop alternating optimization is given by
\begin{align*}
\params &\leftarrow\! \arg\min_{\params} \sum_{t=1}^T E_{\trajdist(\st)\policy_\params(\at|\st)}[\at\tr\lgmut] + \admmrho_t \badmmpol_t(\params,\trajdist) \\
\trajdist &\leftarrow\! \arg\min_{\trajdist} \sum_{t=1}^T E_{\trajdist(\st,\at)}[\cost(\st,\at) \!-\! \at\tr\lgmut] \!+\!
\admmrho_t \badmmtraj_t(\trajdist,\params) \\
\lgmut &\leftarrow \lgmut + \admmrate \admmrho_t (E_{\policy_\params(\at|\st)\trajdist(\st)}[\at] - E_{\trajdist(\at|\st)\trajdist(\st)}[\at]).
\end{align*}
We use a step size of $\admmrate = 0.1$ in all of our experiments, which we found to be more stable than $\admmrate = 1.0$. The weights $\admmrho_t$ are initialized to $0.01$ and incremented based on the following schedule: at every iteration, we compute the average KL-divergence between $\trajdist(\at|\st)$ and $\policy_\params(\at|\st)$ at each time step, as well as its standard deviation over time steps. The weights $\admmrho_t$ corresponding to time steps where the KL-divergence is higher than the average are increased by a factor of 2, and the weights corresponding to time steps where the KL-divergence is two standard deviations or more below the average are decreased by a factor of 2. The rationale behind this schedule is to adjust the KL-divergence penalty to keep the policy and trajectory in agreement by roughly the same amount at all time steps. Increasing $\admmrho_t$ too quickly can lead to the policy and trajectory becoming ``locked'' together, which makes it difficult for the trajectory to decrease its cost, while leaving it too low requires more iterations for convergence. We found this schedule to work well across all tasks, both during trajectory pretraining and while training the visuomotor policy.

To update the dual variables $\lgmut$, we evaluate the expectations over $\trajdist(\st)$ by using the latest batch of sampled trajectories. For each state $\{\st^i\}$ along these sampled trajectories, we evaluate the expectations over $\at$ under $\policy_\params(\at|\st)$ and $\trajdist(\at|\st)$, which correspond simply to the means of these conditional Gaussian distributions, in closed form.

\subsection{Policy Variance Optimization}

As discussed in Section~\ref{sec:gps}, the variance of the Gaussian policy $\policy_\params(\at|\ot)$ does not depend on the observation, though this dependence would be straightforward to add. Analyzing the objective $\lagpol(\params,\trajdist)$, we can write out only the terms that depend on $\polsig$:
\[
\lagpol(\params,\trajdist) = \frac{1}{2N}\sum_{i=1}^N \sum_{t=1}^T E_{\trajdist_i(\st,\ot)}\left[\trace[\ucovar_{ti}\inv\polsig] - \log|\polsig|\right].
\]
Differentiating and setting the derivative to zero, we obtain the following equation for $\polsig$:
\[
\polsig = \left[\frac{1}{NT}\sum_{i=1}^N\sum_{t=1}^T \ucovar_{ti}\inv \right]\inv,
\]
where the expectation under $\trajdist_i(\st)$ is omitted, since $\ucovar_{ti}$ does not depend on $\st$.

\subsection{Dynamics Fitting}
\label{app:dynamics}

Optimizing the linear-Gaussian controllers $\trajdist_i(\at|\st)$ that induce the trajectory distributions $\trajdist_i(\traj)$ requires fitting the system dynamics $\trajdist_i(\state_{t+1}|\st,\at)$ at each iteration to samples generated on the physical system from the previous controller $\hat{\trajdist}_i(\at|\st)$. In this section, we describe how these dynamics are fitted. As in Section~\ref{sec:gps}, we drop the subscript $i$, since the dynamics are fitted the same way for all of the trajectory distributions.

The linear-Gaussian dynamics are defined as $\trajdist(\state_{t+1}|\st,\at) = \gauss(\fxt\st + \fut\at + \fct, \noise_t)$, and the data that we obtain from the robot can be viewed as tuples $\{\state_t^i,\action_t^i,\state_{t+1}^i\}$. A simple way to fit these linear-Gaussian dynamics is to use linear regression to determine $\fx$, $\fu$, and $\fc$, and fit $\noise_t$ based on the errors. However, the sample complexity of linear regression scales with the dimensionality of $\st$. For a high-dimensional robotic system, we might need an impractically large number of samples at each iteration to obtain a good fit. However, we can observe that the dynamics at nearby time steps are strongly correlated, and we can dramatically reduce the sample complexity of the dynamics fitting by bringing in information from other time steps, and even prior iterations. We will bring in this information by fitting a global model to all of the transitions $\{\state_t^i,\action_t^i,\state_{t+1}^i\}$ for all $t$ and all tuples from several prior iterations (we use three prior iterations in our implementation), and then use this model as a prior for fitting the dynamics at each time step. Note that this global model does not itself need to be a good forward dynamics model -- it just needs to serve as a good prior to reduce the sample complexity of linear regression.

To make it more convenient to incorporate a data-driven prior, we will first reformulate this linear regression fit and view it as fitting a Gaussian model to the dataset $\{\state_t^i,\action_t^i,\state_{t+1}^i\}$ at each time step $t$, and then conditioning this Gaussian to obtain $p(\state_{t+1}|\st,\at)$. While this is equivalent to linear regression, it allows us to easily incorporate a normal-inverse-Wishart prior on this Gaussian in order to bring in prior information. Let $\empsig$ be the empirical covariance of our dataset, and let $\empmu$ be the empirical mean. The normal-inverse-Wishart prior is defined by prior parameters $\priorphi$, $\priormu$, $\priorm$, and $\priorn$. Under this prior, the maximum a posteriori estimates for the covariance $\Sigma$ and mean $\mu$ are given by
\begin{align}
\Sigma = \frac{\priorphi + \empn\empsig + \frac{\empn\priorm}{\empn + \priorm}(\empmu - \priormu)(\empmu - \priormu)\tr}{\empn + \priorn} \hspace{0.5in}
\mu = \frac{\priorm \priormu + \priorn \empmu}{\priorm + \priorn}. \nonumber 
\end{align}
Having obtained $\Sigma$ and $\mu$, we can obtain an estimate of the dynamics $p(\state_{t+1}|\st,\at)$ by conditioning the distribution $\gauss(\mu,\Sigma)$ on $[\st;\at]$, which produces linear-Gaussian dynamics $p(\state_{t+1}|\st,\at) = \gauss(\fxt\st + \fut\at + \fct, \noise_t)$. The parameters of the normal-inverse-Wishart prior are obtained from the global model of the dynamics which, as described previously, is fitted to all available tuples $\{\state_t^i,\action_t^i,\state_{t+1}^i\}$.

The simplest prior can be obtained by fitting a Gaussian distribution to vectors $[\state; \action; \state\pr]$. If the mean and covariance of this data are given by $\bar{\mu}$ and $\bar{\Sigma}$, the prior is given by $\priorphi = \priorn\bar{\Sigma}$ and $\priormu = \bar{\mu}$, while $\priorn$ and $\priorm$ should be set to the number of data points in the datasets. In practice, settings $\priorn$ and $\priorm$ to $1$ tends to produce better results, since the prior is fitted to many more samples than are available for linear regression at each time step. While this prior is simple, we can obtain a better prior by employing a nonlinear model.

The particular global model we use in this work is a Gaussian mixture model over vectors $[\state; \action; \state\pr]$. Systems of articulated rigid bodies undergoing contact dynamics, such as robots interacting with their environment, can be coarsely modeled as having piecewise linear dynamics. The Gaussian mixture model provides a good approximation for such piecewise linear systems, with each mixture element corresponding to a different linear mode \citep{kb-ialsn-10}. Under this model, the state transition tuple is assumed to come from a distribution that depends on some hidden state $h$, which corresponds to the mixture element identity. In practice, this hidden state might correspond to the type of contact profile experienced by a robotic arm at step $i$. The prior for the dynamics fit at time step $t$ is then obtained by inferring the hidden state distribution for the transition dataset $\{\state_t^i,\action_t^i,\state_{t+1}^i\}$, and using the mean and covariance of the corresponding mixture elements (weighted by their probabilities) to obtain $\bar{\mu}$ and $\bar{\Sigma}$. The prior parameters can then be obtained as described above.

In our experiments, we set the number of mixture elements for the Gaussian mixture model prior such that there were at least 40 samples per mixture element, or 20 total mixture elements, whichever was lower. In general, we did not find the performance of the method to be sensitive to this parameter, though overfitting did tend to occur in the early iterations when the number of samples is low, if the number of mixtures was too high.

\subsection{Trajectory Optimization}
\label{app:trajopt}

In this section, we show how the LQR backward pass can be used to optimize the constrained objective in Section~\ref{sec:gpstraj}. The constrained trajectory optimization problem is given by
\[
\min_{\trajdist(\traj)\in\gauss(\traj)} \lagtraj(\trajdist,\params) \,\,\text{s.t.}\,\, \kl(\trajdist(\traj)\|\hat{\trajdist}(\traj)) \leq \epsilon.
\]
The augmented Lagrangian $\lagtraj(\trajdist,\params)$ consists of an entropy term and an expectation under $\trajdist(\traj)$ of a quantity that is independent of $\trajdist$. We can locally approximate this quantity with a quadratic by using a quadratic expansion of $\cost(\st,\at)$, and fitting a linear Gaussian to $\policy_\params(\at|\st)$ with the same method we used for the dynamics. We can then solve the primal optimization in the dual gradient descent procedure with a standard LQR backward pass. As discussed in Section~\ref{sec:gps}, $\lagtraj(\trajdist,\params)$ can be written as the expectation of some function $c(\traj)$ that is independent of $\trajdist$, such that $\lagtraj(\trajdist,\params) = E_{\trajdist(\traj)}[c(\traj)] - \admmrho_t\ent(\trajdist(\traj))$. Specifically,
\[
c(\st,\at) = \cost(\st,\at) - \at\tr\lgmut - \admmrho_t\log\policy_\params(\at|\st)
\]
Writing the Lagrangian of the constrained optimization, we have
\[
\mathcal{L}(\trajdist) = E_{\trajdist(\traj)}[c(\traj) - \eta \log \hat{\trajdist}(\traj)] - (\eta + \admmrho_t)\ent(\trajdist(\traj)) - \eta\epsilon,
\]
where $\eta$ is the Lagrange multiplier. Note that $\mathcal{L}(\trajdist)$ is the Lagrangian of the constrained trajectory optimization, which is not related to the augmented Lagrangian $\lagtraj(\traj,\params)$. Grouping the terms in the expectation and omitting constants, we can rewrite the minimization of the Lagrangian with respect to the primal variables as
\begin{equation}
\min_{\trajdist(\traj)\in\gauss(\traj)} E_{\trajdist(\traj)}\left[\frac{1}{\eta+\admmrho_t}c(\traj) \!-\! \frac{\eta}{\eta+\admmrho_t}\log\hat{\trajdist}(\traj)\right] - \ent(\trajdist(\traj)). \label{eqn:trajobjfinal}
\end{equation}
Let $\tilde{c}(\traj) = \frac{1}{\eta+\admmrho_t}c(\traj) \!-\! \frac{\eta}{\eta+\admmrho_t}\log\hat{\trajdist}(\traj)$. The above optimization corresponds to minimizing $E_{\trajdist(\traj)}[\tilde{c}(\traj)] - \ent(\trajdist(\traj))$. This type of maximum entropy problem can be solved using the LQR algorithm, and the solution is given by
\[
\trajdist(\at|\st) = \gauss(\Kpol_t\st + \kpol_t ; \Quut\inv),
\]
where $\Kpol_t$ and $\kpol_t$ are the feedback and open loop terms of the optimal linear feedback controller corresponding to the cost $\tilde{c}(\st,\at)$ and the dynamics $\trajdist(\state_{t+1}|\st,\at)$, and $\Quut$ is the quadratic term in the Q-function at time step $t$. All of these terms can be obtained from a standard LQR backward pass \citep{lt-ilqr-04}, which we summarize below.

Recall that the estimated linear-Gaussian dynamics have the form $\trajdist(\state_{t+1}|\st,\at) = \gauss(\fxt\st + \fut\at + \fct, \noise_t)$. The quadratic cost approximation has the form
\[
\tilde{c}(\st,\at) \approx \frac{1}{2}[\st;\at]\tr\tchesst[\st;\at] + [\st;\at]\tr\tcgradt + \text{const},
\]
\noindent where subscripts denote derivatives, e.g. $\tcgradt$ is the gradient of $\tilde{c}$ with respect to $[\st;\at]$, while $\tchesst$ is the Hessian.\footnote{We assume that all Taylor expansions here are recentered around zero. Otherwise, the point around which the derivatives are computed must be subtracted from $\st$ and $\at$ in all of these equations.} Under this model of the dynamics and cost function, the optimal controller can be computed by recursively computing the quadratic $Q$-function and value function, starting with the last time step. These functions are given by
\begin{align*}
V(\st) &= \frac{1}{2}\st\tr\Vxxt\st + \st\tr\Vxt + \text{const} \\
Q(\st,\at) &= \frac{1}{2}[\st; \at]\tr\Qyyt[\st; \at] + [\st; \at]\tr \Qyt + \text{const}
\end{align*}
We can express them with the following recurrence, which is computed starting at the last time step $t = T$ and moving backward through time:
\begin{align*}
\Qyyt &= \tchesst + \fyt\tr\Vxxtp\fyt \\
\Qyt &= \tcgradt + \fyt\tr\Vxtp + \fyt\tr\Vxxtp\fct \\
\Vxxt &= \Qxxt - \Quxt\tr\Quut\inv\Quxt \\
\Vxt &= \Qxt - \Quxt\tr\Quut\inv\Qut,
\end{align*}
\noindent and the optimal control law is then given by \mbox{$\detpolicy(\state_t) = \Kpol_t\st + \kpol_t$}, where $\Kpol_t = -\Quut\inv \Quxt$ and $\kpol_t = -\Quut\inv \Qut$. If, instead of simply minimizing the expected cost, we instead wish to optimize the maximum entropy objective in Equation~(\ref{eqn:trajobjfinal}), the optimal controller is instead linear-Gaussian, with the solution given by $\trajdist(\at|\st) = \gauss(\Kpol_t\st + \kpol_t ; \Quut\inv)$, as shown in prior work \citep{lk-gps-13}.

\section{Experimental Setup Details}
\label{app:tasks}

In this appendix, we present a detailed summary of the experimental setup for our simulated and real-world experiments.

\subsection{Simulated Experiment Details}
\label{app:taskssim}

All of the simulated experiments used the MuJoCo simulation package~\citep{tet-mjc-12}, with simulated frictional contacts and torque motors at the joints used for actuation. Although no control or state noise was added during simulation, noise was injected naturally by the linear-Gaussian controllers. The linear-Gaussian controllers $\trajdist_i(\at|\st)$ were initialized to stay near the initial state $\state_1$ using linear feedback based on a proportional-derivative control law for all tasks, except for the octopus arm, where $\trajdist_i(\at|\st)$ was initialized to be zero mean with a fixed spherical covariance, and the walker, which was initialized to track a demonstration trajectory with proportional-derivative feedback. The walker was the only task that used a demonstration, as described previously. We describe the details of each system below.

\paragraph{Peg insertion:} The 2D peg insertion task has 6 state dimensions (joint angles and angular velocities) and 2 action dimensions. The 3D version of the task has 12 state dimensions, since the arm has 3 degrees of freedom at the shoulder, 1 at the elbow, and 2 at the wrist. Trials were 8 seconds in length and simulated at 100 Hz, resulting in 800 time steps per rollout. The cost function is given by
\[
\cost(\st,\at) = \frac{1}{2}\torquepen \|\at\|^2 + \pospen \lscnorm{\pos_{\st} - \pos^\star},
\]
\noindent where $\pos_{\st}$ is the position of the end effector for state $\st$, $\pos^\star$ is the desired end effector position at the bottom of the slot, and the norm $\lscnorm{z}$ is given by $\frac{1}{2}\|z\|^2 + \sqrt{\alpha + z^2}$, which corresponds to the sum of an $\ell_2$ and soft $\ell_1$ norm. We use this norm to encourage the peg to precisely reach the target position at the bottom of the hole, but to also receive a larger penalty when far away. The task also works well in 2D with a simple $\ell_2$ penalty, though we found that the 3D version of the task takes longer to insert the peg all the way into the hole without the $\ell_1$-like square root term. The weights were set to $\torquepen = 10^{-6}$ and $\pospen = 1$. Initial states were chosen by moving the shoulder of the arm relative to the hole, with four equally spaced starting states in a 20 cm region for the 2D arm, and four random starting states in a 10 cm radius for the 3D arm.

\paragraph{Octopus arm:} The octopus arm consists of six four-sided chambers. Each edge of each chamber is a simulated muscle, and actions correspond to contracting or relaxing the muscle. The state space consists of the positions and velocities of the chamber vertices. The midpoint of one edge of the first chamber is fixed, resulting in a total of $25$ degrees of freedom: the 2D positions of the $12$ unconstrained points, and the orientation of the first edge. Including velocities, the total dimensionality of the state space is $50$. The cost function depends on the activation of the muscles and distance between the tip of the arm and the target point, in the same way as for peg insertion. The weights are set to $\torquepen = 10^{-3}$ and $\pospen = 1$.

\paragraph{Swimmer:} The swimmer consists of 3 links and 5 degrees of freedom, including the global position and orientation which, together with the velocities, produces a 10 dimensional state space. The swimmer has 2 action dimensions corresponding to the torques between joints. The simulation applied drag on each link of the swimmer to roughly simulate a fluid, allowing it to propel itself. The rollouts were 20 seconds in length at 20 Hz, resulting in 400 time steps per rollout. The cost function for the swimmer is given by
\[
\cost(\st,\at) = \frac{1}{2}\torquepen \vnorm{\at}^t + \frac{1}{2}\velpen \vnorm{{\velx}_{\st}-\velx^\star}^2
\]
\noindent where ${\velx}_{\st}$ is the horizontal velocity, $\velx^\star = 2.0\ms$, and the weights were $\torquepen = 2\cdot 10^{-5}$ and $\velpen = 1$.

\paragraph{Walker:} The bipedal walker consists of a torso and two legs, each with three links, for a total of 9 degrees of freedom and 18 dimensions, with velocity, and 6 action dimensions. The simulation ran for 5 seconds at 100 Hz, for a total of 500 time steps. The cost function is given by
\[
\cost(\st,\at) = \frac{1}{2}\torquepen \vnorm{\at}^t + \frac{1}{2}\velpen \vnorm{{\velx}_{\st}-\velx^\star}^2 + \frac{1}{2}\heightpen \vnorm{{\posy}_{\st}-\posy^\star}^2
\]
\noindent where ${\velx}_{\st}$ is again the horizontal velocity, ${\posy}_{\st}$ is the vertical position of the root, $\velx^\star = 2.1\ms$, $\posy^\star = 1.1\text{m}$, and the weights were set to $\torquepen = 10^{-4}$, $\velpen = 1$, and $\heightpen = 1$.

\subsection{Robotic Experiment Details}
\label{app:tasksviz}

All of the robotic experiments were conducted on a PR2 robot. The robot was controlled at 20 Hz via direct effort control,\footnote{The PR2 robot does not provide for closed loop torque control, but instead supports an effort control interface that directly sets feedforward motor voltages. In practice, these voltages are roughly proportional to feedforward torques, but are also affected by friction and damping.} and camera images were recorded using the RGB camera on a PrimeSense Carmine sensor. The images were downsampled to $240 \times 240 \times 3$. The learned policies controlled one 7 DoF arm of the robot, while the other arm was used to move objects in the scene to automatically vary the initial conditions. The camera was kept fixed in each experiment. Each episode was $5$ seconds in length. For each task, the cost function required placing the object held in the gripper at a particular location (which might require, for example, to insert a shape into a shape sorting cube). The cost was given by the following equation:
\[
\cost(\st,\at) = w_{\ell_2} d_t ^ 2 + w_{\log} \log(d_t^2 + \alpha) + w_{\action}\vnorm{\at}^2,
\]
\noindent where $d_t$ is the distance between three points in the space of the end-effector and their target positions,\footnote{Three points fully define the pose of the end-effector. For the bottle cap task, which is radially symmetric, we use only two points.}, and the weights are set to $w_{\ell_2} = 10^{-3}$, $w_{\log} = 1.0$, and $w_{\action} = 10^{-2}$. The quadratic term encourages moving the end-effector toward the target when it is far, while the logarithm term encourages placing it precisely at the target location, as discussed in prior work \citep{lwa-lnnpg-15}. The bottle cap task used an additional cost term consisting of a quadratic penalty on the difference between the wrist angular velocity and a target velocity.

For all of the tasks, we initialized all of the linear-Gaussian controllers $\trajdist_i(\at|\st)$ to stay near the initial state $\state_1$, with a diagonal noise covariance. The covariance of the noise was chosen to be proportional to a diagonal approximation of the inverse effective mass at each joint, as provided by the manufacturer of the PR2 robot, and the feedback controller was constructed using LQR, with an approximate linear model obtained from the same diagonal inverse mass matrix. The role of this initial controller was primarily to avoid dangerous actions during the first iteration. We discuss the particular setup for each experiment below:

\paragraph{Coat hanger:} The coat hanger task required the robot to hang a coat hanger on a clothes rack. The coat hanger was grasped at one of two angles, about $35^\circ$ apart, and the rack was positioned at three different distances from the robot during training, with differences of about $10\text{ cm}$ between each position. The rack was moved manually between these positions during training. A trial was considered successful if, when the coat hanger was released, it remained hanging on the rack rather than dropping to the ground.

\paragraph{Shape sorting cube:} The shape sorting cube task required the robot to insert a red trapezoid into a trapezoidal hole on a shape sorting cube. During training, the cube was positioned at nine different positions, situated at the corners, edges, and middle of a rectangular region $16\text{ cm} \times 10\text{ cm}$ in size. During training, the shape sorting cube was moved through the training positions by using the left arm. A trial was considered successful if the bottom face of the trapezoid was completely inside the shape sorting cube, such that if the robot were to release the trapezoid, it would fall inside the cube.

\paragraph{Toy hammer:} The hammer task required the robot to insert the claw of a toy hammer underneath a toy plastic nail, placing the claw around the base of the nail. The hammer was grasped at one of three angles, each $22.5^\circ$ apart, for a total variation of $45^\circ$ degrees, and the nail was positioned at five positions, at the corners and center of a rectangular region $10\text{ cm} \times 7\text{ cm}$ in size. During training, the toy tool bench containing the nail was moved using the left arm. A trial was considered successful if the tip of the claw of the hammer was at least under the centerline of the nail.

\paragraph{Bottle cap:} The bottle cap task required the robot to screw a cap onto a bottle at various positions. The bottle was located at nine different positions, situated at the corners, edges, and middle of a rectangular region $16\text{ cm} \times 10\text{ cm}$ in size, and the left arm was used to move the bottle through the training positions. A trial was considered successful if, after completion, the cap could not be removed from bottle simply by pulling vertically.

\vskip 0.2in
\bibliography{references}

\end{document}